\documentclass[11pt,letterpaper]{article}

\usepackage{fix-cm}
\usepackage{amsfonts,amssymb,graphicx,array,amsthm}
\usepackage{tgtermes,newtxtext,newtxmath,bm}
\usepackage{mathtools,booktabs,multirow,tabularx}
\usepackage{algorithm,algpseudocode,enumitem,microtype}
\usepackage[dvipsnames]{xcolor}
\usepackage{tikz}
\usetikzlibrary{arrows.meta,positioning,calc}
\usepackage[sort]{natbib}
\bibpunct[, ]{(}{)}{,}{a}{}{,}

\def\newblock{\ }
\usepackage[hidelinks,hypertexnames=false]{hyperref}
\usepackage[nameinlink,noabbrev,capitalise]{cleveref}

\hypersetup{pdftitle={Flexible Near-Optimal Policies under Model Uncertainty}}
\newcolumntype{L}[1]{>{\raggedright\arraybackslash}p{#1}}
\newcolumntype{Y}{>{\raggedright\arraybackslash}X}
\crefname{assumption}{Assumption}{Assumptions}
\Crefname{assumption}{Assumption}{Assumptions}
\newcommand{\Sset}{\mathcal S}
\newcommand{\Aset}{\mathcal A}
\newcommand{\Uset}{\mathcal U}

\newcommand{\Pset}{\mathcal P}
\newcommand{\Policies}{\mathfrak P}
\newcommand{\Selectors}{\Sigma}
\newcommand{\MenuFamily}{\bm{\Pi}}
\newcommand{\Fexact}{\mathfrak F_{\Delta}}
\newcommand{\Fcert}{\mathfrak F_{\Delta}^{\mathrm{cert}}}
\newcommand{\menu}{\Pi}
\newcommand{\Reg}{\mathcal L}
\newcommand{\Bell}{\mathcal T}
\newcommand{\one}{\bm 1}
\newcommand{\rlo}{\underline r}
\newcommand{\rup}{\overline r}
\newcommand{\plo}{\underline p}
\newcommand{\pup}{\overline p}
\providecommand{\argmax}{}
\renewcommand{\argmax}{\operatorname*{arg\,max}}

\newcommand{\norminf}[1]{\left\lVert #1\right\rVert_\infty}
\newcommand{\abs}[1]{\left|#1\right|}

\usepackage{longtable}

\usepackage{needspace}
\usepackage{placeins}

\usepackage[letterpaper,textwidth=6.5in,textheight=9in,centering]{geometry}
\usepackage{etoolbox}
\usepackage[font=small,labelfont=bf,labelsep=space]{caption}

\makeatletter
\renewcommand\normalsize{\@setfontsize\normalsize{11}{12.66}%
  \abovedisplayskip11pt plus3pt minus5pt
  \belowdisplayskip\abovedisplayskip
  \abovedisplayshortskip0pt plus3pt
  \belowdisplayshortskip6pt plus3pt minus3pt
  \let\@listi\@listI}
\renewcommand\small{\@setfontsize\small{10}{11.5}%
  \abovedisplayskip8pt plus3pt minus5pt
  \belowdisplayskip\abovedisplayskip}
\renewcommand\footnotesize{\@setfontsize\footnotesize{9}{10.35}}
\renewcommand\scriptsize{\@setfontsize\scriptsize{8}{9.5}}
\normalsize
\renewcommand\section{\@startsection{section}{1}{0pt}%
  {-12pt plus-4pt minus-2pt}{5pt}{\large\bfseries\raggedright}}
\renewcommand\subsection{\@startsection{subsection}{2}{0pt}%
  {-9pt plus-3pt minus-1pt}{4pt}{\normalsize\bfseries\raggedright}}
\renewcommand\subsubsection{\@startsection{subsubsection}{3}{0pt}%
  {6pt plus2pt minus1pt}{-1em}{\normalsize\bfseries}}
\renewcommand\@seccntformat[1]{\csname the#1\endcsname.\quad}
\makeatother
\newcommand\ABSTRACT[1]{\par\medskip\noindent\textbf{Abstract.} #1\par}
\newcommand\Halmos{\mbox{\quad$\square$}}
\renewenvironment{proof}[1]{\par\addvspace{4pt}\noindent\hspace*{1em}\textit{#1}\enspace\ignorespaces}{\par\addvspace{4pt}}
\newtheoremstyle{paperplain}{5pt}{5pt}{\itshape}{1em}{\normalfont\sffamily\scshape}{.}{0.5em}{}
\newtheoremstyle{paperdefinition}{5pt}{5pt}{\normalfont}{1em}{\normalfont\sffamily\scshape}{.}{0.5em}{}
\theoremstyle{paperplain}
\newtheorem{theorem}{Theorem}
\newtheorem{lemma}{Lemma}
\newtheorem{proposition}{Proposition}
\newtheorem{corollary}{Corollary}
\newtheorem{claim}{Claim}

\theoremstyle{paperdefinition}
\newtheorem{remark}{Remark}

\newtheorem{definition}{Definition}

\newenvironment{repeattheorem}[1][]{\par\addvspace{5pt}\noindent\hspace*{1em}{\normalfont\sffamily\scshape #1}\enspace\itshape\ignorespaces}{\par\addvspace{5pt}}
\newenvironment{repeatproposition}[1][]{\begin{repeattheorem}[#1]}{\end{repeattheorem}}
\makeatletter
\newcommand\ECNumber[1]{\setcounter{#1}{0}%
  \expandafter\def\csname the#1\endcsname{EC.\arabic{#1}}%
  \expandafter\def\csname theH#1\endcsname{EC.\arabic{#1}}}
\newenvironment{APPENDICES}{%
  \appendix
  \forcsvlist{\ECNumber}{theorem,lemma,proposition,corollary,claim,conjecture,hypothesis,assumption,remark,example,problem,definition,question,answer,exercise,equation,table,figure,algorithm}%
  \small
  \renewcommand\section{\@startsection{section}{1}{0pt}%
    {-10pt plus-4pt minus-1pt}{4pt}{\normalsize\bfseries\raggedright}}%
  \renewcommand\subsection{\@startsection{subsection}{2}{0pt}%
    {-7pt plus-3pt minus-1pt}{3pt}{\small\bfseries\raggedright}}%
  \renewcommand\subsubsection{\@startsection{subsubsection}{3}{0pt}%
    {5pt plus2pt minus1pt}{-1em}{\small\bfseries}}%
  \renewcommand\@seccntformat[1]{\ifstrequal{##1}{section}{Appendix \thesection: }{\csname the##1\endcsname.\quad}}%
  \parindent1em
}{\par}
\makeatother

\hypersetup{
  pdftitle={Flexible Near-Optimal Policies under Model Uncertainty},
  pdfauthor={Zequn Hareton Song; Wesley Marrero},
  pdfsubject={Preprint, November 2026}
}
\begin{document}
\begin{center}
{\LARGE\bfseries Flexible Near-Optimal Policies under Model Uncertainty\par}
\vspace{1em}
\begin{minipage}[t]{0.49\textwidth}\centering
{\large Zequn Hareton Song}\par\medskip
Princeton University\par
Operations Research and Financial Engineering\par
\href{mailto:zs5958@alumni.princeton.edu}{zs5958@alumni.princeton.edu}
\end{minipage}\hfill
\begin{minipage}[t]{0.49\textwidth}\centering
{\large Wesley Marrero}\par\medskip
Dartmouth College\par
Thayer School of Engineering\par
\href{mailto:Wesley.Marrero@dartmouth.edu}{Wesley.Marrero@dartmouth.edu}
\end{minipage}
\par\vspace{1em}
Preprint, November 2026
\end{center}
\ABSTRACT{Decision support systems can accommodate human preferences by recommending several high-performing alternatives at each decision epoch. Repeated choices among these alternatives can reduce performance as small losses accumulate, especially under model uncertainty. To address this challenge, we develop an approach to obtain near-optimal set-valued policies for Markov decision processes with uncertain rewards and transition probabilities. We aim to retain as many actions as possible in the set-valued policy while limiting performance loss across a set of plausible models. This loss is the difference between the optimal expected cumulative reward and that achieved by choosing from the near-optimal recommended sets under the same model, regardless of how actions are selected over time. For rectangular uncertainty, we give an exact verification formulation and establish NP-hardness of finding violations and maximizing flexibility. We derive loss bounds for constructing set-valued policies. For continuous model families, we compare action values under shared uncertain parameters and account for the loss from potential subsequent choices. Controlled experiments on synthetic interval models compare the flexibility of the resulting policies with exact solutions. A separate hypertension case study evaluates personalized models for a large patient population with uncertain future blood pressure and cholesterol. Among approximately 32.0 million adults with multiple admissible treatments, we certify alternatives for 93.6\% at a loss limit of 0.03 quality-adjusted life-years (QALYs). Within certified sets, worst-performing policies lose a survey-weighted mean of 0.0067 QALYs relative to the unrestricted nominal optimum. This approach could help clinical decision support offer treatment choices within an explicit loss limit.}

\par\bigskip
\section{Introduction}

\label{sec:introduction}

Markov decision processes (MDPs) offer a mathematical framework for modeling sequential decision making under uncertainty. Discrete-time MDPs with finite states and actions have been used to inform decisions in various applications, such as healthcare operations and medical decision making. In such critical settings, the ultimate decision is usually made by a human. A single policy, which prescribes one action for each state and decision epoch, may not adequately accommodate how each decision maker chooses among options  \citep{fardpineau}. When multiple actions perform similarly, the performance difference between an optimal policy and other alternatives may not be relevant to the decision maker. In addition, human decision makers often hold information that a model omits, such as their expertise or preferences. A single policy might also be unreliable. Model parameters are typically estimated from observational data and other sources, and estimation error can distort the predicted consequences of an action. Even a policy that is optimal under an estimated model may perform poorly in the actual system \citep{mannor2007}. A reliable recommendation should therefore hold up under all parameter values that the evidence cannot rule out. Decision makers would benefit from alternatives that remain near-optimal under every plausible parameter value, leaving the final choice to their judgment. Each choice may cost little, but the losses accumulate over decision epochs.

These considerations call for a flexible approach to decision support, in which decision makers can choose among actions recommended by a set-valued policy while retaining a guarantee on overall performance. We develop a framework for constructing these policies in MDPs with finite state and action spaces and uncertain rewards and transition probabilities. Our goal is to maximize the total number of recommended actions across states, subject to a prescribed limit on performance loss. Each admissible combination of parameter values defines a plausible model. Under each possible model, every recommended policy must yield an expected total discounted reward within a specified amount of that model’s optimal value. We characterize the resulting optimization problem and develop computational methods to construct and certify these recommendations. Certification verifies that the recommendations satisfy the prescribed loss limit under every model considered. We evaluate the methods in controlled experiments and a hypertension case study with uncertain future blood pressure and cholesterol readings.

\subsection{Medical Decision Making and Atherosclerotic Cardiovascular Disease}

The ideas in this paper are motivated by medical decision making and the management of atherosclerotic cardiovascular disease (ASCVD) via hypertension treatment. Predominantly manifested through myocardial infarction and stroke, ASCVD is a leading cause of death in the United States \citep{kochanek2024}. National data show that coronary heart disease, mainly expressed as myocardial infarction events, and stroke are responsible for 40.3\% and 17.5\% of cardiovascular disease-related deaths, respectively \citep{martin2024}. High blood pressure (BP), or hypertension, is a key risk factor for ASCVD \citep{lamprea2018}. Proper management of hypertension is crucial in reducing the incidence and prevalence of ASCVD.

Hypertension management relies on clinical guidelines \citep{wheltonguideline2017} and analytic approaches \citep{garciahypertension,schell2016}. However, clinical guidelines may be perceived as controversial and subjective \citep{solberg2018,cohen2018}, as they often rely on expert judgment to interpret and synthesize the latest evidence. Analytic approaches typically provide a single treatment recommendation \citep{garciahypertension,steimle,zargoush2018,schell2016}, offering limited flexibility to accommodate physicians' and their patients' preferences. Despite their potential benefit, time constraints in interpreting extensive advice and disagreement with recommendations thwart physicians’ adoption of hypertension treatment protocols \citep{abdalla2023,kataria2021}. Our approach can offer several treatments at each decision epoch, so physicians and patients can choose according to their preferences. Any sequence of these choices stays within a stated loss in quality-adjusted life-years under every plausible forecast of blood pressure and cholesterol.

\subsection{Contributions}

We develop a framework for constructing near-optimal set-valued policies in MDPs with uncertain rewards and transition probabilities. The framework aims to enhance decision-making flexibility by maximizing the total number of recommended actions across states and decision epochs, subject to a prescribed bound on policy performance loss. The recommended actions at a state may reduce to a singleton, so the performance constraint determines how much flexibility can be retained. Our contributions are as follows.

\begin{enumerate}[leftmargin=1.7em]

\item We develop an exact verification method for the performance guarantee and establish its computational difficulty for independent uncertainty across state-action pairs. The guarantee jointly evaluates the optimal policy and the least favorable choices from the recommended sets under the same model. This evaluation provides an exact test of whether the recommendations satisfy the loss limit throughout the uncertainty set. We prove that finding a model that violates this guarantee is NP-hard, even when only one action is recommended at each state, and that maximizing a weighted measure of flexibility is NP-hard.

\item We develop certified augmentation algorithms for constructing set-valued policies with provable performance guarantees. For uncertainty specified independently across state-action pairs, we derive computable bounds on cumulative performance loss that account for uncertainty shared by the recommendations and an optimal benchmark. The algorithms expand the recommended sets by accepting additional actions whenever the resulting sets satisfy the prescribed loss bound. We establish the validity of these bounds and characterize computational cost in terms of problem size, value updates, and action additions.

\item We establish performance guarantees for continuous model families governed by shared uncertain parameters. These parameters can affect multiple rewards and transition probabilities simultaneously. We bound performance loss over regions covering the entire parameter domain, using direct action-value comparisons that retain the dependence induced by shared parameters. By accounting for current and subsequent action choices, the resulting bounds certify cumulative performance across the full model family, regardless of how the recommended actions are selected over time.

\item Through experiments, we quantify the computational performance and recommendation flexibility of the proposed methods. Controlled experiments compare the certified construction with exact solutions, measuring computational cost and the number of recommended actions retained. We examine how these outcomes vary with problem size, the allowed performance loss, and model uncertainty.

\item We further evaluate the framework through a hypertension treatment case study. Using patient-specific models with uncertain future blood pressure and cholesterol readings, we examine whether and to what extent the methods can retain multiple treatment alternatives while satisfying a prescribed limit on expected loss in quality-adjusted life-years (QALYs). We assess the resulting treatment flexibility and the modeled consequences of selecting among these alternatives at the patient and population levels. At a loss limit of 0.03 QALYs, we certify alternatives for 93.6\% of the 32.0 million adults with multiple admissible treatments. Ignoring forecast uncertainty, our construction meets that limit for only 0.2\%.

\end{enumerate}

\subsection{Organization of the Paper}

The remainder of this paper is organized as follows. Section 2 reviews the related literature. Section 3 introduces the MDP setting, the representation of model uncertainty, and the optimization problem of constructing near-optimal set-valued policies. Section 4 presents the exact verification method and certified augmentation algorithms, establishes their computational complexity and performance guarantees, and develops certification methods for continuous model families with shared uncertain parameters. Section 5 reports controlled experiments evaluating computational performance and recommendation flexibility under different performance-loss limits and levels of uncertainty. Section 6 evaluates the framework through a hypertension treatment case study, examining treatment flexibility and modeled outcomes under uncertainty in future blood pressure and cholesterol. Section 7 discusses the main findings, limitations, and directions for future research. Proofs and experimental details are provided in the electronic companion.

\FloatBarrier
\section{Related Literature}
\label{sec:related-work}

Our work draws on research from set-valued policies, robust and regret-based control, permissive controller synthesis, and medical decision making. We examine the choices these approaches preserve and the performance guarantees they provide under model uncertainty.

\subsection{Set-valued Policies and Permissive Control}

Set-valued policies assign a nonempty set of actions to each state, leaving the decision maker to choose among the recommended alternatives. For an MDP with known rewards and transition probabilities, \citet{fardpineau} introduce such policies as nondeterministic policies. They develop mixed-integer optimization and approximation methods to retain actions while ensuring near-optimal performance. \citet{tang} use the same policy representation and develop a model-free learning method. Both formulations evaluate the least favorable choices from the sets, accounting for the cumulative consequences of repeatedly selecting recommended actions within one underlying MDP. Our setting requires the same recommended sets to satisfy a loss limit under every model in a specified uncertainty set. Sets that are near-optimal under one model may exceed this limit under another.

\citet{laber2014} develop set-valued dynamic treatment regimes for competing clinical outcomes. Their recommendations retain treatments with noninferior outcome vectors when a patient's preferences do not specify a single composite objective. We also leave a choice among treatments, but bound cumulative loss for a given scalar reward under every model in an uncertainty set.

Permissive controller synthesis also retains alternative actions while guaranteeing acceptable performance. \citet{drager} represent the available actions through a multi-strategy, which assigns a set of allowed actions to each decision point. Every strategy that selects from these sets must satisfy a specified reachability or expected-reward threshold, including against adversarial environmental choices in stochastic games. Their optimization objective penalizes the exclusion of actions. We share their requirement to protect every allowed selection rule. The nondeterministic-policy and classical permissive-synthesis formulations evaluate choices within one underlying model. In the stochastic-game formulation, the adversary chooses actions within a specified game rather than unknown reward or transition parameters. Our target depends on each model's optimal expected reward, so the guarantee must account for changes in both the recommendations' performance and the optimal benchmark across models.

\subsection{Robust Control}

Estimated model parameters can produce errors in the values used to compare policies. \citet{mannor2007} quantify this effect through approximations to the bias and variance of value estimates. Robust control addresses model uncertainty by maximizing a policy's lowest expected cumulative reward across a specified set of models. \citet{givan} represent model uncertainty through lower and upper bounds on rewards and transition probabilities. \citet{iyengar} and \citet{nilim} develop robust dynamic programming under rectangular uncertainty. This structure allows transition distributions at different state-action pairs to be selected independently, supporting recursive evaluation of worst-case performance. \citet{wiesemann} construct confidence regions for model parameters from observations and optimize worst-case performance over those regions. Other robust-control approaches preserve dependence across states and actions \citep{goyal,tirinzoni}. In our setting, this dependence arises across action values when an uncertain parameter affects several rewards or transition probabilities. Comparing actions under the same parameters preserves shared terms that separate bounds can lose.

\citet{steimle} optimize a weighted average of one policy's performance across several models. Worst-case and average performance objectives provide two ways to evaluate a policy under uncertainty. Either objective can favor a policy whose reward is far below the optimum under some model. Our requirement limits this loss separately in every model and covers every policy formed from the recommended actions.

\subsection{Model-Relative Regret}

We consider a regret metric that measures the expected cumulative reward lost relative to the optimum under the same model. \citet{xumannor} study uncertain rewards and prove that minimizing worst-case regret is NP-hard. \citet{ahmed} and \citet{rigter} develop methods that also accommodate uncertain transitions. While we use the same model-specific benchmark, our objective maximizes the number of recommended actions subject to a loss limit for every policy that chooses from the sets.

\citet{lutz} prepare a small collection of complete policies and select one after the model is identified. Our recommendations allow a new choice at each decision epoch while the model remains unknown. The guarantee covers all combinations of actions selected over time.

\subsection{Robust Permissive Synthesis}

\citet{huynh} provide the closest methodological antecedent, constructing set-valued recommendations under model uncertainty. They study interval MDPs with uncertain transition probabilities but fixed rewards, using mixed-integer optimization to retain as many actions as possible while satisfying a fixed reachability or expected-reward requirement. Their guarantee covers every admissible transition model and every deterministic, memoryless policy that selects from the recommended sets. We allow uncertainty in both rewards and transitions and bound loss relative to each model's own optimum. This comparison requires the optimum and the least favorable performance of the recommended sets to be evaluated under the same model. In interval MDPs, robust dynamic programming verifies fixed requirements. Finding a violation of our loss limit is NP-hard. We develop verification and construction methods that account for this dependence.

\subsection{Medical Decision Making}

Sequential decision models guide treatment for chronic conditions, including liver disease \citep{alagoz2004} and multiple sclerosis \citep{negoescu2018}. Hypertension is another chronic condition for which MDPs support treatment decisions. \citet{schell2016} approximate MDP recommendations with simpler regression rules. \citet{zargoush2018} incorporate patient risk factors into treatment decisions, while \citet{garciahypertension} develop interpretable treatment policies and measure their performance cost. \citet{steimle} account for competing models in blood pressure and cholesterol management. \citet{bonifonte2022} model the joint progression of systolic and diastolic blood pressure to identify treatment thresholds for clinical trials. Their work shows how a model of disease progression can inform the choice of interventions to evaluate. Our computational case study examines whether multiple treatment alternatives can be retained under uncertainty in future blood pressure and cholesterol while satisfying a prescribed loss limit.

The closest clinical precedent for our hypertension application is \citet{marrero2022}, who uses simulation and statistical comparisons to identify actions whose estimated performance does not differ significantly from clinical guidelines. Each alternative is evaluated at a single decision epoch, with the reference policy followed thereafter. These comparisons leave the cumulative effects of choosing alternatives at later decision epochs unresolved. Our framework evaluates these repeated choices and verifies a bound on their cumulative performance loss under every model in the uncertainty set.

\FloatBarrier
\section{Problem Setting}
\label{sec:problem-setting}

We seek a set-valued policy that preserves choices while limiting loss under every plausible model.
To this end, we first define the decision model, compliant policies, and the performance requirement.
We then specify how to measure flexibility and represent model uncertainty.
These definitions determine the certification and construction tasks in \cref{sec:solution}.
Table~\ref{tab:notation-guide} in the electronic companion collects the notation.

\subsection{Decision Model and Policies}
\label{sec:decision-model}

Let $\Sset=\{1,\ldots,n\}$ be a finite state space and $\Aset$ a finite
action space. Each state $s\in\Sset$ has a nonempty set of available actions
$\Aset(s)\subseteq\Aset$. Define
$\mathcal K=\{(s,a):s\in\Sset,\ a\in\Aset(s)\}$ and $m=|\mathcal K|$.
A model $M=(r_M,P_M)$ specifies a reward function
$r_M:\mathcal K\to\mathbb R$ and transition probabilities
$P_M:\mathcal K\times\Sset\to[0,1]$, with
$\sum_{j\in\Sset}P_M(j\mid s,a)=1$ for every $(s,a)\in\mathcal K$.
The rewards form a vector in $\mathbb R^m$, and the transition probabilities
form an $m\times n$ matrix with $P_M(\cdot\mid s,a)$ representing a column
vector.

The model $M$ is uncertain. The possible reward and transition
arrays are contained in a nonempty compact set $\Uset$, whose models share the same state and
action spaces. Compactness in this finite-dimensional space ensures that rewards are uniformly bounded. At decision epoch $t=0,1,\ldots$,
the current state and action determine the expected reward
$r_M(s,a)$ and next-state distribution $P_M(\cdot\mid s,a)$.

A deterministic stationary policy $\pi:\Sset\to\Aset$ assigns one available action to each state.
Write $\Selectors=\{\pi:\Sset\to\Aset:\pi(s)\in\Aset(s)\ \forall s\in\Sset\}$
for this finite policy class. Such a policy selects the same action whenever
the state is revisited. Every finite discounted MDP has an optimal policy of
this form \citep[Theorem~6.2.10]{puterman}.

The value of policy $\pi$ from initial state $s$ is its expected discounted
reward,
\begin{equation*}
 V_M^\pi(s)=\mathbb E_M^\pi\!\left[
 \sum_{t=0}^{\infty}\gamma^t r_M(s_t,a_t)\,\middle|\,s_0=s\right].
\end{equation*}
The expectation $\mathbb E_M^\pi$ is over transitions under model $M$ when policy $\pi$ is followed. The discount factor satisfies $0\le\gamma<1$. An optimal deterministic stationary policy $\pi_M^\star$ attains the optimal value $V_M^\star(s)=\max_{\pi\in\Selectors}V_M^\pi(s)$
from every initial state and provides a benchmark for model $M$.

A finite horizon with the same discount factor is included by adding the
decision epoch to the state and an absorbing state after the terminal
reward \citep[Chapter~4]{puterman}. Terminal states and the absorbing state each have one available action. All main results use the discounted setting above.
Appendices~\ref{app:finite-horizon} and~\ref{app:finite-cell-certificates}
give the finite-horizon representation and specialized recursions used in
the case study.

\subsection{Set-Valued Recommendations and Performance Loss}

A set-valued policy recommends a nonempty set of actions at each state as follows.

\begin{definition}[Set-valued policy]
\label{def:menu}
Let $2^{\Aset}$ denote the collection of subsets of $\Aset$.
A set-valued policy is a mapping $\menu:\Sset\to2^{\Aset}\setminus\{\varnothing\}$ such that
$\varnothing\ne\menu(s)\subseteq\Aset(s)$ for every $s\in\Sset$.
The set $\menu(s)$ consists of the recommended actions at state $s$, and the decision maker may choose any of them.
\end{definition}

We call $\menu(s)$ the menu at state $s$ and write $\MenuFamily$ for the family
of set-valued policies. A singleton policy recommends exactly one action
at every state. A policy is \emph{compliant} if it selects only recommended actions.
The compliant policies form the finite set $ \Selectors(\menu)=
 \{\pi\in\Selectors:\pi(s)\in\menu(s)\quad\forall s\in\Sset\}.$
There are $|\Selectors(\menu)|=\prod_{s\in\Sset}|\menu(s)|$ such policies.
To protect against the least favorable compliant choices, define $W_M^{\menu}(s)=\min_{\pi\in\Selectors(\menu)}V_M^\pi(s)$.
From state $s$, the largest loss relative to each model's own optimum is
\begin{equation*}
 \Reg_s(\menu)=\max_{M\in\Uset}
 \{V_M^\star(s)-W_M^{\menu}(s)\}.
\end{equation*}
Both value functions are continuous in the reward and transition arrays,
so compactness of $\Uset$ ensures that this maximum is attained
(\cref{app:proofs}). The loss is nonnegative. We collect its statewise values in
$\Reg(\menu)=(\Reg_s(\menu))_{s\in\Sset}$ and interpret vector inequalities componentwise.
Given a vector of nonnegative loss limits $\Delta\in\mathbb R_{\ge0}^n$, the
feasible family of set-valued policies is
\begin{equation}
 \Fexact=\{\menu\in\MenuFamily:
       \Reg_s(\menu)\le\Delta(s)\quad\forall s\in\Sset\}.
 \label{eq:universal-safety-main}
\end{equation}
The set $\mathbb R_{\ge0}^n$ includes vectors with zero coordinates. 
A feasible recommendation meets the loss limit for every model, starting
state, and compliant policy. The optimal benchmark can differ across
models, while the same menus must satisfy the requirement under every model in $\Uset$. Throughout the paper, a \emph{safe} policy satisfies
\cref{eq:universal-safety-main}. This term refers to the modeled expected
reward guarantee. A \emph{certificate} is a verified upper bound
on loss that establishes the guarantee when it is within $\Delta$.

\subsection{Flexibility Objective and Ordering}

Our objective is to retain as many actions as possible. More generally,
we define positive weights $\kappa_{sa}>0$ for $(s,a)\in\mathcal K$ to allow different
actions to contribute differently to the objective. Larger weights can prioritize alternatives with greater practical value. State-specific weights can also balance states with many available actions against those with few. When $\Fexact$ is
nonempty, we solve $\max_{\menu\in\Fexact}\sum_{s\in\Sset}\sum_{a\in\menu(s)}\kappa_{sa}$.
Setting every weight to one gives the total number of recommended
state-action pairs. There are finitely many set-valued policies, so a
nonempty feasible family has an optimum.

Write $\menu\preceq\menu'$ if $\menu(s)\subseteq\menu'(s)$ at every state.
Adding actions expands the class of compliant policies, so $\menu\preceq\menu'\Longrightarrow W_M^{\menu'}\le W_M^{\menu}\Longrightarrow\Reg(\menu')\ge\Reg(\menu)$.
Consequently, taking nonempty subsets of a feasible recommendation preserves
a loss guarantee. This property supports an exact search and incremental set-valued policy
construction. It also implies that a feasible set-valued policy exists if
and only if a feasible deterministic singleton policy exists (see Lemma~\ref{lem:menu-monotonicity} in the e-companion).

\subsection{Representations of Model Uncertainty}
\label{sec:uncertainty-representations}
We consider rectangular and parametric representations of uncertainty.
Rectangular sets allow separate estimation of reward and transition bounds and support robust dynamic programming \citep{givan,iyengar,nilim,wiesemann}.
Parametric families instead retain shared parameters, such as risk forecasts that affect several actions and decision epochs together.
This dependence can make a rectangular outer bound conservative, requiring bounds over a joint parameter domain.

We use an interval representation of rectangular uncertainty
in which $\rlo_{sa},\rup_{sa}\in\mathbb R$ are lower and upper reward
bounds, with $\rlo_{sa}\le\rup_{sa}$. For next state $j\in\Sset$, let
$\plo_{sa,j}$ and $\pup_{sa,j}$ be lower and upper bounds on
$P_M(j\mid s,a)$. The admissible transition vectors are
\begin{equation*}
 \Pset_{sa}=\left\{p\in\mathbb R^n:
 \sum_{j\in\Sset}p_j=1,\quad
 \plo_{sa,j}\le p_j\le\pup_{sa,j}\quad\forall j\in\Sset\right\}.
\end{equation*}
The bounds satisfy $0\le\plo_{sa,j}\le\pup_{sa,j}\le1$ and
$\sum_j\plo_{sa,j}\le1\le\sum_j\pup_{sa,j}$, ensuring that every
$\Pset_{sa}$ is nonempty. Rewards and transition vectors can also vary
independently within a state-action pair. Thus,
\begin{equation*}
 \Uset_{\rm rect}=
 \prod_{(s,a)\in\mathcal K}([\rlo_{sa},\rup_{sa}]\times\Pset_{sa}).
\end{equation*}
Every combination of admissible rewards and transition vectors defines a
model in this set. We call an MDP with this uncertainty representation an
interval MDP.

A structured parametric family has the form
$\Uset=\{M(\theta):\theta\in\Theta\}$, where
$\Theta\subseteq\mathbb R^d$ is a nonempty compact parameter domain, $d$ is the number of uncertain parameters, and
$M$ maps parameters continuously to reward and transition arrays. A single
coordinate of $\theta$ may affect several states or actions. These shared
parameters create dependence that independent interval bounds may omit.

Rectangular bounds reduce each update to an optimization over one transition row. Parametric-family certificates retain shared parameters through bounds on cells covering the domain.

\FloatBarrier
\section{Solution Methods and Theoretical Analysis}
\label{sec:solution}
\label{sec:analysis}

\begin{figure}[!htbp]
\centering
\begin{tikzpicture}[font=\fontsize{7}{8}\selectfont,>=Latex,
 box/.style={draw,rounded corners=2pt,align=center,inner sep=3pt,minimum height=13mm},
 flow/.style={->,semithick}]
\node[box,text width=13mm] (input) {\textbf{Input}\\$\Sset,\Aset(s)$\\$\Uset,\Delta$};
\node[box,text width=15mm,right=2.8mm of input] (propose) {\textbf{Propose}\\Seed or\\one-action\\extension $\menu$};
\node[box,text width=42mm,right=4mm of propose,yshift=6mm] (rect) {\textbf{Bound}\\Rectangular uncertainty\\Separated or coupled bound\\Best-margin augmentation};
\node[box,text width=42mm,below=1.2mm of rect] (family) {\textbf{Bound}\\Parametric uncertainty\\Cellwise loss on a complete cover\\Refinement; fixed-order augmentation};
\node[box,text width=15mm,right=4mm of rect,yshift=-6mm] (test) {\textbf{Test}\\Loss bound\\within $\Delta$?};
\node[box,text width=13mm,right=7mm of test] (add) {\textbf{Add}\\Accept passing\\extension};
\node[box,text width=23mm,right=2.8mm of add] (output) {\textbf{Output}\\Certified policy\\Infeasibility certificate\\Unresolved with bounds};
\draw[flow] (input) -- (propose);
\draw[flow] (propose.east) -- ++(1.2mm,0) |- (rect.west);
\draw[flow] (propose.east) -- ++(1.2mm,0) |- (family.west);
\draw[flow] (rect.east) -- ++(1.2mm,0) |- (test.west);
\draw[flow] (family.east) -- ++(1.2mm,0) |- (test.west);
\draw[flow] (test.east) -- node[above,inner sep=1pt]{yes} (add.west);
\draw[flow] (add) -- (output);
\coordinate (addition-loop) at ([yshift=8mm]add.north);
\draw[flow] (add.north) -- (addition-loop) -- node[above,inner sep=1pt]{next addition}
 (propose.north |- addition-loop) -- (propose.north);
\draw[flow] (test.north) -- node[right,align=left,pos=.5,xshift=.4mm]{drop failed\\extension} ++(0,8mm);
\draw[flow] (test.south) -- ++(0,-8mm) -| node[pos=.25,below,align=center]{no passing seed\\finite-list existence test} (output.south);
\end{tikzpicture}
\caption{Policy construction and certification steps for rectangular and parametric uncertainty.}
\label{fig:method-overview}
\end{figure}
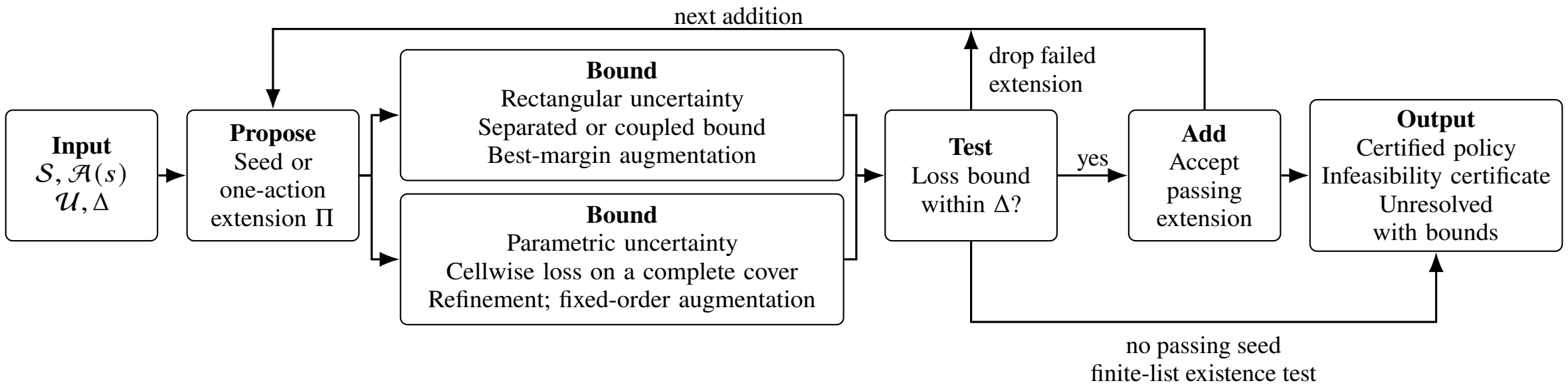

We seek a set-valued policy whose loss stays within the prescribed limit
under every plausible model. The procedure in \cref{fig:method-overview}
first bounds the loss of a candidate, then enlarges a certified candidate
through certified action additions. Exact verification explains what must
be checked and why we use upper bounds in the construction. We develop
these bounds first for independently varying rewards and transition rows,
then for families whose shared parameters link several rewards and transitions.
The two uncertainty representations use the same construction procedure but different
certificates. Our full procedure, proofs, and supporting constructions can be found in the electronic companion (\cref{app:proofs}).

\subsection{A Common Certification and Construction Procedure}
\label{sec:common-procedure}

The \textbf{Bound} step in \cref{fig:method-overview} compares optimal performance with
the least favorable compliant choices under the same model. Maximizing
this difference over plausible models gives the loss $\Reg_s(\menu)$. A certificate is a computable upper bound on
that maximum, and a bound within the loss limit proves safety.

For a model $M$ and a continuation vector
$v\in\mathbb R^n$, the optimal and worst-compliant Bellman operators are
\begin{align*}
 (\Bell_M^\star v)(s)&=\max_{a\in\Aset(s)}
 \{r_M(s,a)+\gamma P_M(\cdot\mid s,a)^\top v\},\\
 (\Bell_M^{\menu}v)(s)&=\min_{a\in\menu(s)}
 \{r_M(s,a)+\gamma P_M(\cdot\mid s,a)^\top v\}.
\end{align*}
The term $P_M(\cdot\mid s,a)^\top v$ is the expected continuation value
after action $a$ at state $s$. Both operators are monotone contractions in
the sup norm $\norminf{v}=\max_s|v(s)|$, with contraction factor $\gamma$.
The unique fixed point of $\Bell_M^\star$ is $V_M^\star$ \citep[Theorem~6.2.5]{puterman}.
Proposition~\ref{prop:worst-selector} gives the fixed point of $\Bell_M^{\menu}$.

\begin{proposition}[Worst compliant policy]
\label{prop:worst-selector}
For each model $M$ and set-valued policy $\menu$, $W_M^{\menu}$ is the
unique fixed point of $\Bell_M^{\menu}$. One deterministic stationary
policy $\sigma\in\Selectors(\menu)$ attains
$W_M^{\menu}(s)=\min_{\sigma'\in\Selectors(\menu)}V_M^{\sigma'}(s)$
simultaneously for every $s\in\Sset$.
\end{proposition}

Proposition~\ref{prop:worst-selector} reduces the worst-compliant value comparison
to one deterministic stationary policy. Restricting actions to
$\menu(s)$ and negating rewards gives an ordinary discounted model whose
optimal policy minimizes the original return \citep[Theorem~6.2.10]{puterman}. Moreover, allowing compliant choices to randomize or depend on past observations does not lower the worst-compliant value, as shown in the following remark (see \cref{app:compliant-choices}).

\begin{remark}[More general compliant choices]
\label{rem:general-compliance}
The infimum over randomized or history-dependent policies
still equals $W_M^{\menu}$.
\end{remark}
Thus, our guarantees also cover a decision maker who chooses differently
on repeated visits to the same state.

Our procedure starts from a candidate set-valued or singleton policy seed. A candidate ``passes'' when its certified loss bound is
within the limit at every state. We retain a passing seed and test policies formed
by adding one missing action to one state at a time. Best-margin augmentation
accepts the passing extension with the largest minimum margin, where a margin
is the loss limit minus the certified bound, and rescans after each acceptance.
It ends when no further extension is certified. Fixed-order augmentation sorts
missing actions once, tries each once, and ends when all have been tried.
For rectangular uncertainty, this order follows state and action indices.
For parametric uncertainty, it follows increasing largest action-loss
bound across regions covering the domain of the parameter $\theta$, breaking ties by decision epoch, state, and action index.
Best-margin augmentation is the default for rectangular uncertainty, and
fixed-order augmentation is the default for parametric uncertainty.
A fixed order requires one certificate check per missing action, while
rescanning can require a number of checks quadratic in the initial number of
missing actions. Either rule stops when a computation limit is reached and
outputs the last certified policy.
If no seed passes, the procedure ends with an infeasibility certificate or an
unresolved outcome with loss bounds as defined in \cref{sec:master-construction}.

This construction is common to both uncertainty representations.
Rectangular uncertainty provides local seeds and a fallback singleton.
Parametric seeds include singleton policies optimal for chosen models of the
family and policies that minimize the largest local loss bound at each state.
Each seed is checked separately.
Algorithm~\ref{alg:master-construction} gives the full procedure and seed list.
We now supply the bounds that \mbox{make each acceptance valid}.

\subsection{Verification and Certificates for Rectangular Uncertainty}
\label{sec:rectangular-methods}

This subsection develops the rectangular uncertainty route in
\cref{fig:method-overview}. We first describe exact verification and its
computational limits. Then, we derive value envelopes and two certificates,
and use them to obtain a seed for the common construction.
The comparison with exact verification identifies what the certificates
can prove and when they may be conservative.

\subsubsection{Verifying a Candidate Set-Valued Policy.}
We now check one set-valued policy $\menu$ under the rectangular model
$\Uset_{\rm rect}$, in which each
reward and transition probability has its own interval. The policy fails if
some admissible model $M$ at state $s$ gives
$V_M^\star(s)-W_M^{\menu}(s)>\Delta(s)$. We therefore search for a witness
of failure, consisting of an admissible model and its two value vectors. The model's rewards $r_{sa}$ and transition rows
$p_{sa}$ lie within their intervals, with $\sum_jp_{sa,j}=1$. The vectors
$v,w\in\mathbb R^n$ solve the optimal and worst-compliant Bellman equations of model $M$ for $s\in\Sset$,
\begin{equation*}
    v_s=\max_{a\in\Aset(s)}\Bigl\{r_{sa}+\gamma\sum_{j}p_{sa,j}v_j\Bigr\}, \qquad w_s=\min_{a\in\menu(s)}\Bigl\{r_{sa}+\gamma\sum_{j}p_{sa,j}w_j\Bigr\}.
\end{equation*}
For a fixed model, each equation has a unique solution, so the Bellman recursions force
$v=V_M^\star$ \citep[Theorem~6.2.5]{puterman} and $w=W_M^{\menu}$
(\cref{prop:worst-selector}). Let $\Phi_{\menu,\Delta}$ denote the resulting
finite system of polynomial equations and inequalities in $r,p,v,w$.
Its constraints impose interval bounds and stochastic transition rows, the two
Bellman equations, and $v_s-w_s>\Delta(s)$ for at least one state $s$ (see Appendix \ref{ec_subsec:exact_verification}).
Each maximum or minimum is expressed by inequalities and the requirement
that at least one attaining action gives equality.

\begin{theorem}[Exact test for one set-valued policy]
\label{thm:compact-membership}
Suppose $\Uset=\Uset_{\rm rect}$. A set-valued policy $\menu$ exceeds its
loss limit at some state, that is, $\menu\notin\Fexact$, if and only if
$\Phi_{\menu,\Delta}$ has a real solution.
\end{theorem}

Theorem~\ref{thm:compact-membership} turns the check of one policy into a
search for a counterexample. A solution names a model and a state at which
$\menu$ loses more than $\Delta$, so the policy is rejected. If no solution
exists, $\menu$ meets its loss limit under every model in $\Uset_{\rm rect}$.
The system has $(n+1)m+2n$ unknowns,
and its constraints have degree at most two, once each maximum and minimum
is written as inequalities with one equality alternative. The difficult
question is whether this polynomial system has a real solution. We take interval endpoints, the discount factor, and loss limits
to be rational numbers so that each input has a finite encoding and a
well-defined length. Under this convention, deciding violation belongs to
$\exists\mathbb R$, the class of problems reducible in polynomial time to
real feasibility of polynomial equations and inequalities
\citep{schaeferstefankovic2017}. Certifying universal safety requires
showing that no solution exists. Small-instance verification results appear
in \cref{app:exact-formulation-results}.

A natural shortcut is to check only the vertices of the transition
uncertainty sets. The following proposition shows why this shortcut can miss a violation.
\begin{proposition}[Transition vertices are insufficient]
\label{prop:nonvertex-main}
There is a three-state interval MDP and a singleton policy whose
model-relative loss at one state is strictly larger at an interior
transition row than at every vertex of that row's uncertainty set.
\end{proposition}

For Proposition~\ref{prop:nonvertex-main}, we use a discount factor $0.9$ and actions $A$ and $B$ available at states 0 and 1. State 2 has only action $C$ and a transition row that varies over a polygon with four vertices. Rewards and all other transition rows are fixed. The singleton policy chooses $B$ at states 0 and 1. Its loss at state 0 is approximately $1.63$ at an interior row, exceeding the maximum vertex loss of approximately $1.43$ (Appendix~\ref{app:nonvertex}).
Vertex-only tests miss the violation for any loss limit between the vertex maximum and the interior loss. A one-step expectation is linear in a transition row, but the
difference between two Bellman solutions can be nonlinear.

\subsubsection{Verification, Optimization, and Enumeration.}
We distinguish three computational problems: verification, weighted
optimization, and enumeration. Verification tests a proposed policy. Weighted
optimization finds a safe policy maximizing the sum of the positive action
weights $\kappa_{sa}$. Enumeration lists every safe policy that admits no
strict enlargement, which we call inclusion maximal. Such a policy may
contain fewer actions than a maximum-cardinality policy, which maximizes
the total action count. The following theorem gives the computational
limits for verification and weighted optimization.

\begin{theorem}[Exact complexity limits]
\label{thm:complexity-limits}
Suppose $\Uset=\Uset_{\rm rect}$.
\begin{enumerate}[label=(\roman*),leftmargin=2.2em]
\item Deciding whether a set-valued policy has a violating model is
NP-hard, even with single-valued rewards, any fixed rational discount factor $\gamma\in(0,1)$, and
singleton menus. Consequently, deciding whether the policy is safe is
coNP-hard under the same restrictions.
\item Finding a safe policy with the largest weighted action count is
NP-hard, even for one known model.
\end{enumerate}
\end{theorem}

Part (i) of \cref{thm:complexity-limits} shows that exact verification can
be computationally difficult even when the proposed policy allows no
choice. We prove it by encoding a graph cut in uncertain transition rows.
The singleton policy exceeds its loss limit exactly when a cut of the
required size exists. Part (ii) shows that maximizing flexibility is hard
even without model uncertainty. Our reduction assigns each optional
action the cost and profit of a knapsack item. Related knapsack reductions
establish hardness in weighted permissive synthesis \citep{drager}.
These limits motivate the certificate-based construction.

For the enumeration problem, exact downward search repeatedly applies the
verification test and provides a reference method for small problems
(\cref{app:complete-family-search}). It starts with every action enabled
and removes actions from unsafe policies while keeping the recommended
set at each state nonempty. Taking nonempty subsets preserves safety, so
further removal is unnecessary once a policy is verified safe. A family
with $N$ independent components can have exponentially many
inclusion-maximal safe policies (\cref{app:exponential-output}). Thus, exact
enumeration can require exponential output even with an inexpensive test.
The construction below seeks one certified policy instead.

\subsubsection{Attained Value Envelopes.}
Rectangular uncertainty allows us to compute value bounds by optimizing
each transition row independently. For a continuation vector $v$, define
$\phi^-_{sa}(v)=\min_{p\in\Pset_{sa}}p^\top v$ and
$\phi^+_{sa}(v)=\max_{p\in\Pset_{sa}}p^\top v$.
Each row problem is a linear optimization over a box intersected with the
probability simplex, solved by standard greedy allocation \citep{givan,nilim}.

Write
$G^-_{sa}(v)=\rlo_{sa}+\gamma\phi^-_{sa}(v)$ and
$G^+_{sa}(v)=\rup_{sa}+\gamma\phi^+_{sa}(v)$.
At each state, the operators $\Bell^+$ and $\Bell^-$ maximize $G^+_{sa}$ and $G^-_{sa}$ over $\Aset(s)$,
respectively. The operator $\Bell^-_{\menu}$ minimizes
$G^-_{sa}$ over $\menu(s)$.
These standard robust Bellman operators are monotone $\gamma$-contractions
with fixed points $V^+$, $V^-$, and $W^-_{\menu}$ \citep{iyengar,nilim}.
We call the componentwise extreme value vectors over the model set value
envelopes. These are the upper and lower optimal values, and the lower
worst-compliant value. Each extreme could
come from a different model at each state. The next theorem shows that each extreme is attained at all states simultaneously by its own admissible model.

\begin{theorem}[Attained value envelopes]
\label{thm:envelope-attainment-main}
Suppose $\Uset=\Uset_{\rm rect}$. There are admissible models $M^+$, $M^-$,
and $M^-_{\menu}$ such that, at every state $s$, $V^+(s)=V^\star_{M^+}(s)=\max_{M\in\Uset}V_M^\star(s)$, $V^-(s)=V^\star_{M^-}(s)=\min_{M\in\Uset}V_M^\star(s)$, and $W^-_{\menu}(s)=W^{\menu}_{M^-_{\menu}}(s)=\min_{M\in\Uset}W_M^{\menu}(s)$, respectively.
\end{theorem}

Each envelope is the value vector of one model and can be computed by
robust value iteration \citep{iyengar,nilim,givan}. To build
$M^+$, choose at every state-action pair the upper reward and a transition
row maximizing expected continuation under $V^+$. Rectangularity lets these
choices form one model, and $V^+$ solves its Bellman equation. Lower rewards
and minimizing rows give $M^-$ and $M^-_{\menu}$. The three models can
differ, so $V^+-W^-_{\menu}$ need not be the loss under any single model.

\subsubsection{Separated and Coupled Certificates.}
This subsection bounds performance loss with
different treatments of shared uncertainty. We obtain the separated bound
$B(\menu)=V^+-W^-_{\menu}$ because $V^+$ bounds the optimal benchmark from above
and $W^-_{\menu}$ bounds the worst compliant value from below. Although the extrema may
occur in different models, their difference bounds the loss in each common
model. The bound can be conservative because variation that cancels within
a model remains in the separate values. To retain part of that dependence,
we compare any available action $a'\in\Aset(s)$ with any compliant action
$a\in\menu(s)$ at the same state. When $a'=a$, their immediate rewards
cancel, and their common transition row acts on the continuation loss,
which we represent by a nonnegative continuation-loss vector $g\in\mathbb R_{\ge0}^n$. Each entry of $g$ bounds the loss from a possible next state.
Define
\begin{equation*}
 (\mathcal H_{\menu}g)(s)=
 \max_{\substack{a'\in\Aset(s)\\a\in\menu(s)}}
 \begin{cases}
 \gamma\phi^+_{sa}(g),&a'=a,\\
 G^+_{sa'}(V^+)-G^-_{sa}(W^-_{\menu}),&a'\ne a.
 \end{cases}
\end{equation*}
The operator $\mathcal H_{\menu}$ turns a bound $g$ on future loss into a
bound on current loss. When the optimal and compliant actions coincide, the
only loss is what passes on through the next state. Otherwise, the branch
uses constant bounds from the value envelopes.
The operator is a monotone $\gamma$-contraction. Its unique fixed point
$g^{\menu}$ is the coupled bound, whose existence and uniqueness follow from
the contraction theorem \citep[Theorem~6.2.3]{puterman}. Write
$\Fcert=\{\menu:B(\menu)\le\Delta\}$ for the policies with a separated
certificate. The next theorem bounds loss and characterizes this family's feasibility.

\begin{theorem}[Sound certificates and their feasibility floor]
\label{thm:certificate-soundness}
Suppose $\Uset=\Uset_{\rm rect}$.
\begin{enumerate}[label=(\roman*),leftmargin=2.2em]
\item For every set-valued policy $\menu$, the coupled and separated
bounds cover the loss,
\begin{equation}
 \Reg(\menu)\le g^{\menu}\le B(\menu).
 \label{eq:certificate-upper-bound}
\end{equation}
\item Some set-valued policy satisfies $B(\menu)\le\Delta$ if and only if
$\Delta\ge V^+-V^-$.
\end{enumerate}
\end{theorem}

Part (i) of \cref{thm:certificate-soundness} justifies either bound in the
acceptance test, and establishes that coupling can only improve the bound for a fixed
policy. Part (ii) gives a floor for the separated certificate. If the loss
limit is below $V^+-V^-$ at any state, no policy can pass that test.

Coupling can certify safe policies below this floor. If each state has
only one available action, every compliant policy is optimal in every
model, so its loss is zero. The separated bound can still be positive
because its two values come from different models. The coupled operator
has only same-action branches in this example, whose fixed point is zero.
More generally, strict improvement at a state requires the same action to
uniquely maximize $G^+_{sa}(V^+)$ over $\Aset(s)$ and uniquely minimize
$G^-_{sa}(W^-_{\menu})$ over $\menu(s)$. Otherwise, a distinct-action
branch attains the separated bound, so $g^{\menu}(s)=B_s(\menu)$.

\subsubsection{Seeds and Guarantees for Certified Augmentation.}
\label{sec:rectangular-construction}

We now supply a seed for the common construction in
\cref{sec:common-procedure}. For loss limits $\Delta$, set the target value
$u_\Delta=V^+-\Delta$ and retain
$\menu_{\rm loc}(s)=\{a\in\Aset(s):G^-_{sa}(u_\Delta)\ge u_\Delta(s)\}$.
The local rule admits an action if, under the least favorable rewards and transition rows, taking it
now and earning $u_\Delta$ afterward still earns $u_\Delta(s)$. If every menu
is nonempty, then $\Bell^-_{\menu_{\rm loc}}u_\Delta\ge u_\Delta$, and Bellman
iteration gives $W^-_{\menu_{\rm loc}}\ge u_\Delta$. Hence,
$B(\menu_{\rm loc})\le V^+-u_\Delta=\Delta$, so the local menus are
certified. We call this step local construction. If any local menu is empty, we replace all local menus with a singleton policy satisfying $\sigma^-(s)\in\argmax_a G^-_{sa}(V^-)$ at every state as a fallback mechanism. Its separated bound equals the floor $V^+-V^-$ under exact evaluation, so
by \cref{thm:certificate-soundness}(ii) it passes whenever any policy has a
separated certificate. Replacing all menus preserves a certificate based on the same value vector at every state. The e-companion gives the detailed procedure (\cref{alg:scalable-main}) along with the computational cost of seeds, augmentation counts, and precision requirements (\cref{app:numerical-cost}).

The following theorem presents the separated construction guarantees
under exact evaluation.
\begin{theorem}[Safety of certified augmentation]
\label{thm:scalable-safety}
If its seed checks are completed with exact evaluation, the separated
construction returns a policy
whenever some policy satisfies $B(\menu)\le\Delta$, and only then. Every
returned policy is safe. If every remaining one-action extension fails the
exact separated test, no policy in $\Fcert$ strictly contains the returned
policy.
\end{theorem}

Every returned policy is safe by \cref{eq:certificate-upper-bound}, and
exact evaluation finds a policy whenever a separated certificate exists. The local solution or verified singleton supplies the seed, and each
addition requires a fresh certificate. Adding actions can only lower $W^-_{\menu}$ and raise $B(\menu)$, so an extension that fails the exact separated test still fails after other actions are added.
If every remaining one-action extension fails, every larger policy also
fails, establishing inclusion maximality. As with enumeration, an
inclusion-maximal policy can retain fewer actions than another certified
policy. Either rule meets this condition after testing every remaining extension under exact
separated evaluation, since every rejected extension still fails after later
additions. The rules can end at different policies and differ in the number
of checks (Table~\ref{tab:scalable-complexity-main}). Every returned policy supplies a lower bound on the largest certified action count. The coupled construction uses the test $g^{\menu}\le\Delta$ in
place of $B(\menu)\le\Delta$ when it checks seeds and additions. Because
$g^{\menu}\le B(\menu)$, the coupled test accepts every policy accepted by
the separated test. It can also certify a policy even when no separated
certificate exists. The coupled bound is nondecreasing under policy inclusion,
so either rule also yields inclusion maximality for the coupled test after
evaluating every remaining extension exactly, as justified in the companion proof.

These certificates exploit independent reward and transition choices.
When a shared parameter changes several transition rows or rewards together, a
rectangular enclosure admits combinations that the parameter cannot
produce. The resulting bounds can be conservative. We next retain this
shared structure by bounding models on cells of their parameter domain.

\subsection{Continuous Families with Shared Uncertain Parameters}
\label{sec:continuous-certificates}

This subsection develops the structured parametric family route in
\cref{fig:method-overview}. We consider nonempty subsets of the parameter domain $\Theta$, which we call cells. Our approach bounds values on parameter cells, compares
actions within each cell, and covers the full domain to obtain a
certificate. The bounds use discounted Bellman fixed points on the common
state space.

Recall that $\Theta\subseteq\mathbb R^d$ is the compact domain of the
$d$ uncertain parameters. The models share finite state and action sets and discount factor $0\le\gamma<1$.
The reward $r_\theta(s,a)$ and transition row
$P_\theta(\cdot\mid s,a)$ depend continuously on $\theta$.
The values $V^\star_\theta(s)$ and $W^{\menu}_\theta(s)$ are its optimal and
worst-compliant discounted returns. Shared parameters may create dependence
across states and actions.
We bound models within regions of the parameter domain and use those bounds
to control cumulative loss. With valid local enclosures, this construction
also applies to nonrectangular families. Cells may be parameter boxes or
regions within a box on which a continuous scalar score lies in a closed
interval. The latter can tighten enclosures when transitions depend on the
parameters mainly through that score.

\subsubsection{Cellwise Value and Action-Loss Bounds.}
\label{sec:cellwise-bounds}

Within each cell, we bound action values and cumulative
losses. For a compact cell
$C\subseteq\Theta$, let the finite interval $[\rlo^C_{sa},\rup^C_{sa}]$
and the nonempty compact set of stochastic transition rows $\Pset^C_{sa}$ enclose
the reward and transition row for every $\theta\in C$. This set serves the
same role as $\Pset_{sa}$ in Section~\ref{sec:uncertainty-representations},
now enclosing the transition rows on one cell.
For a continuation vector $v$, define
$\underline\phi^C_{sa}(v)=\min_{p\in\Pset^C_{sa}}p^\top v$ and
$\overline\phi^C_{sa}(v)=\max_{p\in\Pset^C_{sa}}p^\top v$.
The lower and upper optimal-value bounds are the unique solutions of
\begin{equation*}
    \underline V^C(s)=\max_{a\in\Aset(s)}
 \{\rlo^C_{sa}+\gamma\underline\phi^C_{sa}(\underline V^C)\} \quad \text{and} \quad \overline V^C(s)=\max_{a\in\Aset(s)}
 \{\rup^C_{sa}+\gamma\overline\phi^C_{sa}(\overline V^C)\}.
\end{equation*}
These monotone $\gamma$-contractions give
$\underline V^C\le V^\star_\theta\le\overline V^C$ throughout the cell.
The corresponding action-value bounds are
$\underline Q^C_{sa}=\rlo^C_{sa}+\gamma\underline\phi^C_{sa}(\underline V^C)$
and
$\overline Q^C_{sa}=\rup^C_{sa}+\gamma\overline\phi^C_{sa}(\overline V^C)$.

We build three certificates from these enclosures. The raw certificate uses
the cellwise separated bound. The advantage certificate bounds each
action's one-step loss, and the joint certificate bounds the difference
between two actions directly. The raw certificate computes
\begin{equation*}
 \underline W^{C,\menu}(s)=\min_{a\in\menu(s)}
 \{\rlo^C_{sa}+\gamma\underline\phi^C_{sa}(\underline W^{C,\menu})\},
\end{equation*}
and uses
$B^{C,\mathrm{raw}}=\overline V^C-\underline W^{C,\menu}$.
The defining operator is again a monotone $\gamma$-contraction.
The ordering $\underline W^{C,\menu}\le\underline V^C\le\overline V^C$
makes this bound nonnegative, including when rewards are negative.

We first measure the loss from one action followed by optimal choices. Define
\begin{equation*}
 \ell_\theta(s,a)=V^\star_\theta(s)-r_\theta(s,a)
 -\gamma P_\theta(\cdot\mid s,a)^\top V^\star_\theta.
\end{equation*}
The nonnegative loss $\ell_\theta(s,a)$ is the cost of taking action $a$
now and acting optimally thereafter under model $\theta$, or the negative
of its advantage. The advantage certificate uses
\begin{equation*}
 \overline\ell^{C,\mathrm{adv}}_{sa}
 =\max\{0,\overline V^C(s)-\underline Q^C_{sa}\}.
\end{equation*}
Thus $0\le\ell_\theta(s,a)\le\overline\ell^{C,\mathrm{adv}}_{sa}$ for every
$\theta\in C$. Bounding the two values separately can be conservative,
because their extrema may occur at different parameter values.
The total loss $\rho_\theta=V^\star_\theta-W^{\menu}_\theta$
satisfies
\begin{equation*}
 \rho_\theta(s)=\max_{a\in\menu(s)}
 \{\ell_\theta(s,a)+\gamma P_\theta(\cdot\mid s,a)^\top\rho_\theta\}.
\end{equation*}
The worst compliant choice pays its one-step loss now and the loss from the
next state on. Upper bounds on $\ell$ therefore confine the total loss through
the same fixed-point comparison.

For the joint certificate, we directly bound the action-value difference
between a rival action $a'$ and candidate action $a$ under the same parameter
values. Shared terms can therefore cancel before we bound the difference.
This difference, called a contrast, is
\begin{equation}
 D_\theta(s;a',a)=r_\theta(s,a')-r_\theta(s,a)
 +\gamma\bigl(P_\theta(\cdot\mid s,a')-P_\theta(\cdot\mid s,a)\bigr)^\top
 V^\star_\theta.
 \label{eq:clinical-joint-contrast}
\end{equation}
Let $\overline D^C_{s,a',a}$ bound this entire difference for every
$\theta\in C$, with $\overline D^C_{s,a,a}=0$. A valid choice maximizes
the expression over a single shared $\theta\in C$ and a continuation
vector in $[\underline V^C,\overline V^C]$.
Compactness and continuity ensure this maximum exists.
Although this bound permits extra
continuation vectors, it retains the dependence between the two actions'
rewards and transitions. We exclude any rival whose upper
action value is below another action's lower value because it cannot be
optimal. The rival set is
$R^C_s=\{a'\in\Aset(s):\overline Q^C_{sa'}\ge\max_{a''}\underline Q^C_{sa''}\}$.
For every model in the cell, it contains every optimal action and may
contain actions that are never optimal. Since
$\ell_\theta(s,a)=\max_{a'}D_\theta(s;a',a)$ is attained at an optimal
action, $\max_{a'\in R^C_s}\overline D^C_{s,a',a}$ bounds it. Both this quantity and the advantage bound cover the same one-step loss.
Their minimum is therefore a valid upper bound, giving the joint action-loss bound
\begin{equation*}
 \overline\ell^{C,\mathrm{joint}}_{sa}
 =\min\!\left\{\overline\ell^{C,\mathrm{adv}}_{sa},
       \max\!\left(0,\max_{a'\in R^C_s}\overline D^C_{s,a',a}\right)\right\}.
\end{equation*}
By bounding differences before maximizing, we can cancel common reward
and transition terms.

To account for all subsequent choices, we combine the one-step loss bounds
with continuation losses. For $\chi\in\{\mathrm{adv},\mathrm{joint}\}$, define
the loss operator on nonnegative vectors by
\begin{equation}
 (\mathcal H^{C,\chi}_{\menu}v)(s)=\max_{a\in\menu(s)}
 \{\overline\ell^{C,\chi}_{sa}+\gamma\overline\phi^C_{sa}(v)\},
 \qquad B^{C,\chi}=\mathcal H^{C,\chi}_{\menu}B^{C,\chi}.
 \label{eq:clinical-loss-recursion}
\end{equation}
Each loss operator is a monotone $\gamma$-contraction and has a unique
nonnegative fixed point. Maximizing the sum of current and continuation
loss bounds covers every allowed action. The following proposition establishes the cellwise
guarantee and compares the three bounds.

\begin{proposition}[Cellwise loss bounds]
\label{prop:continuous-cell-safety}
Suppose the cell enclosures contain all models $M(\theta)$, $\theta\in C$,
and the contrast bounds of \cref{eq:clinical-joint-contrast} are valid.
For every nonempty set-valued policy and every $\theta\in C$,
$0\le V^\star_\theta(s)-W^{\menu}_\theta(s)\le B^{C,\chi}(s)$
for $\chi\in\{\mathrm{raw},\mathrm{adv},\mathrm{joint}\}$, with
$B^{C,\mathrm{raw}}(s)\le B^{C,\mathrm{adv}}(s)$ and
$B^{C,\mathrm{joint}}(s)\le B^{C,\mathrm{adv}}(s)$.
\end{proposition}

Each bound can certify the loss limit on a cell. For the same cell and policy,
both raw and joint bounds are no larger than the advantage bound.
Neither the raw nor the joint bound is always the smaller one, as examples
accompanying the proof show (\cref{app:continuous-family-proofs}). Substituting the
local bounds into the loss operators and comparing their fixed points proves
the cumulative guarantee.

\subsubsection{Complete Covers and Certified Search.}
\label{sec:clinical-cover-search}
Covering the entire parameter domain extends the cellwise guarantee to
every model in the family. Let $\mathcal C$ be a finite collection of
nonempty compact cells whose union is all of $\Theta$, including its boundary.
Set $B^{\mathcal C,\chi}(s)=\max_{C\in\mathcal C}B^{C,\chi}(s)$.
A policy passes when this bound is at most the loss limit at every
state.

\begin{proposition}[Safety on a complete parameter cover]
\label{prop:clinical-cover-safety}
For a complete cover with valid cell bounds,
$\sup_{\theta\in\Theta}\{V^\star_\theta(s)-W^{\menu}_\theta(s)\}
\le B^{\mathcal C,\chi}(s)$.
\end{proposition}

Passing the complete-cover test guarantees the loss limit under every
model, since each belongs to a covered cell. Complete coverage by score
regions gives the same guarantee (\cref{lem:risk-score-cover}).
The common construction in
\cref{sec:common-procedure} uses this bound for seeds and action additions.
A policy optimal for one model is a candidate singleton seed, but it must
pass the complete-cover test before the construction retains it.

A loose enclosure can leave a cell's bound above the loss limit even when
the candidate is safe. Subdividing the cell can tighten the bound.
For a fixed candidate, refinement starts from a complete cover and splits
selected cells whose bounds exceed the limit. Each selected parent is
replaced by children contained in it whose union equals the parent.
Each child receives the componentwise minimum of its computed bound and
the parent's bound for that same candidate. Both bounds are valid on the
child, so repeated refinement preserves coverage and cannot increase the
complete-cover bound for the fixed candidate. Unchanged \mbox{cells retain their bounds}.

\subsection{Certified Policies, Infeasibility, and Unresolved Cases}
\label{sec:master-construction}

Our construction outputs a certified policy whenever a seed passes, maintaining
certification through each accepted action addition. When the seed search
has not produced a certificate, the question remains whether any policy
satisfies the loss limit at every state under every model in the family.
Finite lists of models can propose further candidates or prove that no
such policy exists.

An infeasibility certificate consists of models from the family and a proof
that every deterministic policy exceeds the loss limit at some state under
at least one listed model. This proof rules out a nonempty safe set-valued policy,
because such a policy contains a safe singleton by
\cref{lem:menu-monotonicity}, and safety for the family implies safety
for every member of the list. Finite lists are a proof device and also
supply candidate seeds.

An unresolved outcome means that neither a policy certificate nor an
infeasibility certificate was obtained. For a common scalar loss limit,
write $d^\star=\min_{\sigma\in\Selectors}\max_s\Reg_s(\{\sigma\})$ for the
smallest achievable worst-case loss. Finite-list analysis supplies a lower
bound $\underline d\le d^\star$, and the smallest complete-cover bound
among the policies found supplies an upper bound $d^\star\le\overline d$.
These bounds describe how close $d^\star$ is to the loss limit.

The rectangular construction and the parametric cover procedure are given
in Algorithms~\ref{alg:scalable-main} and~\ref{alg:family-search}.
The infeasibility arguments and controlled examples appear in
\cref{app:clinical-existence,app:synthetic-controls}.

\FloatBarrier
\section{Computational Experiments}
\label{sec:numerical}
\label{sec:computational}

This section examines how certification affects choice and computational effort in models with up to 144 states and eight actions per state. We compare the rectangular certificates of \cref{sec:rectangular-methods} with exact answers, then test when a separated certificate exists and how widening uncertainty reduces choice. Further experiments show that ignoring shared parameters can overstate loss, while comparisons made one epoch at a time can understate it (\cref{sec:shared-uncertainty-controls}). In a shared-reward example, retaining dependence reduces the bound from about $0.21$ to the true loss of $0.01$. The parametric-family certificates of \cref{sec:continuous-certificates} recover exact maximum action counts in ten small one-parameter families (\cref{app:parametric-reference}). All experiments in \cref{sec:computational,sec:clinical-study-design} use unit action weights in the flexibility objective. The e-companion gives additional results in \cref{app:extra-results}.

\subsection{Reference Comparisons}
Small models allow exhaustive enumeration, which reveals how close each construction comes to the largest feasible action count. Evaluating every policy under each certificate then separates shortfalls due to certificate conservatism from those due to the search. The exact maximum is the largest number of retained state-action pairs among policies satisfying the loss limit.

We draw nominal rewards uniformly from $[0,1]$ and transition rows uniformly from the probability simplex. Each nominal reward or transition coordinate $c$ has interval $[(1-\eta)c,(1-\eta)c+\eta]$, with transition rows restricted to the simplex. Local construction admits actions using the conservative continuation values of \cref{sec:rectangular-construction}. Best-margin augmentation repeatedly adds the action leaving the largest margin under the separated bound. Coupled census tests every policy and finds the largest count its certificate admits.

\begin{figure}[!htbp]
\centering
\includegraphics[width=\linewidth]{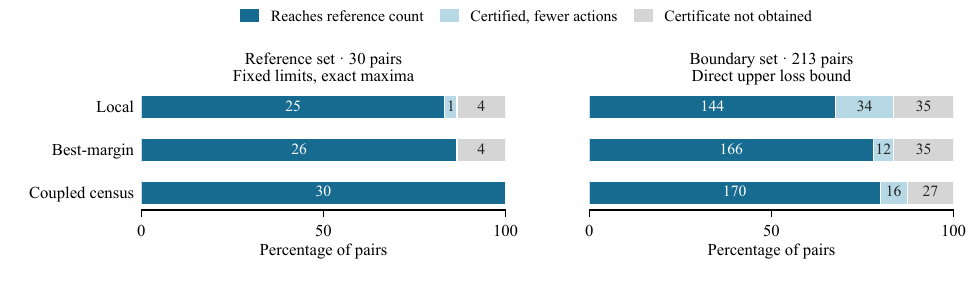}
\caption{Action counts from local construction, best-margin augmentation, and coupled census. The left panel compares them with exact maxima at fixed loss limits. The right compares them with the reference counts at loss limits where the direct upper loss bound first permits more actions. Bar segments distinguish reaching the reference count, certifying fewer actions, and obtaining no certificate. Bar labels count instance--limit pairs.}
\label{fig:exact-scalable-capacity}
\label{fig:compact-runtime}
\end{figure}

\paragraph{Small reference problems.}
We generate two nominal models at each of five sizes, giving ten interval families. Each model has one state with two, three, or four actions, or two states with two or three actions per state. We use uncertainty fraction $\eta=0.25$ and discount factor $0.5$. Each family is tested at loss limits $0.5$, $1$, and $1.5$. An \emph{instance--limit pair} combines a family with a loss limit. The \emph{reference set} contains 30 such pairs, all admitting a feasible policy. Local construction certifies 26 and attains the exact maximum in 25. Best-margin augmentation reaches the maximum in all 26 certified cases. No policy can be certified by the separated bound in the other four. Coupled census reaches the exact maximum in all 30 pairs (Figure~\ref{fig:exact-scalable-capacity}, left). Each selected policy passes the exact test of \cref{thm:compact-membership}. Augmentation therefore reaches the exact maximum wherever a separated certificate exists in this set.

\paragraph{Loss-limit boundaries.}
For each policy, branch and bound encloses its worst-case loss in an interval. We use the upper endpoint, the \emph{direct upper loss bound}, as the reference for the certificates in \cref{sec:rectangular-methods}. We test limits where this bound first admits a larger action count, so conservatism can cost an action. We enumerate policies in 68 families with up to three states and three actions per state. Discount factors are $0$, $0.5$, or $0.75$. Reward and transition uncertainty fractions are $0$, $0.5$, or $1$. For each target action count, the tested limit is the smallest direct upper loss bound among enumerated policies reaching that count. The \emph{boundary set} contains 213 instance--limit pairs. The reference count at each pair is the largest admitted by exhaustive evaluation.

Local construction reaches the reference at 144 pairs, best-margin augmentation at 166, and coupled census at 170 (Figure~\ref{fig:exact-scalable-capacity}, right). Where a separated certificate exists, best-margin augmentation attains the largest action count that certificate permits. At the other 35 pairs, it returns no policy. Thus, the shortfall comes from the bound in these cases. Coupling leaves 27 pairs without a certificate. It recovers some lost actions, consistent with the tighter bound in Theorem~\ref{thm:certificate-soundness}(i).

To measure the cost of local construction as models grow, we extend the reference design to 128 states at loss limit $1$. With four actions per state, local construction takes a median 0.44 seconds (Figure~\ref{fig:reference-construction-time}).

\subsection{Certificate Floors and Uncertainty Width}
Our theory establishes when a separated certificate exists. We test that claim in synthetic interval models, examine what coupling adds beyond it, and track flexibility in a sepsis simulator as uncertainty widens.

\paragraph{Synthetic interval models.}
Our synthetic grid uses eight state counts, doubling from one to 128. Each model has two or four actions per state. Reward and transition half-widths independently take values $0$, $0.05$, or $0.25$. Three replications per combination give 432 cases. Nominal rewards are uniform on $[-1,2]$, and transition rows are uniform on the simplex. Reward intervals extend by the half-width in each direction. Transition intervals are clipped to $[0,1]$ and intersected with the simplex. We use discount factor $0.9$ and loss limit $0.5$. We use fixed-order augmentation in state and action order. Every certified policy \mbox{passes an exact-rational check}.

The separated floor satisfies $V^+-V^-\le0.5$ at every state in exactly 64 cases. Separated augmentation certifies those 64 and no others, as Theorem~\ref{thm:certificate-soundness}(ii) predicts. The coupled construction certifies 23 further cases with separated floors up to about three times the loss limit. The coupled comparison cancels a common reward for identical actions and certifies 20 small cases with uncertain rewards. Of the 345 remaining cases, we prove that 244 admit no nonempty set-valued policy meeting the loss limit.

All 48 cases with zero uncertainty are certified. With positive uncertainty, none of the constructions certifies a case with eight or more states. These infeasibility proofs show that independent intervals can preclude a common near-optimal policy. \Cref{sec:clinical-study-design} uses shared parameters to preserve dependence.

\paragraph{Sepsis uncertainty widths.}
We consider a 144-state projection of a synthetic sepsis simulator with eight actions and 1,152 state-action pairs. States describe heart rate, blood pressure, oxygen level, and treatment status, and actions combine antibiotics, ventilation, and vasopressors (\citealt{oberst}). Ten models represent the combinations of diabetes status and current glucose category omitted from the state description. At full width $\lambda=1$, reward and transition intervals span all ten models. Width $\lambda$ scales their endpoints toward the average model, which remains at $\lambda=0$.

We choose a loss limit that allows some choice at full uncertainty and then examine how narrowing the intervals expands it. At full width, local construction first recommends an action in every state at a limit of 89.11 and retains every action at 186.22. We place the limit 8\% of the way between these thresholds, at 96.88 on the simulator's reward scale. At a discount factor of 0.99 and loss limit 96.88, fixed-order augmentation keeps 1,152 pairs at width zero and 950 at full width, compared with 1,020 and 712 for \mbox{local construction (Figure~\ref{fig:sepsis-width})}.

A separate experiment uses widths $0.25$, $0.5$, and $1$, with intervals contracted toward one constituent model. At each width, we test loss limits $0.1$, $0.5$, and $1$. No nonempty set-valued policy meets these limits in any of the nine cases (\cref{app:synthetic-controls}). At full width, a separated certificate requires a limit above 26.41, yet one policy loses less than 4.45 on the ten models themselves (\cref{app:sepsis-width-detail}).

\FloatBarrier
\begingroup
\let\preprintfigure\figure
\renewcommand\figure[1][t]{%
  \ifnum\value{figure}<4
    \preprintfigure[p]%
  \else
    \preprintfigure[t]%
  \fi}
\section{Hypertension Treatment Case Study}
\label{sec:clinical-study-design}

Our case study asks whether uncertainty about future cardiovascular risk leaves room for treatment preferences within a QALY loss limit. In our primary experiment, we vary only forecasts of untreated risk factors across models. We describe the patient models and experimental design before examining certification, treatment flexibility across patient groups, and the consequences of choosing among recommended treatments. Our sensitivity analyses vary the loss limit, uncertainty width, and modeling assumptions.

\subsection{Patient Models and Uncertainty in Future Risk Factors}

We adapt the hypertension model of \citet{garciahypertension}, using its cohort, forecasts, treatment responses, and health-state conventions. \citet{marrero2024wsc} use a similar version of the data and model. Uncertain forecasts share parameters across years and actions, which motivates a parametric family and joint certificate. We refer to a patient whose admissible action sets contain at least two actions at some year and living health state as \emph{choice-capable}. Table~\ref{tab:clinical-settings} gives the model inputs and their sources.

Our cohort comprises 4,590 patients from the 2009--2016 National Health and Nutrition Examination Survey, representing 66.5 million adults under survey weights \citep{nhanes}. The extraction includes non-Hispanic Black and White adults aged 40--60 at baseline and excludes patients reporting a previous myocardial infarction or stroke (\cref{app:clinical-survey-weights}). We forecast each patient's risk factors over ten annual decision epochs (\cref{app:clinical-factor-bands}). Conditional on that trajectory, the model has ten health states at each year per patient. Six describe living conditions, distinguishing no cardiovascular history, prior coronary heart disease, prior stroke, both histories, and survival of a current coronary or stroke event. The other four distinguish deaths due to a coronary, stroke, or non-ASCVD event and an absorbing death state.

Each action specifies a combination of antihypertensive drug classes. A set-valued treatment plan recommends treatments for each health state and year. We consider up to 196 combinations, including no treatment, before admissibility restrictions \citep{garciahypertension}. Medication count measures treatment intensity. Combinations with the same count can differ in blood pressure (BP) reductions and medication burdens, estimated from trials and treatment models \citep{law2003,law2009,schell2016}. The revised pooled cohort equation determines untreated cardiovascular risk from each forecast \citep{yadlowsky2018}. The transition model applies treatment effects, event-history adjustments, and competing mortality (\cref{app:clinical-common-feasibility}). Rewards are QALYs, reduced by medication burden, with an annual discount factor of 0.97. Terminal continuation accounts for expected QALYs beyond the 10-year decision horizon, conditional on the final health state (\cref{app:clinical-risk-bridge}). This formulation allows us to use the finite-horizon recursions in \cref{app:finite-cell-certificates}.

Since untreated cardiovascular risk depends on systolic BP, total cholesterol (TC), and high-density lipoprotein (HDL) cholesterol, we consider their forecasts as uncertain. We base each factor's forecast band on the 2.5th and 97.5th percentiles of its assumed regression-based forecast distribution (\cref{app:clinical-factor-bands}). A parameter $z_j\in[-1,1]$ moves that factor's future path within the band. Values $z_j=-1$, $0$, and $1$ give the lower band end, forecast, and upper band end, respectively. The width $w$ scales the logarithmic distance from the forecast, with $w=1$ giving the original band. Logarithmic interpolation keeps values positive. Each model keeps a factor on the same side of its forecast across future years. Thus, $w=0$ gives the nominal model, which takes each forecast as exact, while $w=0.5$ halves the distance and $w=2$ doubles it. In year ten, the median systolic BP band extends about 44 mmHg below and above the forecast among choice-capable adults. Wider families require a plan to accommodate more forecasts and can reduce choices satisfying the guarantee. For patient $i$, the family is
\[
\Uset_i^{\rm clin}(w)=\{M_i(z;w):z=(z_{\rm SBP},z_{\rm TC},z_{\rm HDL})\in[-1,1]^3\}.
\]
Baseline values remain fixed, and each coordinate is shared across subsequent years, states, and actions (\cref{eq:clinical-factor-scale,eq:clinical-factor-map}). A value of $z$ fixes how the three risk factors evolve, while events and deaths remain random as in the source model. We do not know the correct $z$ and do not assign probabilities to it.

Below 120 mmHg systolic or 55 mmHg diastolic, the benefit of further BP reduction is not well established. As a result, we exclude treatments predicted to lower BP beyond either level \citep{garciahypertension}. The no-treatment action is admissible when nominal systolic BP is at most 150 mmHg and diastolic BP at most 90 mmHg. A fallback allows no treatment when every action fails the rules. Under these restrictions, 34.5 million adults have only one admissible action, no treatment, at every living health state and year. The primary certification analysis considers the remaining choice-capable population of 32.0 million adults.

The least systolic-BP-reducing drug in our model, an ACE inhibitor, requires nominal systolic BP of approximately 126 mmHg or above to satisfy the floor. Diastolic BP restrictions can further exclude treatment. Some choice-capable patients have treatment alternatives from the first year, while others gain them only later. This input-based classification precedes our treatment plan optimization.

We fix treatment responses and medication burdens at nominal values to isolate uncertainty in future risk factors. Treatment admissibility uses the nominal BP predictions, with the BP floors enforced at those predictions. Our certificate bounds the QALY loss on these action sets across the uncertain family. Details of the range checks and the nonempty-action fallback appear in \cref{app:clinical-risk-bridge,app:clinical-common-feasibility}.

\subsection{Experimental Design and Evaluation Measures}

We ask whether physicians can offer treatment choices under forecast uncertainty and how many adults retain those choices. We then identify which patients retain flexibility and what preferences for fewer medications or greater continuity cost.

The primary analysis uses width $w=1$ for the 32.0 million choice-capable adults. We set a clinically informed loss limit of $\delta=0.03$ QALYs, approximately eleven quality-adjusted days, at every living health state and year. \citet{horsman2003} identify a mean Health Utilities Index difference of 0.03 as clinically important, with reported minimal important differences of 0.02--0.04. We interpret that utility difference sustained for one year as a cumulative QALY loss limit. A certified plan recommends treatments whose selections stay within this limit of each forecast model's own optimum, however the choices change over time. Our construction uses the seeds and fixed-order augmentation of Section~\ref{sec:common-procedure}. We bound loss over regions of $z$ covering every model in the family and check certificates independently (\cref{sec:clinical-cover-search,app:clinical-risk-cover}). The e-companion lists computation limits and stopping reasons in Table~\ref{tab:clinical-computation-settings}.

We classify patients by certification outcome (\cref{sec:master-construction}). A \emph{certified flexible plan} offers multiple treatments at some living health state and year, while a \emph{certified singleton} offers one everywhere. An \emph{infeasible} patient admits no plan meeting the loss limit across the family. For an \emph{unresolved} patient, neither a qualifying plan nor a proof of infeasibility is established. The share with certified flexible plans is a lower bound on the share admitting flexibility (\cref{app:clinical-results}).

We compare certified plans with the nominal-model plan to measure the cost of ignoring forecast uncertainty and with the guidelines to assess choices against guideline care. The \emph{nominal-model plan} uses the same construction, admissible treatments, and loss limit at width zero, controlling loss under the nominal forecast alone (\cref{app:clinical-results}). At $w=1$, it is \emph{rejected} when an allowed selection loses over 0.03 QALYs under some model at a living health state and year, and certified otherwise. As a clinical reference, we evaluate the 2017 ACC/AHA medication-count guidelines \citep{wheltonguideline2017}, mapped to the nearest admissible medication count (\cref{app:decision-outcome-design}). The guidelines set each year's medication count from BP and cardiovascular risk, and the model uses the lowest-burden drug combination with that count.

We measure both the amount of choice and how widely it is available. For each patient, we average the number of recommended treatments over the 60 combinations of living health state and year and count how many offer multiple treatments. Population shares and summaries use the survey weights \citep{nhanesweighting} (\cref{app:clinical-survey-weights}).

A selection rule chooses a recommended action according to a preference. The \emph{best-QALY rule} maximizes expected QALYs and serves as the reference rule. Other rules choose the fewest current medications, least medication burden, fewest regimen changes, or median medication count. Class-avoidance rules maximize nominal QALYs after removing one drug class from all recommended sets, provided every living health state and year retains an option. Best and worst selections within each plan bound the QALYs any rule can achieve. We also report selection losses in \cref{sec:clinical-selection-results} against the unrestricted nominal optimum, the best plan under the nominal forecast over all admissible treatments. A regimen change is a switch to a different drug combination between consecutive years while alive. This outcome measures continuity over the ten years. Expected regimen changes average that count over possible health trajectories during the 10-year horizon. Outcome comparisons are nominal-model predictions from an initial healthy state, whereas the certificate covers every model and every compliant selection. These comparisons concern adults with certified flexible plans. QALYs include terminal continuation (\cref{app:decision-outcome-design}).

Sensitivity analyses use 244 patients selected to span model inputs (\cref{app:clinical-panel-design}). Higher loss limits test how much more choice a larger limit allows. Forecast widths test how much uncertainty those choices withstand (\cref{app:clinical-width-results}). Dose variants ask whether more medication combinations increase flexibility. Shorter horizons test how accumulating loss over time constrains choice (\cref{app:dose-horizon-design}). Post-event and terminal-reward variants test how health after illness and beyond year ten affects certified choices (\cref{app:clinical-structure-results}). The responsive analysis varies treatment responses across models to test how requiring admissibility under every model narrows choice (\cref{app:clinical-physical-results}). We test whether sets chosen by statistical comparisons at each decision epoch \citep{marrero2022} keep cumulative loss within the limit (\cref{app:clinical-statistical-results}).

\subsection{Numerical Results}

We report certification, flexibility across patients, and outcomes within certified plans.

\subsubsection{Certification and Retained Treatment Flexibility}
\label{sec:clinical-certification-results}

At $\delta=0.03$ and $w=1$, certified flexible treatment plans cover 93.6\% of the choice-capable population, or 30.0 million adults. For these patients, we tried adding every missing treatment to the plan. In contrast, across the forecast family, nominal-model plans are certified for 0.2\% of adults and rejected for 99.8\% (Figure~\ref{fig:clinical-search-outcomes}). To assess the size of this loss, we evaluate all 27 combinations of the three factors' low, central, and high values. Among rejected plans, the largest loss across these models from the healthy initial state has a median of 0.30 QALYs, ten times the limit. Its interquartile range is 0.17--0.55 QALYs. Accounting for forecast uncertainty therefore changes which choices physicians can offer under the guarantee.

Certified singletons account for 0.7\% of adults. Infeasible and unresolved patients account for 3.4\% and 2.2\%, respectively. Adults with certified flexible plans have a median of 3.45 recommended treatments, averaged across all 60 combinations. Multiple treatments are available in a median of 35 combinations. Thus a physician could offer alternatives at many visits while keeping cumulative modeled loss within the limit, whichever forecast in the family holds.

Unresolved patients represent 0.70 million adults. Within the computation limits, we found neither a qualifying plan nor a proof of infeasibility. The 47 unresolved patients have lower bounds from 0.0247 to 0.0300 QALYs and upper bounds from 0.0300 to 0.0345 on the smallest achievable worst-case loss. Each patient has a plan certified at 0.035 QALYs, but whether any plan meets 0.03 remains unknown (\cref{app:unresolved-existence-bounds}). More precise bounds could decide whether these patients admit a plan at the primary limit.

Every plan certified at 0.03 QALYs remains valid at the larger loss limits, establishing flexibility for at least 93.6\% of adults at each. At 0.05 QALYs, bounds establish some treatment plan for at least 97.6\% of adults. At the primary loss limit, all 244 panel patients have certified flexible plans at widths zero and one half. At widths one and two, at least 234 and 155 patients do, respectively (\cref{app:clinical-width-results}). Wider forecast ranges make the guarantee more demanding.

\begin{figure}[t]
\centering
\includegraphics[width=\linewidth]{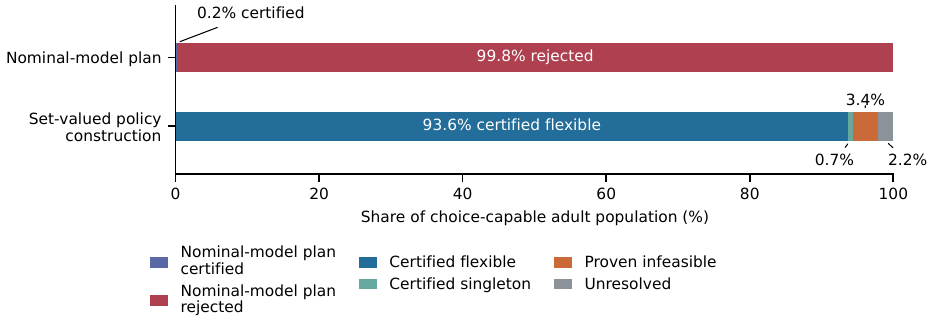}
\caption{Survey-weighted outcome shares among the 32.0 million choice-capable adults at $\delta=0.03$ and $w=1$. The top bar tests the plan built from the nominal forecast alone. Rejection means an allowed selection loses over 0.03 QALYs relative to a width-one model's optimum at some living health state and year. The lower bar distinguishes certified flexible plans, certified singletons, infeasible patients, and unresolved patients.}
\label{fig:clinical-search-outcomes}
\end{figure}

\subsubsection{Treatment Flexibility Across Patient Characteristics}

Physicians can retain treatment choices even for adults with high BP or cardiovascular risk. The adult shares with certified flexible plans are 99.4\%, 94.7\%, and 79.6\% across baseline systolic BP groups 120--129, 130--139, and 140--159 mmHg. Among the 1.49 million adults with systolic BP at least 160 mmHg, 89.3\% retain certified treatment alternatives (Figure~\ref{fig:clinical-flexibility-profile}A). Across nominal annual risk groups below 0.5\%, 0.5--1\%, 1--2\%, and at least 2\%, the shares are 92.4\%, 96.4\%, 99.2\%, and 100.0\%, with 0.42 million adults in the highest-risk group (Figure~\ref{fig:clinical-flexibility-profile}B). Absolute benefits from BP-lowering treatment increase with cardiovascular risk, making choice relevant to patients who stand to benefit substantially \citep{sundstrom2014}.

Choice is more constrained when moderately high systolic BP accompanies low nominal risk. Adults with systolic BP 140--159 mmHg and risk below 0.5\% number 3.34 million. Of these adults, 64.1\% have certified flexible plans. Infeasible and unresolved patients account for 19.4\% and 13.9\% of the group, respectively. At the same systolic BP with risk at least 0.5\%, 97.2\% of 2.93 million adults have certified flexibility. Below systolic BP 140 mmHg, 97.5\% of 24.26 million adults have certified flexibility. At 160 mmHg or above, 22.6\% of adults are low-risk and most are higher-risk. BP floors keep every 120--129 mmHg adult below twenty admissible treatments on average.

Among adults classified as infeasible, 94.8\% have annual risk below 0.5\%, 91.9\% average more than twenty admissible treatments across living health states and years, and 71.1\% are younger than fifty, with a median age of 46 compared with the median age of 52 among adults with certified flexibility. Low-risk forecasts can favor few medications, while high-risk forecasts can favor much more intensive treatment. For adults classified as infeasible, no single plan accommodates those differences while staying within the 0.03-QALY loss limit under every forecast.

Flexible shares across groups averaging 1--5, more than 5--20, more than 20--100, and more than 100 admissible treatments per living health state and year are 99.2\%, 98.5\%, 82.6\%, and 87.1\% (Figure~\ref{fig:clinical-flexibility-profile}C).

Some patients gain choice as their risk factors evolve. Among adults with certified flexible plans, 73.3\% have multiple admissible treatments in the healthy state in the first year. Certification preserves that choice for 98.9\% of this group. The other 26.7\% begin with one admissible treatment and have choices later. A treatment plan can therefore support preferences at future visits even when only one option is available initially.

\begin{figure}[t]
\centering
\includegraphics[width=\linewidth]{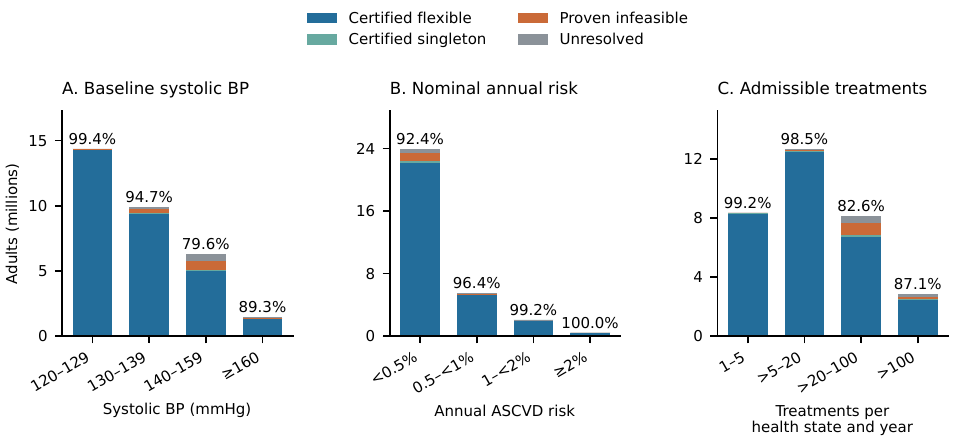}
\caption{Treatment flexibility across patient characteristics. Panels A--C group adults by baseline systolic BP, nominal annual cardiovascular risk, and the number of admissible treatments averaged over living health states and years. Labels give survey-weighted shares with certified flexible plans within each group.}
\label{fig:clinical-flexibility-profile}
\end{figure}

\subsubsection{Consequences of Selecting among Certified Treatments}
\label{sec:clinical-selection-results}

Selection-rule comparisons concern the 30.0 million adults with certified flexible plans (Figure~\ref{fig:clinical-selection-consequences}). Under the best-QALY rule, adults have a mean of 4.03 expected regimen changes over ten years. Choosing the fewest current medications reduces that mean by 0.63 changes per adult, or 15.7\%, at a mean cost of 0.0030 QALYs. Relative to the best-QALY rule, it gains 0.00053 QALYs during the 10-year horizon and loses 0.0035 QALYs afterward. Lower medication burden can improve the immediate reward while less treatment reduces expected health beyond the decision horizon. Expected regimen changes fall by at least 0.5 per adult over ten years for 52.0\% of this population. They increase by at least that amount for 6.9\%.

The fewest-regimen-changes rule reduces the mean by 1.6 changes per adult over ten years, costing 0.0023 QALYs relative to the best-QALY rule. For 84.4\% of adults, the reduction is at least 0.5 expected regimen changes over ten years. This rule loses 0.00058 and 0.0017 QALYs during and after the 10-year horizon, respectively. Preferences for fewer medications or greater continuity therefore cost about one quality-adjusted day on average under the nominal model.

The mean gap between the best and worst selections within the sets is 0.0062 QALYs. Relative to the unrestricted nominal optimum, their mean losses are 0.00043 and 0.0067 QALYs, respectively. At least one drug class can be avoided throughout the plan for 98.4\% of adults with certified flexible plans. Drug class-specific shares range from 15.1\% for angiotensin receptor blockers to 95.0\% for thiazides, with each restriction evaluated separately. These alternatives could honor drug-class preferences within the guarantee.

The guarantee also supports comparison with the 2017 ACC/AHA guidelines. Their recommended actions are admissible throughout the forecast family because we fix admissibility at nominal values. Under each model, no admissible treatment rule exceeds that model's optimum, including a rule that responds to treatment history. Every certified selection loses at most 0.03 QALYs relative to that optimum. It therefore loses at most 0.03 QALYs relative to the guidelines at every living health state and year. Under the nominal model, the best and worst selections gain means of 0.072 and 0.066 QALYs over the guidelines. The worst selection gives a lower bound on the gain for every rule.

\begin{figure}[t]
\centering
\includegraphics[width=\linewidth]{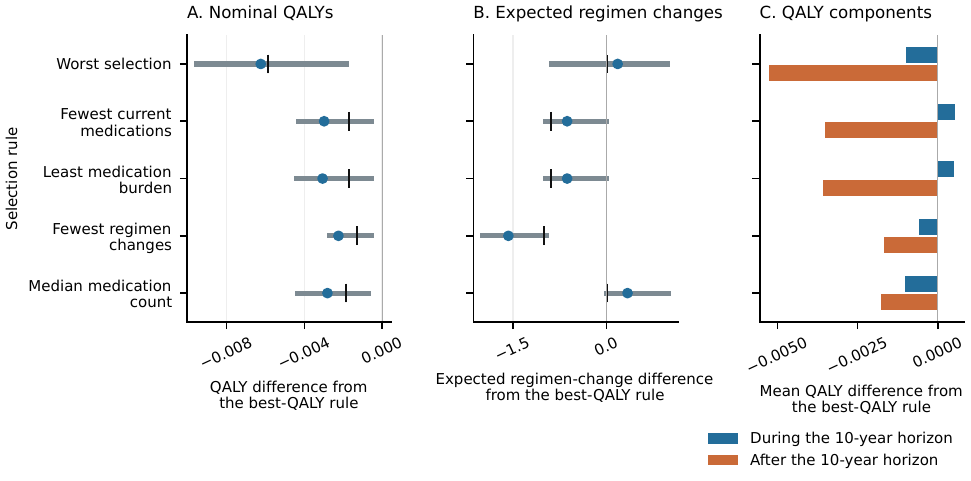}
\caption{Nominal outcomes of selection rules, which choose a recommended treatment according to a preference, within certified treatment plans. Comparisons concern adults with certified flexible plans and use the best-QALY rule as reference. Panels A and B show QALY and expected regimen-change differences, with dots for means, ticks for medians, and lines for interquartile ranges. Panel C shows mean QALY differences during and after the 10-year horizon, which sum to panel A’s mean. After-horizon values are expected QALYs after year ten, given the health state reached. Differences are calculated beginning from the healthy state.}
\label{fig:clinical-selection-consequences}
\end{figure}

\FloatBarrier
\endgroup
\section{Discussion and Conclusions}
\label{sec:discussion}

We construct set-valued policies that limit cumulative loss under every model in a specified uncertainty set. Each model's optimum serves as a benchmark, and our guarantee covers every compliant policy. For rectangular uncertainty, we give an exact decision formulation and establish the hardness of finding violations and maximizing weighted flexibility. Bellman certificates support incremental construction, and cellwise bounds extend it to continuous parametric families. Controlled experiments examine computation and choice, while a hypertension study evaluates treatment alternatives and their potential consequences.

The theoretical results distinguish exact decisions from sufficient certification. Theorem~\ref{thm:compact-membership} gives an exact test, and Theorem~\ref{thm:complexity-limits} establishes its computational difficulty. Theorems~\ref{thm:envelope-attainment-main} and~\ref{thm:certificate-soundness} give a test for whether any policy can satisfy the separated certificate. Coupled bounds reduce conservatism by comparing compliant and optimal choices within the same model. Theorem~\ref{thm:scalable-safety} permits certified action addition, and Propositions~\ref{prop:continuous-cell-safety} and~\ref{prop:clinical-cover-safety} extend the certification to a cover of the parameter domain. The guarantee includes randomized and history-dependent compliant choices, as shown in Remark \ref{rem:general-compliance}.

Our controlled experiments show how preserving shared dependence can certify alternatives that separate bounds miss. The e-companion presents these experiments in \cref{sec:shared-uncertainty-controls}. In the hypertension study, a clinically informed loss limit of 0.03 QALYs permits certified alternatives for 93.6\% of the choice-capable population, representing 30.0 million adults. Under the nominal model, the mean gap between the best and worst selections within these sets is 0.0062 QALYs. Their corresponding mean losses against the unrestricted nominal optimum are 0.00043 and 0.0067 QALYs. Choosing the fewest current medications reduces expected regimen changes by 0.63 per adult over ten years relative to the best-QALY rule, a reduction of 15.7\%. Its mean cost is 0.0030 QALYs. Presenting these consequences alongside certified alternatives can help clinicians and patients express preferences while maintaining health outcomes within an explicit loss limit.

Future clinical work could calibrate forecast families for simultaneous coverage across factors and years. Our certificates apply to the supplied model family. We can apply the same method to improved forecasts. The clinical predictions use a model calibrated against national fatal and nonfatal cardiovascular event rates, with clinical review of its parameters and logic \citep[Appendix B.2]{marrero2021ranges}. Future models could allow treatment responses and admissibility to vary, as the responsive analysis shows how requiring common admissibility can restrict choices. The certified share is a lower bound on the share admitting flexible plans. Tighter bounds and stronger existence tests could settle the 2.2\% of adults whose status remains unresolved. Prospective evaluation could establish how clinicians use these alternatives and how the choices affect care.

Models of multiple chronic conditions and uncertain treatment responses would account for competing benefits, burdens, and differences in who benefits from each option. Combining risk calculators would represent additional forecast uncertainty, especially for adults with moderately high systolic BP and low nominal risk, where infeasibility and unresolved cases concentrate. Preference-based ranking could help physicians display the alternatives most relevant to a patient.

Future theoretical questions extending our work include unweighted maximum cardinality and structural conditions relating flexibility to patient characteristics. Budgeted nonrectangular sets and families of probability distributions can offer additional uncertainty representations. Conformal prediction could also support calibration from data \citep{vovk2005}. For example, conformal regions can provide uncertainty sets for robust optimization \citep{johnstone2021}, and conformal intervals could cover a policy's realized cumulative reward in finite-horizon MDPs \citep{foffano2023}. Applying these methods in our setting would require simultaneous bounds on the expected values or model parameters that enter the certificate. Sublinear algorithms offer another direction for models with compact structure or sampling access. Reducing the states, actions, and parameter cells inspected by certification would require structural assumptions that preserve the full all-state, all-model guarantee.

Decision support does not have to choose between prescribing one action and offering choices with no guarantee. This paper shows how to offer choices with a guarantee when the model is uncertain. Our framework builds set-valued policies that limit the expected loss from repeated choices among their recommended actions. The bound holds under every plausible model, relative to that model's optimal policy. Our analysis also offers proof for when no policy can meet a given loss limit. Because the guarantee covers every way of choosing among the recommended actions, it does not require decision makers to specify their preferences in advance. The same principle applies wherever a finite decision model supports several acceptable alternatives, leaving room for human judgment under a stated performance limit.
\FloatBarrier

\clearpage
\section*{Electronic Companion}
\addcontentsline{toc}{section}{Electronic Companion}
This companion gives notation, proofs, construction procedures, and supplementary results for certified near-optimal set-valued policies under model uncertainty. It specifies the controlled experiments, finite-model infeasibility certificates, and hypertension models and analyses.
\par\medskip
\begin{APPENDICES}
\crefalias{section}{appendix}
\crefalias{subsection}{appendix}
\crefalias{subsubsection}{appendix}
\section{Notation and Finite-Horizon Representation}
\label{app:notation-representation}

The notation guide collects the symbols used in the manuscript. The finite-horizon representation below explains how the discounted framework applies when the state includes the decision epoch.

\begingroup
\small
\begin{longtable}{L{0.27\linewidth}L{0.66\linewidth}}
\caption{The guide includes main-text symbols, symbols in tables, figures, algorithms and theorem statements, and symbols used in multiple companion passages. Decision-epoch indices are added for finite horizons.}
\label{tab:notation-guide}\\
\toprule
Symbol & Meaning\\
\midrule
\endfirsthead
\multicolumn{2}{l}{Table~\thetable\ continued}\\
\toprule
Symbol & Meaning\\
\midrule
\endhead
\bottomrule
\endfoot
$\Sset$, $n$ & State space and number of states. In the controlled-outcome table, $n$ counts instances.\\
$\mathbb R$, $\mathbb R_{\ge0}$, $\mathbb E$ & Real numbers, nonnegative real numbers, and expectation.\\
$\boldsymbol 1\{E\}$ & Indicator of event or condition $E$.\\
$\Aset(s)$, $\mathcal K$, $m$ & Available actions at state $s$, available state-action pairs, and number of such pairs.\\
$a$, $a'$, $a''$ & Candidate or compliant action, comparator, and second comparator in action comparisons. Comparators may also be enabled.\\
$M=(r_M,P_M)$, $\Uset$ & Model rewards and transitions, and the uncertainty set.\\
$\gamma$, $T$, $t$ & Discount factor, finite horizon, and decision epoch.\\
$(t,s)$ & An epoch--state pair, called a position.\\
$\Selectors$, $\pi$, $\sigma$ & Deterministic stationary policy class and individual policies.\\
$\menu$, $\menu(s)$, $\MenuFamily$ & Set-valued policy, its menu at state $s$, and all set-valued policies. The mapping takes values in $2^{\Aset}\setminus\{\varnothing\}$.\\
$\Selectors(\menu)$ & Deterministic stationary policies satisfying $\pi(s)\in\menu(s)$ at every state.\\
$\Policies(\menu)$, $\mathfrak h_t$ & Randomized history-dependent compliant policies and an observed state-action history.\\
$V_M^\pi$, $V_M^\star$, $W_M^{\menu}$ & Policy value, optimal value, and worst compliant value in model $M$.\\
$\Reg_s(\menu)$, $\Reg(\menu)$ & Largest model-relative loss at state $s$, and the vector of these losses.\\
$\Delta(s)$, $\one$, $\norminf{v}$ & Statewise loss limit, vector of ones, and sup norm. Vector inequalities are componentwise.\\
$\Fexact$, $\Fcert$ & Exactly feasible policies and policies passing the separated certificate.\\
$\kappa_{sa}$ & Positive contribution of action $a$ at state $s$ to flexibility.\\
$\menu\preceq\menu'$ & Containment of set-valued policies at every state.\\
$\rlo_{sa},\rup_{sa}$, $\plo_{sa,j},\pup_{sa,j}$ & Lower and upper bounds on rewards and transition probabilities.\\
$\Pset_{sa}$, $\Uset_{\rm rect}$ & Admissible vectors for one transition row and the rectangular model set.\\
$\Bell_M^\star$, $\Bell_M^{\menu}$ & Optimal and worst-compliant Bellman operators for a fixed model.\\
$\Phi_{\menu,\Delta}$ & Polynomial constraint system for a model violating the loss requirement.\\
$\phi^-_{sa}(v),\phi^+_{sa}(v)$ & Minimum and maximum continuation expectations over $\Pset_{sa}$.\\
$\Bell^+,\Bell^-,\Bell^-_{\menu}$ & Optimistic optimal, pessimistic optimal, and pessimistic compliant operators.\\
$V^+,V^-,W^-_{\menu}$ & Attained fixed points of these three operators.\\
$B(\menu)$, $g^{\menu}$ & Separated and coupled upper bounds on cumulative loss.\\
$\mathcal H_{\menu}$, $\mathcal H^{C,\chi}_{\menu}$ & Rectangular coupled-loss operator and cellwise cumulative-loss operator.\\
$G^-_{sa}$, $G^+_{sa}$ & Lower and upper one-action Bellman backups.\\
$u_\Delta$, $\menu_{\rm loc}$, $\sigma^-$ & Local target, local set-valued policy, and pessimistic-greedy singleton.\\
$\bar u$, $\widehat T$, $\tau$ & Numerically enclosed local target, implemented Bellman backup, and requested iteration error.\\
$T$, $x$, $x^\star$ & Exact contraction map, comparison or iterate vector, and its fixed point in numerical bounds.\\
$\mathcal M$, $\mathcal X$ & Maximal-policy family in exact enumeration and candidate action additions in augmentation.\\
$Q$, $\mathrm{Seen}$, $\menu_{\rm full}$, $F$ & Search queue, visited policies, full-action policy, and a candidate maximal policy in exact enumeration.\\
$\mathcal B(\menu)$ & Indices of enabled bad actions in the weighted-optimization and exponential-output constructions.\\
$\mathcal Q$ & Probability polytope in the three-state example.\\
$q$, $q^1,\ldots,q^4$, $q^\circ$ & Uncertain transition row, its vertices, and the interior witness in the three-state example.\\
$M^+,M^-,M^-_{\menu}$ & Models simultaneously attaining the value envelopes.\\
$N_{\rm pol}$ & Number of nonempty set-valued policies.\\
$d^\star$, $\underline d$, $\overline d$ & Smallest achievable worst-case loss and its lower and upper existence bounds.\\
$N_{\rm add}$, $N_{\rm check}$, $K_{\max}$ & Maximum additions, number of certificate checks, and an upper bound on value-iteration sweeps per check.\\
$D(n,m)$, $J(\menu)$ & Dense transition-sweep cost and number of action pairs in a coupled sweep.\\
$K_+$, $K_-$, $K_{\menu}$, $K_H$, $K_e$ & Sweep counts for optimistic, pessimistic, compliant, coupled, and individual policy-check calculations.\\
$I_{\rm seed}$, $E_0$ & Indicator that singleton fallback is used and initial error bound for contraction iteration.\\
$\theta$, $\Theta$, $d$ & Shared parameter vector, its compact domain, and number of uncertain parameters.\\
$C$, $\mathcal C$ & One parameter cell and a finite cover of the domain.\\
$\underline V^C$, $\overline V^C$, $\underline W^{C,\menu}$ & Cellwise lower and upper optimal-value bounds and lower compliant-value bound.\\
$R_s^C$, $\overline D^C_{s,a',a}$ & Rival actions retained on a cell and upper action-value contrast bound.\\
$\overline\ell^{C,\chi}_{sa}$ & Upper one-step loss bound for cell certificate type $\chi$.\\
$B^{\mathcal C,\chi}$ & Largest cellwise loss bound over a complete cover.\\
$h_\theta$, $\underline h^C,\overline h^C$ & Terminal value and cellwise lower and upper enclosures.\\
$\underline\phi^C_{sa},\overline\phi^C_{sa}$ & Minimum and maximum continuation expectations over a cell's row enclosure.\\
$\overline Q^C_{sa},\underline Q^C_{sa}$ & Upper and lower optimal action-value bounds on a cell.\\
$\ell_\theta(s,a)$ & Loss from action $a$ followed by optimal continuation under $M(\theta)$.\\
$D_\theta(s;a',a)$ & Same-model action-value contrast between rival $a'$ and candidate $a$.\\
$\rho_M$, $\rho_\theta$ & True cumulative loss of compliant choices relative to the same model's optimum.\\
$\chi$ & Certificate type, one of raw, advantage, and joint.\\
$B^{C,\chi}(s)$ & Cumulative loss bound for certificate $\chi$ on cell $C$.\\
$\psi_t(\theta)$, $r$ & Epoch risk score and reference decision epoch in the risk-score cover.\\
$B$, $B_j$, $I$ & Outer parameter box, its covering subboxes, and a reference-score interval.\\
$R_j$, $D_{tj}$ & Interval enclosures of $\psi_r(B_j)$ and $(\psi_t-\psi_r)(B_j)$.\\
$F_{B,I}$, $J_{B,I}$ & Parameter points in $B$ with reference score in $I$, and subboxes whose score enclosure intersects $I$.\\
$E_t(B,I)$ & Enclosure of the epoch score on $F_{B,I}$.\\
$z$, $w$, $\delta$ & Clinical latent factors, uncertainty width multiplier, and common QALY-loss limit.\\
$\lambda$, $\Uset_\lambda$ & Rectangular width multiplier and the corresponding nested uncertainty set.\\
$w_r,w_P$ & Reward and transition interval widths in the synthetic experiments.\\
$\xi$, $\eta$ & Shared reward offset in controlled examples and the opposing-means example's half-width.\\
$\varpi_i$ & Pooled NHANES examination weight of source record $i$.\\
$\beta_a$ & Medication burden assigned to action $a$.\\
$\Delta x_{ijt}$ & Change in a regression-design coordinate from baseline.\\
$c_{ijt}$, $s_{ijt}$, $L_{ijt},U_{ijt}$ & Nominal forecast center, uncertainty scale, and positive forecast-band endpoints.\\
$\widehat V_j$, $\widehat\sigma_j^2$, $x_{ijt}(z_j;w)$ & Fitted coefficient covariance, residual variance, and factor forecast under the shared parameter.\\
$q_{it}(S,C,H)$, $S,C,H$ & Annual untreated cardiovascular risk and its systolic BP, total-cholesterol, and HDL inputs.\\
$D^S_{ita},D^D_{ita}$, $RR_{ite}$, $E_{itae}$, $u_{its}$ & Treatment-induced systolic and diastolic BP reductions, event relative risk, resulting risk multiplier, and quality factor.\\
$n,h$ & Numbers of standard-dose and half-dose agents in the generic-dose catalogs.\\
$n_B,n_K$, $\alpha$, $Y_{abk}$ & Numbers of inference batches and observations, comparison significance level, and simulated action-value observation.\\
$\bar Y_{ab}$, $s^2_{ab}$, $\widehat Q_a$, $S_a^2$ & Within-batch mean and variance, average estimated action value, and between-batch variance.\\
$\widehat v_t$, $Z_{ab}$, $z_{1-\alpha}$, $L_a$ & Supplied continuation values, studentized comparison root, empirical quantile, and comparison lower bound.\\
$c$, $Q^*$ & Independently selected control action and optimal action values targeted by statistical coverage.\\
$M_k$, $N_{\rm mod}$, $K$ & A listed model and the model counts in the finite-horizon and sepsis existence tests.\\
$Y_{M_k,t}$, $L_{M_k,t}$ & Exact loss supersolutions and loss variables in the outer relaxation.\\
$q$, $\lfloor z\rfloor_q$ & Binary rounding exponent and downward rounding to multiples of $2^{-q}$.\\
$\ell_k(s,a)$, $\rho_k$ & One-step loss and cumulative policy loss under finite-list model $k$.\\
$\underline\rho_k(s)$, $\zeta_k(s,a)$, $\rho_\Sigma$ & Lower bound on cumulative loss, action-elimination bound, and sum of losses over finite-list models.\\
$\underline\ell_{k,tsa}$, $\underline c_{k,tsa,j}$ & Direct lower bounds on original-model action loss and discounted continuation coefficients.\\
$\mathcal P_{k,tsa}$, $A'_t(s)$ & Point-model transition-row enclosure intersected with the probability simplex, and actions surviving a checked screen.\\
$\underline P_{k,tsa,j}$, $\mu_{k,ts}(a',a)$ & Point-model probability lower endpoint and strict continuation-box dominance margin.\\
$\Delta P_{M_k,tsj}(a',a)$ & Difference between two actions' transition probabilities in the dominance test.\\
$w_j=j/N$, $N$, $b_{t,s,j}$, $\mathcal B_{t,s}(w)$ & Rational two-model mixture weights, number of grid intervals, lower loss bounds on the grid, and their linear interpolation.\\
$x_{s'}$, $y_{s'}$, $m_{s'}$ & Nonnegative transition masses for two listed models and their sum in the pair-loss bound.\\
$\mathcal J$, $z$ & Coupled subsolution operator and a lower comparison vector for the sum of model losses in the sepsis existence screen.\\
\end{longtable}
\endgroup

\subsection{Finite-horizon representation}
\label{app:finite-horizon}

Suppose decisions occur at epochs $t=0,\ldots,T-1$, with reward
$r_{M,t}(s,a)$, transition probabilities $P_{M,t}(\cdot\mid s,a)$,
nonempty action sets $\Aset_t(s)$, and terminal value $h_M(s)$.
For $0\le\gamma<1$, replace each decision state by $(t,s)$, pay
$h_M(s)$ once at $(T,s)$, and then enter a zero-reward absorbing state.
At each terminal state and at the absorbing state, one action is compulsory.
The value from $(t,s)$ is
\begin{equation*}
 \mathbb E_M^\pi\!\left[
 \sum_{u=t}^{T-1}\gamma^{u-t}r_{M,u}(S_u,A_u)
 +\gamma^{T-t}h_M(S_T)\,\middle|\,S_t=s\right].
\end{equation*}
This augmented finite-state discounted MDP represents the original
finite-horizon problem. A deterministic stationary policy on the enlarged
space specifies a deterministic Markov policy on the original states
\citep[Chapter~4]{puterman}. Both the recommendation and the optimal benchmark
receive the same terminal value in each model, so their loss at the terminal epoch $T$
is zero. The compulsory terminal and absorbing actions add a constant to
the flexibility objective.

For $\gamma=1$, the finite-horizon values and loss bounds follow directly
by backward induction. The contraction arguments for an infinite horizon
require $\gamma<1$ and are used only under that restriction.

\subsection{Finite-horizon cell certificates}
\label{app:finite-cell-certificates}

The clinical implementation works backward over the finite horizon.
This gives specialized bounds that exploit the zero terminal loss.
For a cell $C$, enclose the terminal value by
$\underline h^C\le h_\theta\le\overline h^C$ and use epoch-specific
reward and row enclosures. Initialize
$\underline V_T^C=\underline W_T^{C,\menu}=\underline h^C$ and
$\overline V_T^C=\overline h^C$. At decision epochs, set
\begin{align*}
 \underline Q^C_{tsa}
 &=\rlo^C_{tsa}+\gamma\underline\phi^C_{tsa}(\underline V^C_{t+1}),&
 \overline Q^C_{tsa}
 &=\rup^C_{tsa}+\gamma\overline\phi^C_{tsa}(\overline V^C_{t+1}),\\
 \underline V^C_t(s)&=\max_a\underline Q^C_{tsa},&
 \overline V^C_t(s)&=\max_a\overline Q^C_{tsa},\\
 \underline W_t^{C,\menu}(s)
 &=\min_{a\in\menu_t(s)}
 \{\rlo^C_{tsa}+\gamma\underline\phi^C_{tsa}
                         (\underline W_{t+1}^{C,\menu})\}.&&
\end{align*}
The row functions have the same definitions as in the main text, with
the decision-epoch index added. Backward induction yields
$\underline W_t^{C,\menu}\le W^{\menu}_{\theta,t}$ and
$\underline W_t^{C,\menu}\le\underline V_t^C
\le V^\star_{\theta,t}\le\overline V_t^C$.

For $t<T$, define the one-step loss and its bounds as in Section~\ref{sec:cellwise-bounds},
using $V^\star_{\theta,t+1}$ in the contrasts and the corresponding
epoch-specific action-value enclosures. In particular,
\begin{equation*}
 D_{\theta,t}(s;a',a)=r_{\theta,t}(s,a')-r_{\theta,t}(s,a)
 +\gamma\bigl(P_{\theta,t}(\cdot\mid s,a')-P_{\theta,t}(\cdot\mid s,a)\bigr)^\top
 V^\star_{\theta,t+1}.
\end{equation*}
The rival set contains every action optimal in any model of the cell.
Taking the minimum of the advantage bound and the nonnegative upper
contrast bound gives $\overline\ell^{C,\mathrm{joint}}_{tsa}$, as in
the main construction. Define
\begin{align*}
 B_T^{C,\chi}&=0\quad(\chi\in\{\mathrm{raw},\mathrm{adv},\mathrm{joint}\}),\\
 B_t^{C,\mathrm{raw}}&=\overline V_t^C-\underline W_t^{C,\menu}
 \quad(t<T),\\
 B_t^{C,\chi}(s)&=\max_{a\in\menu_t(s)}
 \{\overline\ell^{C,\chi}_{tsa}
          +\gamma\overline\phi^C_{tsa}(B_{t+1}^{C,\chi})\}
 \quad(\chi\in\{\mathrm{adv},\mathrm{joint}\},\ t<T).
\end{align*}
These bounds satisfy the soundness and ordering conclusions of
Proposition~\ref{prop:continuous-cell-safety} at every position, including
when $\gamma=1$. To verify soundness directly, subtract the two modelwise
Bellman recursions to obtain
\begin{equation*}
 \rho_{\theta,T}=0,\qquad
 \rho_{\theta,t}(s)=\max_{a\in\menu_t(s)}
 \{\ell_{\theta,t}(s,a)
       +\gamma P_{\theta,t}(\cdot\mid s,a)^\top\rho_{\theta,t+1}\}.
\end{equation*}
The value enclosures prove raw soundness. The one-step contrast argument
in \cref{app:continuous-family-proofs}, followed by backward induction
in this identity, proves advantage and joint soundness. The ordering of
local losses proves $B_t^{C,\mathrm{joint}}\le B_t^{C,\mathrm{adv}}$.

For the raw--advantage comparison, the two bounds are equal at $T-1$
because $\underline W_T^{C,\menu}=\underline V_T^C$. At each earlier
epoch, the difference-of-row-minima inequality used in the discounted
proof gives
\begin{equation*}
 B_t^{C,\mathrm{raw}}(s)\le\max_{a\in\menu_t(s)}
 \{\overline\ell^{C,\mathrm{adv}}_{tsa}
       +\gamma\overline\phi^C_{tsa}(B_{t+1}^{C,\mathrm{raw}})\}.
\end{equation*}
Backward induction now proves $B_t^{C,\mathrm{raw}}\le B_t^{C,\mathrm{adv}}$.
Taking the maximum over a complete cell cover proves the finite-horizon
analogue of Proposition~\ref{prop:clinical-cover-safety}.

When terminal values depend on the uncertain parameters, these specialized
bounds need not equal the generic cell bounds on the augmented discounted
MDP. Generic raw and advantage bounds may retain the gap between terminal
value enclosures even though the single compulsory terminal action has
zero true loss. The backward recursions use this exact zero-loss boundary
condition. They retain terminal-value uncertainty in the earlier action
comparisons. The model values and true losses are unchanged by the
augmented-state representation. The specialized certificates exploit
additional structure when bounding them.

\FloatBarrier
\section{Proofs, Constructions, and Verification Procedures}
\label{app:proofs}
\label{app:extra-methods}

This appendix follows the main mathematical argument. Containment and existence support the problem formulation. The repeated results then proceed from exact verification and its limits to rectangular certificates, parametric-family certificates, and construction. Implementation details follow the proofs.

Algorithm~\ref{alg:master-construction} links the construction procedures
and follows the labeled steps in Figure~\ref{fig:method-overview}.
The exact family search of Algorithm~\ref{alg:complete-exact-appendix}
is the reference method of Section~\ref{sec:rectangular-methods}.

\begin{algorithm}[htbp]
\caption{Policy construction with certified outcomes}
\label{alg:master-construction}
\begin{algorithmic}[1]
\Require \textbf{Input} state and action sets, uncertainty family $\Uset$, loss limit $\Delta$, and computation limits
\State \textbf{Propose} local or singleton candidates for rectangular uncertainty;
point-model optima, finite-model minimax proposals, minimum worst local-loss choices,
and bounded single-position repairs for parametric uncertainty
\State \textbf{Bound} separated or coupled loss for rectangular uncertainty;
cellwise loss on a complete parameter cover using Algorithm~\ref{alg:family-verify}
for parametric uncertainty
\State \textbf{Test} compare the candidate's certified bound with $\Delta$
\If{a candidate passes}
  \State \textbf{Add} use best-margin augmentation in Algorithm~\ref{alg:scalable-main}
or fixed-order augmentation in Algorithm~\ref{alg:family-search}; for a rectangular
fixed-order pass, use state and action indices
  \State \textbf{Output} the final certified policy and its checked bound
\Else
  \State Analyze existence on finite models from $\Uset$ and check any proof
  \State Use the finite list to propose and test further candidates while computation permits
  \If{a policy passes}
    \State Apply \textbf{Add} and \textbf{Output} its certified policy
  \ElsIf{an infeasibility proof passes its check}
    \State \textbf{Output} an infeasibility certificate
  \Else
    \State \textbf{Output} unresolved with existence bounds and stop reason
  \EndIf
\EndIf
\end{algorithmic}
\end{algorithm}

All vector comparisons in this appendix are componentwise. The sup norm is $\norminf{x}=\max_s|x(s)|$. We use two
standard facts about finite discounted MDPs. Their Bellman operators are
monotone contractions, and deterministic stationary policies attain their
optimal values \citep[Proposition~6.2.4 and Theorem~6.2.10]{puterman}. The fixed-point comparison used below has a
short proof. If $T$ is a monotone contraction with fixed point $x^\star$ and
$x\le Tx$, then
$x\le Tx\le T^2x\le\cdots$, and convergence gives $x\le x^\star$.
Similarly, $Tx\le x$ gives $x^\star\le x$. The first inequality makes $x$
a lower comparison vector, often called a subsolution. The second makes it
an upper comparison vector, or supersolution.

The compactness condition in the problem setting also has a direct role.
For a fixed deterministic policy, its value is
$(I-\gamma P)^{-1}r$. This inverse exists for every stochastic $P$ when
$\gamma<1$, and it varies continuously with $P$ and $r$. There are finitely
many deterministic policies. Their componentwise maxima, and the minima
over policies of any fixed set-valued policy, are therefore continuous. The optimal
value, worst-compliant value, and their difference attain their extrema on
the compact model set. A different model may attain each extremum unless
an additional argument gives simultaneous attainment.

\subsection{Containment and existence}
\begin{lemma}[Containment and singleton existence]
\label{lem:menu-monotonicity}
If $\menu\preceq\menu'$, then $W_M^{\menu'}\le W_M^{\menu}$ for every model and $\Reg(\menu)\le\Reg(\menu')$.
A nonempty safe set-valued policy exists if and only if a safe deterministic stationary policy exists.
\end{lemma}
\begin{proof}{Proof.}
We first compare the two policy sets, then prove both directions of the existence claim.
Every policy compliant with $\menu$ is also compliant with $\menu'$. Minimizing over the larger policy set gives $W_M^{\menu'}\le W_M^{\menu}$.
Subtracting from $V_M^\star$ and maximizing over models gives the loss ordering.

For the forward direction, choose any policy $\sigma$ compliant with a safe set-valued policy $\menu$.
This single policy is chosen independently of the model. For every $M\in\Uset$,
\[
V_M^\sigma\ge W_M^{\menu}\ge V_M^\star-\Delta.
\]
Thus $\sigma$ is safe in every model. For the reverse direction, a safe deterministic stationary policy defines a safe singleton set-valued policy.
\Halmos
\end{proof}
\subsection{Worst compliant value}
\begin{repeatproposition}[Proposition~\ref{prop:worst-selector} (Worst compliant policy).]
For each model $M$ and set-valued policy $\menu$, $W_M^{\menu}$ is the
unique fixed point of $\Bell_M^{\menu}$. One deterministic stationary
policy $\sigma\in\Selectors(\menu)$ attains
$W_M^{\menu}(s)=\min_{\sigma'\in\Selectors(\menu)}V_M^{\sigma'}(s)$
simultaneously for every $s\in\Sset$.
\end{repeatproposition}
\begin{proof}{Proof of \cref{prop:worst-selector}.}
We reduce the minimizing problem to an ordinary discounted maximization problem. Fix a model and restrict its available actions to $\menu(s)$. Negate the
rewards. The Bellman operator for maximizing the resulting return is
\begin{equation*}
 (\widetilde\Bell v)(s)=\max_{a\in\menu(s)}
 \{-r_M(s,a)+\gamma P_M(\cdot\mid s,a)^\top v\}.
\end{equation*}
It satisfies $-\widetilde\Bell(-w)=\Bell_M^{\menu}w$. Each policy's value in the negated problem is $-V_M^\pi$, so that problem's optimal value is $-W_M^{\menu}$. Thus the negative of
its optimal value is the unique fixed point of the minimizing operator
$\Bell_M^{\menu}$. A deterministic stationary policy maximizes the negated
return from every state, even when compared with randomized,
history-dependent policies \citep[Theorem~6.2.10]{puterman}. Negating returns shows that this same policy
attains the minimum defining $W_M^{\menu}$ at every state. It chooses only
recommended actions, so it belongs to $\Selectors(\menu)$.
\Halmos
\end{proof}

\subsubsection{Extension to randomized and history-dependent choices.}
\label{app:compliant-choices}
A history at epoch $t$ is $\mathfrak h_t=(s_0,a_0,\ldots,s_t)$. Let $\Policies(\menu)$ contain all policies that assign a probability distribution supported on $\menu(s_t)$ after each history.
The model remains fixed throughout the trajectory. The negated-reward result used above holds over this larger class \citep[Theorem~6.2.10]{puterman}.
Consequently,
\[
\inf_{\pi\in\Policies(\menu)}V_M^\pi(s)
=\min_{\sigma\in\Selectors(\menu)}V_M^\sigma(s)
=W_M^{\menu}(s).
\]
The deterministic stationary policy attains the infimum from every initial state. Hence every loss certificate proved using policies also protects randomized and history-dependent compliant choices.
\subsection{Exact verification}\label{ec_subsec:exact_verification}
\begin{repeattheorem}[Theorem~\ref{thm:compact-membership} (Exact test for one set-valued policy).]
Suppose $\Uset=\Uset_{\rm rect}$. A set-valued policy $\menu$ exceeds its
loss limit at some state, that is, $\menu\notin\Fexact$, if and only if
$\Phi_{\menu,\Delta}$ has a real solution.
\end{repeattheorem}
\begin{proof}{Proof of \cref{thm:compact-membership}.}
We describe the complete system and then prove that its solutions are exactly the violating models with their value vectors.
The real variables are rewards $r_{sa}$, probabilities $p_{sa,j}$, and values $v_s,w_s$, for $(s,a)\in\mathcal K$ and $s,j\in\Sset$.
For each available state-action pair, impose
\begin{align*}
\rlo_{sa}\le r_{sa}\le\rup_{sa},\qquad
\plo_{sa,j}\le p_{sa,j}\le\pup_{sa,j}\quad(j\in\Sset),\qquad
\sum_jp_{sa,j}=1.
\end{align*}
The Bellman constraints are
\begin{align*}
&v_s\ge r_{sa}+\gamma\sum_jp_{sa,j}v_j
&&\text{for every }a\in\Aset(s),\\
&\bigvee_{a\in\Aset(s)}\left(v_s=r_{sa}+\gamma\sum_jp_{sa,j}v_j\right),\\
&w_s\le r_{sa}+\gamma\sum_jp_{sa,j}w_j
&&\text{for every }a\in\menu(s),\\
&\bigvee_{a\in\menu(s)}\left(w_s=r_{sa}+\gamma\sum_jp_{sa,j}w_j\right),
\end{align*}
for every $s\in\Sset$. The final condition is
\[
\bigvee_{s\in\Sset}\bigl(v_s-w_s>\Delta(s)\bigr).
\]
Here a disjunction means that at least one of its alternatives holds. Together these conditions define $\Phi_{\menu,\Delta}$.

\emph{A solution implies violation.}
The reward and probability variables specify one admissible complete model $M$.
The Bellman inequalities and equality alternatives imply $v=\Bell_M^\star v$ and $w=\Bell_M^{\menu}w$.
Uniqueness of their fixed points gives $v=V_M^\star$ and $w=W_M^{\menu}$.
The final disjunction therefore identifies a state at which the policy exceeds its loss limit.

\emph{Violation implies a solution.}
Suppose an admissible model $M$ has $V_M^\star(s)-W_M^{\menu}(s)>\Delta(s)$ at some state.
Assign the model's rewards and probabilities to $r,p$, and assign $v=V_M^\star$ and $w=W_M^{\menu}$.
The model satisfies all interval and row-sum constraints. Its Bellman equations supply the inequalities and at least one equality at each state.
The violating state satisfies the final disjunction. These assignments solve the system, proving the converse.

\emph{Description size and complexity.}
There are $(n+1)m+2n$ real variables. All polynomial constraints have degree at most two because their only products pair a probability with a value.
With rational input, the number of terms and their encoding lengths are polynomial in the input size.
Thus violation belongs to $\exists\mathbb R$, the class defined by polynomial-size existential real formulas \citep{schaeferstefankovic2017}.
Safety is the complementary decision problem. Its weak loss inequality is why the violation formula uses strict excess.\Halmos
\end{proof}
\subsection{An interior transition witness}
\label{app:nonvertex}

\begin{repeatproposition}[Proposition~\ref{prop:nonvertex-main} (Transition vertices are insufficient).]
There is a three-state interval MDP and a singleton policy whose
model-relative loss at one state is strictly larger at an interior
transition row than at every vertex of that row's uncertainty set.
\end{repeatproposition}
\begin{proof}{Proof of \cref{prop:nonvertex-main}.}
We compare exact policy values at every transition vertex with one interior row. The construction has
$\Sset=\{0,1,2\}$ and $\gamma=9/10$. States 0 and 1 have actions $A,B$, while
state 2 has only action $C$. The rewards are
\begin{equation*}
\begin{array}{c|cc}
 & A & B\\ \hline
s=0 & -3 & -2\\
s=1 & -2 & -1
\end{array}
\qquad r(2,C)=3.
\end{equation*}
The fixed transition rows at states 0 and 1 are
\begin{align*}
 p_{0A}&=(1/5,1/5,3/5),
 &p_{0B}&=(3/10,1/10,3/5),\\
 p_{1A}&=(1/10,2/5,1/2),
 &p_{1B}&=(4/5,1/5,0).
\end{align*}
Only the row $p_{2C}=q$ is uncertain. Its feasible set is
\begin{equation*}
 \mathcal Q=\{q\in\mathbb R_{\ge0}^3:
 \one^\top q=1,\ (1/5,1/10,3/10)\le q\le(2/5,7/10,7/10)\}.
\end{equation*}
The four vertices of this polygon are
\begin{align*}
q^1&=(1/5,1/10,7/10),
&q^2&=(1/5,1/2,3/10),\\
q^3&=(2/5,3/10,3/10),
&q^4&=(2/5,1/10,1/2).
\end{align*}

For a two-letter policy $\tau$, the first letter is the action at state 0 and
the second is the action at state 1. Let the set-valued policy have the
singleton menu $\{B\}$ at both states, so its value is $V^{BB}$. Solving
\begin{equation*}
  (I-\gamma P^\tau(q))V^\tau(q)=r^\tau
\end{equation*}
with rational arithmetic shows that policy $BA$ is optimal at state 0 for each
row in \cref{tab:nonvertex-values}.

\begin{table}[htbp]
\centering
\caption{Exact loss at the transition vertices and one interior row.}
\label{tab:nonvertex-values}
\begin{tabular}{lcc}
\toprule
Row & exact $V^{BA}(0)-V^{BB}(0)$ & decimal \\
\midrule
$q^1$ & $807480/604513$ & 1.335753 \\
$q^2$ & $37167480/25908419$ & 1.434572 \\
$q^3$ & $11886300/10744187$ & 1.106301 \\
$q^4$ & $98460/103441$ & 0.951847 \\
\midrule
$q^\circ=(21/100,11/40,103/200)$
& $1065691260/655194841$ & 1.626526 \\
\bottomrule
\end{tabular}
\end{table}

Exact comparisons give
\begin{equation*}
  V^{BA}(0)>V^{BB}(0)>V^{AA}(0),V^{AB}(0)
\end{equation*}
at the five reported rows.
Every coordinate of $q^\circ$ lies strictly between its bounds, and the
coordinates sum to one. Thus $q^\circ$ lies in the relative interior of
$\mathcal Q$. Its loss is strictly larger than the loss at all four
vertices. 
\Halmos
\end{proof}

\subsection{Complexity limits}
\begin{repeattheorem}[Theorem~\ref{thm:complexity-limits} (Exact complexity limits).]
Suppose $\Uset=\Uset_{\rm rect}$.
\begin{enumerate}[label=(\roman*),leftmargin=2.2em]
\item Deciding whether a fixed set-valued policy has a violating model is
NP-hard, even with single-valued rewards, meaning rewards fixed at one
known value, any fixed rational discount factor $\gamma\in(0,1)$, and
singleton menus. Consequently, deciding whether the policy is safe is
coNP-hard under the same restrictions.
\item Finding a safe policy with the largest weighted action count is
NP-hard, even for one known model.
\end{enumerate}
\end{repeattheorem}
\begin{proof}{Proof of \cref{thm:complexity-limits}.}
We use a graph-cut reduction for verification and a knapsack reduction for weighted optimization.

\emph{Part (i), violation.}
\label{app:maxcut-proof}

We prove part (i) of \cref{thm:complexity-limits} by reducing Max-Cut
\citep{gareyjohnson}. Let $G=(\mathcal N,E)$ be an undirected graph with $n$ vertices and
$m$ edges. The decision problem asks whether $G$ has a cut of size at least
$k$. We may restrict to simple graphs with $n\ge1$ and $1\le k\le m$.
Give every edge an arbitrary orientation, denoted by $E^\to$, and set
$H=m+1$.

For each vertex $i$, introduce two numbers $x_i,y_i\in[0,1]$. Consider the
multiaffine polynomial
\begin{align}
 F(x,y)
 ={}&H\sum_{i=1}^n(1-x_i-y_i+2x_iy_i)\notag\\
 &+\sum_{(i,j)\in E^\to}(x_i+y_j-2x_iy_j).
 \label{eq:maxcut-polynomial-appendix}
\end{align}
The first sum rewards $x_i$ and $y_i$ agreeing, after which the second counts edges with unequal binary endpoints.

Holding all other coordinates fixed, an affine function attains its maximum
at an endpoint of the remaining coordinate interval. Applying this step
successively to all coordinates shows that $F$ has a binary maximizer. At a binary point, changing a disagreeing $x_i$ to $y_i$ gains $H$
in the first sum. It can lose at most one on each outgoing edge, for a total of
at most $m$. Since $H=m+1$, every binary maximizer has $x=y$. The remaining terms
count a cut, and therefore
\begin{equation*}
 \max_{x,y\in[0,1]^n}F(x,y)=Hn+\operatorname{MaxCut}(G).
\end{equation*}

Write the polynomial as
\begin{equation*}
 F(x,y)=c+\sum_i\alpha_i x_i+\sum_j\beta_j y_j
       +\sum_{i,j}q_{ij}x_iy_j.
\end{equation*}
Here $c=Hn$, $\alpha_i=-H+\deg^+(i)$, and
$\beta_i=-H+\deg^-(i)$, where $\deg^+$ and $\deg^-$ count outgoing and
incoming oriented edges. The coefficients are $q_{ii}=2H$,
$q_{ij}=-2$ for $(i,j)\in E^\to$, and zero otherwise.
The following layered MDP converts this polynomial into a difference of
action values under a common model.
Fix any rational $\gamma\in(0,1)$ and define
\begin{equation*}
 Q_{\max}=\max_i\sum_j\abs{q_{ij}},
 \qquad
 \lambda=\frac{\gamma^2}{1+Q_{\max}},
 \qquad
 \omega_{ij}=\frac{\lambda q_{ij}}{\gamma^2}.
\end{equation*}
For $[z]_+=\max\{z,0\}$, both
$\sum_j[\omega_{ij}]_+$ and $\sum_j[-\omega_{ij}]_+$ are at most one. They can therefore
serve as parts of transition rows.

The construction uses a fixed singleton policy, so the uncertain rows alone encode the graph-cut choices. Absorbing states fix the values zero and one, and the states $R_j$ encode the variables $y_j$. The states $A_i^+$, $A_i^-$, and $L_i$ assemble the bilinear terms. Two actions at the start state compare their sum, and a half-unit offset will separate the two cut outcomes.

Create two absorbing states $z,h$. State $z$ has reward zero. State $h$ has
self-loop reward $1-\gamma$. Their values are zero and one. For each $j$, add a
one-action state $R_j$ with zero reward and uncertain transition row
\begin{equation*}
 P(h\mid R_j)=y_j,
 \qquad P(z\mid R_j)=1-y_j.
\end{equation*}
Its value is $V(R_j)=\gamma y_j$.

For each $i$, add one-action states $A_i^+$ and $A_i^-$. Their rows are
\begin{align*}
 P(R_j\mid A_i^+)&=[\omega_{ij}]_+,
 &P(z\mid A_i^+)&=1-\sum_j[\omega_{ij}]_+,\\
 P(R_j\mid A_i^-)&=[-\omega_{ij}]_+,
 &P(z\mid A_i^-)&=1-\sum_j[-\omega_{ij}]_+.
\end{align*}
Set $r(A_i^+)=\lambda\alpha_i$ and $r(A_i^-)=0$. The two values are
\begin{equation*}
 V(A_i^+)=\lambda\alpha_i+\gamma^2\sum_j[\omega_{ij}]_+y_j,
 \qquad V(A_i^-)=\gamma^2\sum_j[-\omega_{ij}]_+y_j.
\end{equation*}
Using $[\omega]_+-[-\omega]_+=\omega$ and $\gamma^2\omega_{ij}=\lambda q_{ij}$ gives
\begin{equation*}
 V(A_i^+)-V(A_i^-)
 =\lambda\left(\alpha_i+\sum_jq_{ij}y_j\right).
\end{equation*}
Next add a one-action state $L_i$ with zero reward and row
\begin{equation*}
 P(A_i^+\mid L_i)=x_i,
 \qquad P(A_i^-\mid L_i)=1-x_i.
\end{equation*}
It follows that
\begin{equation*}
 V(L_i)-\gamma V(A_i^-)
 =\gamma\lambda x_i
 \left(\alpha_i+\sum_jq_{ij}y_j\right).
\end{equation*}

The start state $s_0$ has actions $a',a$. Define a signed vector over the
internal states by
\begin{equation*}
 \Xi(L_i)=\frac{1}{\gamma^2\lambda},\qquad
 \Xi(A_i^-)=-\frac{1}{\gamma\lambda},\qquad
 \Xi(R_j)=\frac{\beta_j}{\gamma^2}.
\end{equation*}
Let $\eta=(1+\sum_u\abs{\Xi(u)})^{-1}$. The two start-state rows are
\begin{align*}
 P(u\mid s_0,a')&=\eta[\Xi(u)]_+,
 &P(u\mid s_0,a)&=\eta[-\Xi(u)]_+.
\end{align*}
The displayed entries are nonnegative and sum to at most $\eta\sum_u\abs{\Xi(u)}<1$ in each row. The unused probability in each row goes to $z$. Finally, set
$r(s_0,a')=\eta c$ and $r(s_0,a)=0$. If $\mathsf Q(s_0,a')$ denotes the
start-state action value, $V(z)=0$ gives
\begin{equation*}
 \mathsf Q(s_0,a')-\mathsf Q(s_0,a)
 =\eta c+\gamma\eta\sum_u\Xi(u)V(u).
\end{equation*}
The terms for $L_i$ and $A_i^-$ in the sum combine to
\begin{equation*}
 \frac{V(L_i)-\gamma V(A_i^-)}{\gamma^2\lambda}
 =\frac{x_i}{\gamma}\left(\alpha_i+\sum_jq_{ij}y_j\right),
\end{equation*}
and the term for $R_j$ is $\beta_jy_j/\gamma$. Substitution gives
\begin{equation*}
 \mathsf Q(s_0,a')-\mathsf Q(s_0,a)=\eta F(x,y).
\end{equation*}

The uncertain rows at $R_j$ and $L_i$ each have interval bounds zero and one on their two listed entries, with all other entries zero. All other rows and all rewards are fixed. Thus the model set is $\Uset_{\rm rect}$, and the coordinates of $(x,y)\in[0,1]^{2n}$ vary independently.
The fixed set-valued policy enables only $a$ at $s_0$. Every other state has
one action.
The agreement term equals $x_i y_i+(1-x_i)(1-y_i)$, and an edge term
equals $x_i(1-y_j)+(1-x_i)y_j$. Both are nonnegative throughout the unit
box. Thus $a'$ is optimal at $s_0$ for every admissible model. The policy loss there is $\eta F(x,y)$, while
loss at every other state is zero. Set
\begin{equation*}
 \Delta(s_0)=\eta(Hn+k-1/2),
 \qquad \Delta(s)=0\quad(s\ne s_0).
\end{equation*}
If the maximum cut has size at least $k$, a maximizing binary model has
loss at least $\eta(Hn+k)$, strictly above the threshold. If it has size
at most $k-1$, every model has loss at most $\eta(Hn+k-1)$, strictly below
the threshold. The half-unit offset therefore gives exactly the required
strict violation test. The construction has $4n+3$
states, at most two actions per state, known rewards, and rational data of
polynomial encoding length. The construction uses only $0<\gamma<1$, so
the reduction holds for every fixed rational $\gamma\in(0,1)$. Since Max-Cut is NP-hard and this reduction has polynomial size, deciding whether the fixed policy has a violation is NP-hard.

\emph{Part (i), safety.}
The same mapping sends a graph with no cut of size at least $k$ to a safe policy, and a graph with such a cut to an unsafe policy. It is therefore a reduction from the complement of Max-Cut to safety. This proves the stated coNP-hardness.

\emph{Part (ii), weighted optimization.}
Start from a 0--1 knapsack instance with positive item
costs $c_i$, profits $u_i$, and capacity $B_{\rm cap}$. Use a root $s_0$ with one zero-reward action leading uniformly to item states $s_1,\ldots,s_n$. Each item state has a good action of reward zero and a bad action of reward $-c_i$. Both actions lead to a zero-reward absorbing state. Fix any rational $\gamma\in(0,1)$ and set
\begin{equation*}
 \Delta(s_0)=\frac{\gamma B_{\rm cap}}{n},
 \qquad \Delta(s_i)=c_i,\qquad \Delta(s_{\rm term})=0.
\end{equation*}
Write $\mathcal B(\menu)=\{i:\text{the bad action is enabled at }s_i\}$. Give bad action $i$ weight $u_i$. Give every remaining action any fixed
positive weight. The good actions cannot decrease a worst-compliant value when added,
because their reward is larger and they have the same transition as the
bad action. Their positive weights therefore put them in every optimum.
The remaining required actions are compulsory. Together these actions
contribute a constant. The worst compliant policy takes every enabled bad action, so
\begin{equation*}
 \Reg_{s_i}(\menu)=c_i\boldsymbol 1\{i\in\mathcal B(\menu)\}\le\Delta(s_i),
 \qquad \Reg_{s_0}(\menu)=\frac{\gamma}{n}\sum_{i\in\mathcal B(\menu)}c_i.
\end{equation*}
The limit at $s_0$ holds exactly when
\begin{equation*}
  \sum_{i\in\mathcal B(\menu)}c_i\le B_{\rm cap},
\end{equation*}
while the enabled bad actions contribute
$\sum_{i\in\mathcal B(\menu)}u_i$ to the objective. An optimal weighted set-valued policy therefore solves
the knapsack instance. This polynomial reduction from 0--1 knapsack establishes
NP-hardness of the weighted objective.
It neither proves strong NP-hardness nor settles unweighted
maximum-cardinality complexity.
\Halmos
\end{proof}

\subsection{Complete exact family search}
\label{app:complete-family-search}

There are
\begin{equation*}
 N_{\rm pol}=\prod_{s\in\Sset}(2^{\abs{\Aset(s)}}-1)
\end{equation*}
nonempty set-valued policies. A one-action child removes one action from one
state menu and keeps
at least one action there. Algorithm \ref{alg:complete-exact-appendix} searches
this finite lattice from the full set-valued policy downward. The set $\mathcal M$ stores
the maximal safe set-valued policies found so far.

\begin{algorithm}[htbp]
\caption{Complete identification of all maximal exact-safe set-valued policies}
\label{alg:complete-exact-appendix}
\begin{algorithmic}[1]
\State $Q\gets[\menu_{\rm full}]$, $\mathrm{Seen}\gets\varnothing$,
$\mathcal M\gets\varnothing$
\While{$Q$ is nonempty}
  \State remove one set-valued policy $\menu$ from $Q$
  \If{$\menu\in\mathrm{Seen}$}
    \State \textbf{continue}
  \EndIf
  \State $\mathrm{Seen}\gets\mathrm{Seen}\cup\{\menu\}$
  \If{$\menu\preceq F$ for some $F\in\mathcal M$}
    \State \textbf{continue}
  \EndIf
  \If{$\Phi_{\menu,\Delta}$ has no real solution}
    \State $\mathcal M\gets\{F\in\mathcal M:F\not\prec\menu\}\cup\{\menu\}$
  \Else
    \State add every one-action child of $\menu$ to $Q$
  \EndIf
\EndWhile
\State \Return $\mathcal M$
\end{algorithmic}
\end{algorithm}

The search expands only unsafe set-valued policies. Once a safe policy is
found, downward closure represents all policies with smaller state menus,
so there is no reason to visit them.
The lattice is finite. With rational data, a complete real-algebraic
decision procedure also terminates on each membership query.

To prove completeness, take any maximal safe set-valued policy $F$. Every strict ancestor
of $F$ is unsafe. Starting at the full set-valued policy, remove only actions outside the
corresponding state menus of $F$.
All set-valued policies on this path are expanded until the search reaches $F$.
The rule that skips policies contained in an already safe policy cannot
remove this path. Such a skip would require a previously found safe
set-valued policy containing an ancestor of $F$ statewise, which would
contradict maximality unless that policy were $F$ itself. Hence every maximal
safe set-valued policy is returned. Any safe policy that is not maximal lies below some maximal safe policy $F$. It is skipped if $F$ is already stored and removed when $F$ is stored later. The returned set is therefore exactly $\operatorname{Max}(\Fexact)$. Downward
closure then gives
\begin{equation*}
 \Fexact
 =\bigcup_{F\in\operatorname{Max}(\Fexact)}
 \{\menu:\menu\preceq F\}.
\end{equation*}

\subsection{Exponential output}
\label{app:exponential-output}
\begin{claim}[Exponential number of maximal policies]
A known discounted model with a number of states linear in $n$ can have $\binom{n}{\lfloor n/2\rfloor}$ inclusion-maximal safe set-valued policies.
\end{claim}
\begin{proof}{Proof.}
The output construction uses a root state $s_0$, intermediate states
$s_1,\ldots,s_n$, and a terminal state. The root has one zero-reward action and
moves uniformly to these states. The terminal state is a zero-reward
absorbing state. Each intermediate state has a good action with reward zero
and a bad action with reward $-1$. Both actions move to the terminal state. Fix
any rational $\gamma\in(0,1)$ and set
\begin{equation*}
 \Delta(s_0)=\frac{\gamma k}{n},
 \qquad \Delta(s_i)=1,
 \qquad \Delta(s_{\rm term})=0.
\end{equation*}
For a set-valued policy $\menu$, let
\begin{equation*}
  \mathcal B(\menu)=\{i:\text{the bad action is enabled at }s_i\}.
\end{equation*}
The worst policy takes every enabled bad action. Hence
\begin{equation*}
 \Reg_{s_0}(\menu)=\frac{\gamma\abs{\mathcal B(\menu)}}{n},
 \qquad
 \Reg_{s_i}(\menu)=\boldsymbol 1\{i\in\mathcal B(\menu)\}.
\end{equation*}
The set-valued policy is safe exactly when
$\abs{\mathcal B(\menu)}\le k$. Every maximal safe policy includes every
good action and exactly $k$ bad actions. There are $\binom nk$ such policies. Taking $k=\lfloor n/2\rfloor$ gives
$\Theta(2^n/\sqrt n)$ maximal outputs.
\Halmos
\end{proof}

\subsection{Row optimization and attained envelopes}

The transition-row optimization is the interval-simplex step of robust dynamic programming \citep{iyengar,nilim}. We give its exchange argument to make the envelope construction explicit. Consider a feasible probability row with lower and upper bounds $\plo,\pup$.
To maximize $p^\top v$, start at $p=\plo$ with remaining mass
$c=1-\one^\top\plo$. Order coordinates so that
$v_{j_1}\ge\cdots\ge v_{j_n}$. At coordinate $j_k$, add
$\min\{c,\pup_{j_k}-\plo_{j_k}\}$ and reduce $c$ by the amount added.
The inequalities $\one^\top\plo\le1\le\one^\top\pup$ ensure that all mass can
be assigned. If another feasible row assigns movable mass to a later
coordinate while an earlier coordinate has spare capacity, moving an amount
$\varepsilon$ forward changes its objective by
$\varepsilon(v_{j_k}-v_{j_l})\ge0$. Repeating the exchange produces a row
returned by this allocation without decreasing the objective. Reversing the
coordinate order proves the minimizing rule. Ties can be resolved in any
order.

The three rectangular operators are
\begin{equation*}
 (\Bell^+v)(s)=\max_{a\in\Aset(s)}G^+_{sa}(v),\quad
 (\Bell^-v)(s)=\max_{a\in\Aset(s)}G^-_{sa}(v),\quad
 (\Bell^-_{\menu}v)(s)=\min_{a\in\menu(s)}G^-_{sa}(v).
\end{equation*}

For stochastic $p$, the map $v\mapsto p^\top v$ is monotone and satisfies
$|p^\top(v-w)|\le\norminf{v-w}$. Taking a minimum or maximum over a fixed
row set preserves these properties. Taking the finite action maximum or
minimum and multiplying continuation values by $\gamma$ therefore gives a
monotone $\gamma$-contraction for each rectangular Bellman operator.

\begin{repeattheorem}[Theorem~\ref{thm:envelope-attainment-main} (Attained value envelopes).]
Suppose $\Uset=\Uset_{\rm rect}$. There are admissible models $M^+$, $M^-$,
and $M^-_{\menu}$ such that, at every state $s$,
\begin{align*}
 V^+(s)&=V^\star_{M^+}(s)=\max_{M\in\Uset}V_M^\star(s),\\
 V^-(s)&=V^\star_{M^-}(s)=\min_{M\in\Uset}V_M^\star(s),\\
 W^-_{\menu}(s)&=W^{\menu}_{M^-_{\menu}}(s)=\min_{M\in\Uset}W_M^{\menu}(s).
\end{align*}
\end{repeattheorem}
\begin{proof}{Proof of \cref{thm:envelope-attainment-main}.}
We first compare fixed points, then construct models that attain each bound. For every admissible complete model and every vector $v$,
$\Bell^-v\le\Bell_M^\star v\le\Bell^+v$. Comparing fixed points gives
$V^-\le V_M^\star\le V^+$.

For each state-action pair, choose its upper reward and a row maximizing
$p^\top V^+$. A maximizing row exists because the row set is compact.
Rectangularity allows all of these choices in one complete model $M^+$.
Its operator satisfies
$\Bell_{M^+}^\star V^+=\Bell^+V^+=V^+$, so uniqueness gives
$V_{M^+}^\star=V^+$. Choosing each lower reward and a row minimizing
$p^\top V^-$ gives a model $M^-$ whose optimal operator fixes $V^-$.
Thus both bounds on the optimal value are attained as full vectors.

For a fixed set-valued policy, $\Bell^-_{\menu}w\le\Bell_M^{\menu}w$ gives
$W^-_{\menu}\le W_M^{\menu}$. Choose lower rewards and rows minimizing
$p^\top W^-_{\menu}$ for every enabled action, and fill disabled rows
arbitrarily. The resulting model's restricted operator fixes
$W^-_{\menu}$, so it attains that whole vector as well. The three
constructions need not choose the same model.
\Halmos
\end{proof}

\subsection{Rectangular certificates}
\begin{repeattheorem}[Theorem~\ref{thm:certificate-soundness} (Sound certificates and their feasibility floor).]
Suppose $\Uset=\Uset_{\rm rect}$.
\begin{enumerate}[label=(\roman*),leftmargin=2.2em]
\item For every set-valued policy $\menu$, the coupled and separated
bounds cover the loss,
\begin{equation*}
 \Reg(\menu)\le g^{\menu}\le B(\menu).
 \end{equation*}
\item Some set-valued policy satisfies $B(\menu)\le\Delta$ if and only if
$\Delta\ge V^+-V^-$.
\end{enumerate}
\end{repeattheorem}
\begin{proof}{Proof of \cref{thm:certificate-soundness}.}
\emph{Part (i), separated bound.}
The attained envelopes give $V_M^\star\le V^+$ and $W^-_{\menu}\le W_M^{\menu}$ for every $M$. Subtracting these inequalities gives
\begin{equation*}
 V_M^\star-W_M^{\menu}\le V^+-W^-_{\menu}=B(\menu).
\end{equation*}
Taking the componentwise maximum over models proves $\Reg(\menu)\le B(\menu)$.

\emph{Part (i), coupled bound and ordering.}
Each branch of $\mathcal H_{\menu}$ is monotone
and has Lipschitz constant at most $\gamma$ in the sup norm. The finite
maximum has the same properties, so its fixed point $g^{\menu}$ exists
and is unique by the contraction theorem \citep[Theorem~6.2.3]{puterman}.

Fix $M$ and let $\rho_M=V_M^\star-W_M^{\menu}$. At state $s$, choose an
optimal action $a'$ and an action $a$ attaining the restricted Bellman
minimum. If $a'=a$, their common reward cancels and
\begin{equation*}
 \rho_M(s)=\gamma P_M(\cdot\mid s,a')^\top \rho_M
 \le\gamma\phi^+_{sa'}(\rho_M)\le(\mathcal H_{\menu}\rho_M)(s).
\end{equation*}
If $a'\ne a$, the reward bounds, value envelopes, and nonnegative transition
probabilities give
\begin{equation*}
 \rho_M(s)\le G^+_{sa'}(V^+)-G^-_{sa}(W^-_{\menu})
 \le(\mathcal H_{\menu}\rho_M)(s).
\end{equation*}
Thus $\rho_M\le\mathcal H_{\menu}\rho_M$, which implies $\rho_M\le g^{\menu}$.
This inequality holds for every model. Taking the componentwise maximum of its left side gives $\Reg(\menu)\le g^{\menu}$, because the same model defines both values in $\rho_M$.

To compare $g^{\menu}$ with $B=B(\menu)$, first consider a branch with
$a'\ne a$. The fixed-point equations for $V^+$ and $W^-_{\menu}$ imply
$G^+_{sa'}(V^+)-G^-_{sa}(W^-_{\menu})\le B(s)$.
On a branch with $a'=a$, the action is enabled, and
\begin{align*}
 \gamma\phi^+_{sa'}(B)
 &\le\gamma\{\phi^+_{sa'}(V^+)-\phi^-_{sa'}(W^-_{\menu})\}\\*
 &\le G^+_{sa'}(V^+)-G^-_{sa'}(W^-_{\menu})\le B(s).
\end{align*}
The first inequality follows by separately bounding the two terms of
$p^\top(V^+-W^-_{\menu})$ for the same feasible $p$. The second adds the
nonnegative reward width $\rup_{sa'}-\rlo_{sa'}$. Consequently,
$\mathcal H_{\menu}B\le B$, and fixed-point comparison gives
$g^{\menu}\le B$. Maximizing $\rho_M\le g^{\menu}$ over models proves
the bound ordering in \cref{thm:certificate-soundness}.
If distinct actions $a',a$ attain the optimistic maximum and restricted
pessimistic minimum at state $s$, their off-diagonal branch equals $B(s)$
for every input vector. In particular, $g^{\menu}(s)\ge B(s)$, so equality
holds. If either extremum has more than one attaining action, a distinct
pair can be chosen. Thus strict improvement requires both extrema to be
attained uniquely by the same action. This necessary condition alone does
not guarantee strict improvement.

\emph{Part (ii), feasibility floor.}
Removing actions cannot decrease
$W^-_{\menu}$ (\cref{lem:menu-monotonicity,thm:envelope-attainment-main}). Its largest possible value is therefore attained among
singleton policies. For every deterministic policy $\sigma$, its lower
Bellman operator is bounded above by $\Bell^-$, so
$W^-_{\{\sigma\}}\le V^-$. Choose
\begin{equation*}
 \sigma^-(s)\in\argmax_{a\in\Aset(s)}G^-_{sa}(V^-).
\end{equation*}
Its lower operator fixes $V^-$, giving
$W^-_{\{\sigma^-\}}=V^-$ at all states. Thus every set-valued policy has
$B(\menu)\ge V^+-V^-$, and this one singleton policy attains equality
throughout the vector. If $\Fcert$ is nonempty, its member satisfies $V^+-V^-\le B(\menu)\le\Delta$, proving necessity. Conversely, if $V^+-V^-\le\Delta$, the singleton $\sigma^-$ has $B(\{\sigma^-\})\le\Delta$ and belongs to $\Fcert$. This proves sufficiency and completes part (ii).
\Halmos
\end{proof}

\subsection{Certified augmentation}
\begin{repeattheorem}[Theorem~\ref{thm:scalable-safety} (Safety of certified augmentation).]
If its seed checks are completed with exact evaluation, the separated construction
described above returns a policy whenever some policy satisfies $B(\menu)\le\Delta$, and only then. Every
returned policy is safe. If every remaining one-action extension fails the
exact separated test, no policy in $\Fcert$ strictly contains the returned
policy.
\end{repeattheorem}
\begin{proof}{Proof of \cref{thm:scalable-safety}.}
We verify the seed, then use monotonicity for augmentation and maximality. Set $u_\Delta=V^+-\Delta$. Every action admitted by the local rule satisfies
$G^-_{sa}(u_\Delta)\ge u_\Delta(s)$. If each menu is nonempty, taking the minimum over
its actions gives $\Bell^-_{\menu_{\rm loc}}u_\Delta\ge u_\Delta$. Fixed-point comparison
yields $W^-_{\menu_{\rm loc}}\ge u_\Delta$, so the separated loss bound is at
most $\Delta$.

If some local menu is empty and $\Fcert$ is nonempty, the floor
result gives $V^+-V^-\le\Delta$. Its attaining singleton therefore passes
the fallback test. Conversely, every accepted seed belongs to $\Fcert$.
With exact evaluation, the algorithm returns a policy exactly when
$\Fcert$ is nonempty.

Every accepted extension passes the exact separated certificate and is therefore safe. Induction over the accepted additions preserves safety.

Finally, suppose every remaining one-action extension is proved to have
$B_s>\Delta(s)$ for some state. Any strict enlargement contains one such
extension. Menu monotonicity makes its separated bound at least as large,
so it cannot belong to $\Fcert$. The final policy is therefore inclusion
maximal in that certified family. Failure of an upper bound to pass is not
proof that $B_s>\Delta(s)$, and an unresolved numerical comparison cannot
support this maximality conclusion.
\Halmos
\end{proof}

The coupled test has the same persistence under policy enlargement. If
$\menu\preceq\menu'$, then $W^-_{\menu'}\le W^-_{\menu}$, so each existing
off-diagonal branch of $\mathcal H_{\menu'}$ is at least as large as its
counterpart in $\mathcal H_{\menu}$. Diagonal branches are unchanged, and
the maximum includes every branch of the smaller policy. Thus
$\mathcal H_{\menu}g\le\mathcal H_{\menu'}g$ for every $g\ge0$.
Fixed-point comparison gives $g^{\menu}\le g^{\menu'}$. Every rejected
one-action extension therefore remains rejected after later additions,
which proves the same inclusion-maximality conclusion once every remaining
extension has been evaluated exactly and verified to fail the coupled test.

\subsection{Continuous-family certificates}
\label{app:continuous-family-proofs}

The cell operators use the discounted setting of the main text. The
finite-horizon recursions and their terminal-loss tightening are derived
separately in \cref{app:finite-cell-certificates}.

\begin{repeatproposition}[Proposition~\ref{prop:continuous-cell-safety} (Cellwise loss bounds).]
Suppose the cell enclosures contain all models $M(\theta)$, $\theta\in C$,
and the contrast bounds of \cref{eq:clinical-joint-contrast} are valid.
For every nonempty set-valued policy and every $\theta\in C$,
$0\le V^\star_\theta(s)-W^{\menu}_\theta(s)\le B^{C,\chi}(s)$
for $\chi\in\{\mathrm{raw},\mathrm{adv},\mathrm{joint}\}$, with
$B^{C,\mathrm{raw}}(s)\le B^{C,\mathrm{adv}}(s)$ and
$B^{C,\mathrm{joint}}(s)\le B^{C,\mathrm{adv}}(s)$.
\end{repeatproposition}
\begin{proof}{Proof of \cref{prop:continuous-cell-safety}.}
We first enclose values, then bound one-step losses, and finally compare
the fixed points that accumulate those losses.

\emph{Value enclosures.} The minimum and maximum row expectations are
monotone and nonexpansive in the sup norm, because every transition row is stochastic.
Adding rewards and multiplying continuation by $\gamma$ gives monotone
$\gamma$-contractions for the three cell value operators. They therefore
have unique fixed points. For each $\theta\in C$, the lower optimal
operator is bounded above by the model's optimal Bellman operator, and
the upper optimal operator is bounded below by it. Iterating these ordered
operators and taking limits gives
\begin{equation*}
 \underline V^C\le V^\star_\theta\le\overline V^C,
 \qquad \underline W^{C,\menu}\le W^{\menu}_\theta.
\end{equation*}
The latter comparison uses the restricted minimum over the same menu.
Since that minimum is bounded above by the lower optimal maximum,
$\underline W^{C,\menu}\le\underline V^C$. Subtracting proves the
nonnegative raw bound. Substituting the optimal-value enclosures into
each action gives $\underline Q^C_{sa}\le Q^\star_\theta(s,a)
\le\overline Q^C_{sa}$, where $Q^\star_\theta(s,a)=r_\theta(s,a)+\gamma P_\theta(\cdot\mid s,a)^\top V^\star_\theta$.

\emph{One-step losses.} The advantage bound consequently satisfies
$\overline\ell^{C,\mathrm{adv}}_{sa}\ge\ell_\theta(s,a)\ge0$.
An excluded rival has an upper action value strictly below another
action's lower value, so it cannot be optimal in any model in the cell.
The rival set is nonempty and contains every optimal action for the fixed
model $M(\theta)$. Hence
\begin{equation*}
 \ell_\theta(s,a)=\max_{a'\in R_s^C}D_\theta(s;a',a)
 \le\max\!\left(0,\max_{a'\in R_s^C}\overline D^C_{s,a',a}\right).
\end{equation*}
Taking the minimum with the advantage bound preserves validity. This
establishes $0\le\ell_\theta\le\overline\ell^{C,\mathrm{joint}}
\le\overline\ell^{C,\mathrm{adv}}$. A cell contrast must enclose the
entire cell. Evaluating only parameter-box corners is insufficient in
general. When a state's available action set is a singleton, its sole
contrast is zero. A singleton menu within a larger available action set
need not have zero one-step loss.

\emph{Cumulative loss and the joint comparison.} Subtracting the restricted
Bellman equation from the optimal value and adding and subtracting the
optimal continuation gives
\begin{equation*}
 \rho_\theta(s)=\max_{a\in\menu(s)}
 \{\ell_\theta(s,a)+\gamma P_\theta(\cdot\mid s,a)^\top\rho_\theta\},
 \qquad \rho_\theta=V^\star_\theta-W^{\menu}_\theta\ge0.
\end{equation*}
Both cell loss operators are monotone contractions and map nonnegative
vectors to nonnegative vectors. Their fixed points are nonnegative by
iteration from zero. Enclosing the local loss and transition row gives
$\rho_\theta\le\mathcal H^{C,\chi}_{\menu}\rho_\theta$.
Repeated application preserves this inequality and converges to
$B^{C,\chi}$, proving soundness for both loss certificates. The ordering
of the local bounds gives ordered operators. Fixed-point comparison
therefore yields $B^{C,\mathrm{joint}}\le B^{C,\mathrm{adv}}$.

\emph{The raw comparison.} Suppress the cell index and put
$L=\underline V^C$, $U=\overline V^C$, $Z=\underline W^{C,\menu}$,
so $Z\le L\le U$ and $B^{C,\mathrm{raw}}=U-Z$. For each state,
\begin{align*}
 U(s)-Z(s)
 &=\max_{a\in\menu(s)}\{U(s)-\rlo^C_{sa}
                          -\gamma\underline\phi^C_{sa}(Z)\}\\
 &\le\max_{a\in\menu(s)}\{
 \overline\ell^{C,\mathrm{adv}}_{sa}
 +\gamma[\underline\phi^C_{sa}(L)-\underline\phi^C_{sa}(Z)]\}\\
 &\le\max_{a\in\menu(s)}\{
 \overline\ell^{C,\mathrm{adv}}_{sa}
 +\gamma\overline\phi^C_{sa}(U-Z)\}.
\end{align*}
The first inequality uses the definition of the advantage bound. For
the second, let a row attain the minimum at $Z$ and evaluate the minimum
at $L$ using that row. This gives
$\underline\phi(L)-\underline\phi(Z)\le\overline\phi(L-Z)
\le\overline\phi(U-Z)$. Thus
$B^{C,\mathrm{raw}}\le\mathcal H^{C,\mathrm{adv}}_{\menu}
B^{C,\mathrm{raw}}$. Iterating this contraction proves the remaining
ordering.\Halmos
\end{proof}

\begin{claim}[Raw and joint bounds are not ordered]
Neither certificate is uniformly smaller than the other over valid cell enclosures.
\end{claim}
\begin{proof}{Proof.}
Both examples use $\theta\in[0,10]$, $\gamma=1/2$, known transitions,
and a zero-reward absorbing state. First, a state with one action of
reward $\theta$ that immediately enters the absorbing state has raw bound
$10$ and joint bound $0$.

For the reverse ordering, state $x$ has zero-reward actions $a,a'$ leading
to states $u,v$, respectively. Actions at $u$ pay $0$ or $\theta$ and
then absorb. The sole action at $v$ pays $\theta$ and then absorbs.
The policy retains $a$ at $x$ and the zero-reward action at $u$. At $u$,
both bounds are $10$. The true optimal values at $u$ and $v$ are both
$\theta$, and their separate enclosures are $[0,10]$. The raw bound at
$x$ is $5$. The valid contrast construction in the main text permits
these two continuation coordinates to vary independently within their
enclosures, giving a contrast bound of $5$. The advantage action-loss
bound is also $5$, so the joint local bound is $5$. Adding continuation
loss $\gamma\times10$ gives joint bound $10$ at $x$. The true worst
loss there is $5$. \Halmos
\end{proof}

\begin{repeatproposition}[Proposition~\ref{prop:clinical-cover-safety} (Safety on a complete parameter cover).]
For a complete cover with valid cell bounds,
$\sup_{\theta\in\Theta}\{V^\star_\theta(s)-W^{\menu}_\theta(s)\}
\le B^{\mathcal C,\chi}(s)$.
\end{repeatproposition}
\begin{proof}{Proof of \cref{prop:clinical-cover-safety}.}
Fix any parameter $\theta\in\Theta$, including a boundary point. Some
cell $C$ contains it. Proposition~\ref{prop:continuous-cell-safety}
bounds its loss at every state by $B^{C,\chi}$, hence by the maximum over
the finite cover. Taking the supremum over $\theta$ proves the result.
Checking this bound after each addition preserves safety throughout
construction.

Refinement replaces a parent cell by children whose union equals the
parent. For a factor interval $[a,b]$, the intervals $[a,m]$ and $[m,b]$
with $a<m<b$ preserve coverage, including the shared face. A parent's
bound remains valid on either child. Taking the minimum of it and any
independently valid child bound preserves validity. No region may be
omitted because its bound is difficult to compute. Computation limits
affect what can be certified, while every accepted certificate still
covers the full domain.\Halmos
\end{proof}

\subsection{Complete covers through a risk score}
\label{app:risk-score-cover}

The next lemma justifies the change of regions in
\cref{sec:clinical-cover-search}. It uses the same cellwise loss bound as
\cref{prop:continuous-cell-safety}. The score intervals organize the
original models and do not define new independent risks at each epoch.

\begin{lemma}[Risk-score cover]
\label{lem:risk-score-cover}
Consider a family with fixed admissible actions, treatment responses,
and rewards. Suppose its uncertain transition inputs are monotone
functions of continuous epoch scores $\psi_t(\theta)$. Fix a reference
epoch $r$. Let a compact parameter box $B\subseteq\Theta$ have a finite cover by compact
subboxes $B_j\subseteq B$. For every subbox, let the closed intervals
$R_j$ and $D_{tj}$ satisfy
\[
 \psi_r(B_j)\subseteq R_j,
 \qquad
 (\psi_t-\psi_r)(B_j)\subseteq D_{tj}.
\]
For a closed score interval $I$, define
\[
 F_{B,I}=\{\theta\in B\mid \psi_r(\theta)\in I\},
 \qquad
 J_{B,I}=\{j\mid R_j\cap I\ne\varnothing\}.
\]
When $J_{B,I}$ is nonempty, the interval
\[
 E_t(B,I)=\operatorname{hull}_{j\in J_{B,I}}
            \bigl((I\cap R_j)+D_{tj}\bigr)
\]
contains $\psi_t(F_{B,I})$. If $J_{B,I}$ is empty, then $F_{B,I}$ is
empty. If finitely many boxes cover $\Theta$ and each box's finitely many score intervals cover
$\psi_r(B)$, the nonempty sets $F_{B,I}$ form a complete compact cover
of $\Theta$. Valid cellwise loss bounds on these sets therefore certify
the same loss limit over the original family.
\end{lemma}
\begin{proof}{Proof.}
Fix $\theta\in F_{B,I}$. Some subbox $B_j$ contains $\theta$, so
$\psi_r(\theta)\in I\cap R_j$ and $j\in J_{B,I}$. The identity
\[
 \psi_t(\theta)=\psi_r(\theta)
                  +\bigl(\psi_t(\theta)-\psi_r(\theta)\bigr)
\]
places $\psi_t(\theta)$ in $(I\cap R_j)+D_{tj}$ and hence in
$E_t(B,I)$. This argument also shows that an empty $J_{B,I}$ rules out
every point of $F_{B,I}$. Continuity makes each $F_{B,I}$ closed in the
compact box $B$, so it is compact.

Every $\theta\in\Theta$ belongs to a covered box and has a reference
score in one of that box's covered intervals. Thus it belongs to a
retained region. Applying the monotone risk functions to the score
enclosures gives valid transition-input enclosures on that region.
The transition rows computed from these enclosures and valid contrast bounds then satisfy
the hypotheses of \cref{prop:continuous-cell-safety}. Maximizing the
resulting bounds over the complete cover proves the claim by
\cref{prop:clinical-cover-safety}. The interval calculations may enclose
risk combinations that no original parameter attains. Such enlargement
can weaken the bound, but it does not invalidate it.\Halmos
\end{proof}

A score interval must remain in the cover until its preimage is proved
empty or covered by replacement intervals. A bound that fails to certify
a candidate does not justify removing that interval. The baseline epoch
in the clinical implementation is fixed and is evaluated directly.
Outward rounding must enclose each interval in the lemma. Additional
parameter-dependent treatment effects or admissibility require their
own enclosures before this argument can be used.

\subsection{Nesting of rectangular uncertainty sets}
\label{app:rectangular-nesting}

Let $M^0=(r^0,P^0)$ belong to the full rectangular set. At width
$\lambda\in[0,1]$, interpolating every endpoint toward this nominal model gives
\begin{align*}
 \rlo^\lambda_{sa}&=(1-\lambda)r^0_{sa}+\lambda\rlo_{sa},
 &\rup^\lambda_{sa}&=(1-\lambda)r^0_{sa}+\lambda\rup_{sa},\\
 \plo^\lambda_{sa,j}&=(1-\lambda)P^0_{sa,j}+\lambda\plo_{sa,j},
 &\pup^\lambda_{sa,j}&=(1-\lambda)P^0_{sa,j}+\lambda\pup_{sa,j}.
\end{align*}
The nominal model belongs to every contracted set. In particular,
\begin{equation*}
 \sum_j\plo^\lambda_{sa,j}=(1-\lambda)+\lambda\sum_j\plo_{sa,j}\le1,
 \qquad
 \sum_j\pup^\lambda_{sa,j}=(1-\lambda)+\lambda\sum_j\pup_{sa,j}\ge1,
\end{equation*}
so each transition box intersects the simplex. Coordinatewise monotonicity
of the endpoints gives
\begin{equation*}
 \{M^0\}=\Uset_0\subseteq\Uset_{\lambda_1}
 \subseteq\Uset_{\lambda_2}\subseteq\Uset_1,
 \qquad 0\le\lambda_1\le\lambda_2\le1.
\end{equation*}
Worst-case loss cannot increase when the model set shrinks, so a policy safe for a wider set is safe for every contained set. Width measures interpolation distance. A statistical coverage probability requires separate calibration.
\subsection{Rectangular shared-advantage refinement}
\label{app:rectangular-advantage}

The shared-advantage variant combines the coupled operator with a bound on
the loss from each action. Using the rectangular envelopes from the main
text, define
\begin{align*}
 \bar\ell_{sa}&=\max\{0,V^+(s)-G^-_{sa}(V^-)\},\\
 (\mathcal A_{\menu}g)(s)&=\max_{a\in\menu(s)}
       \{\bar\ell_{sa}+\gamma\phi^+_{sa}(g)\},\\
 \widetilde{\mathcal H}_{\menu}g
       &=\min\{\mathcal H_{\menu}g,\mathcal A_{\menu}g\},
\end{align*}
where the last minimum is componentwise. The value envelopes imply that
$\bar\ell_{sa}$ bounds the loss from taking $a$ and then following the
optimal policy in every admissible model. For
$\rho_M=V_M^\star-W_M^{\menu}$, the Bellman loss identity therefore gives
$\rho_M\le\mathcal A_{\menu}\rho_M$. The coupled proof also gives
$\rho_M\le\mathcal H_{\menu}\rho_M$. The scalar inequality
\begin{equation*}
 |\min\{a,b\}-\min\{c,d\}|\le\max\{|a-c|,|b-d|\}
\end{equation*}
shows that their minimum is a monotone $\gamma$-contraction, so its unique fixed point bounds $\rho_M$ for every model.
Moreover, $\widetilde{\mathcal H}_{\menu}g^{\menu}\le\mathcal H_{\menu}g^{\menu}=g^{\menu}$, so fixed-point comparison places that fixed point below $g^{\menu}$.

\subsection{Detailed construction procedures}
\label{app:construction-procedures}
The following procedure supplies the rectangular seed and augmentation
steps of the construction in \cref{sec:common-procedure,sec:master-construction}.
Its guarantees concern the separated certificate.

\begin{algorithm}[!t]
\caption{Certified augmentation for rectangular uncertainty}
\label{alg:scalable-main}
\begin{algorithmic}[1]
\Require rectangular model set, loss limits $\Delta$, exact evaluation or valid enclosures
\State compute a verified upper bound for $V^+$
\State form the local menus using a target at least $V^+-\Delta$
\State retain an action only when its local admission inequality is verified
\If{a local menu is empty}
 \State compute $V^-$ and propose its pessimistic-greedy singleton
 \If{the singleton is verified to satisfy $B\le\Delta$}
  \State use the singleton as the seed
 \ElsIf{$V^+-V^->\Delta$ is verified at some state}
  \State \Return \textit{no separated certificate}
 \Else
  \State \Return \textit{seed unresolved}
 \EndIf
\EndIf
\State store the verified seed as $\menu$
\State $\mathcal X\gets\{(s,a)\in\mathcal K:a\notin\menu(s)\}$
\While{untested or unresolved extensions remain and computation permits}
 \State evaluate the one-action extensions in $\mathcal X$
 \If{no extension is certified to satisfy $B\le\Delta$}
  \State \textbf{break}
 \EndIf
 \State accept the passing extension with largest minimum margin
 \State remove its action and extensions verified to fail from $\mathcal X$
\EndWhile
\State \Return the certified policy $\menu$
\end{algorithmic}
\end{algorithm}

\subsubsection{Complete-cover certification and refinement.}
\label{app:continuous-protocol}
The parametric-family construction separates checking a fixed policy from
proposing seeds and action additions. Algorithm~\ref{alg:family-verify}
bounds the loss of a fixed policy on a complete cover.
Algorithm~\ref{alg:family-search} uses that test in fixed-order augmentation.

\begin{algorithm}[htbp]
\caption{Certify one parametric-family policy on a complete cover}
\label{alg:family-verify}
\begin{algorithmic}[1]
\Require Family, fixed policy $\menu$, certificate type $\chi$, loss limit, complete cover, and computation limits
\State Build valid cell enclosures and evaluate the certificate on every cell
\While{the complete-cover bound exceeds the limit and computation permits}
  \If{a checked model gives a loss above the limit}
    \State \Return rejection of this policy with the violating model
  \EndIf
  \State Select a cell attaining the largest unresolved loss bound
  \State Split its parameter or score interval into children covering the parent
  \State Enclose each child and cap its bound by the valid parent bound
  \State Check complete coverage and reevaluate the fixed policy
\EndWhile
\If{the complete-cover bound is within the limit at every position}
  \State \Return certified policy, complete cover, and loss bound
\Else
  \State \Return unresolved comparison, loss bound, and stop reason
\EndIf
\end{algorithmic}
\end{algorithm}

A cell remains in the cover when finite endpoint precision prevents an
interior split. Every refinement preserves all parameter points, including
boundaries. Score intervals use the enclosures of
Lemma~\ref{lem:risk-score-cover}. Parameter boxes use their coordinate intervals.

\begin{algorithm}[htbp]
\caption{Parametric-family proposals and fixed-order augmentation}
\label{alg:family-search}
\begin{algorithmic}[1]
\Require Family, certificate type, loss limit, complete cover, and computation limits
\State Propose singleton seeds from point-model optima, finite-model minimax proposals,
minimum worst local-loss choices, and bounded single-position repairs
\State Check each distinct seed using Algorithm~\ref{alg:family-verify}
\If{no seed passes}
  \State Analyze policy existence on a finite list of models from the family
  \State Add checked violating models to the list and use it to propose further seeds while computation permits
  \If{a checked infeasibility proof is obtained}
    \State \Return infeasibility certificate
  \ElsIf{no seed passes}
    \State \Return unresolved outcome with existence bounds and stop reason
  \EndIf
\EndIf
\State Initialize the incumbent with a certified policy and its complete cover
\State Order missing actions once by increasing largest cellwise action-loss bound,
then decision epoch, state, and action index
\For{each missing action, while computation permits}
  \State Add that action to the incumbent and apply Algorithm~\ref{alg:family-verify}
  \State Accept the addition only if its complete-cover bound is within the limit
\EndFor
\State Check the final incumbent and output its certificate and enlargement stop reason
\end{algorithmic}
\end{algorithm}

Point-model optima include the nominal model and parameter-domain corners.
A finite-model minimax proposal seeks a common policy minimizing the largest
loss over a finite list of models.
The minimum worst local-loss policy chooses at each position an action
minimizing its largest cellwise one-step loss bound. Bounded single-position
repairs change one action of a candidate before another full-family check.
An available certified policy can initialize the incumbent directly.
The order of missing actions is fixed when augmentation starts and each
action is considered once. A certified incumbent is preserved when an
addition remains unresolved or computation ends. Its action count is a
lower bound on the largest feasible count.

\subsection{Numerical Acceptance and Computational Cost}
\label{app:numerical-cost}
\begin{remark}[Verified numerical construction]
For a numerical implementation, the same proof applies to any vector
$\bar u\ge V^+-\Delta$ for which every admitted action is verified to
satisfy $G^-_{sa}(\bar u)\ge\bar u(s)$. Then
$W^-_{\menu_{\rm loc}}\ge\bar u\ge V^+-\Delta$. A numerical singleton
proposal is also accepted only after its loss bound is verified. If this
comparison remains unresolved, no seed is returned. Each subsequent
acceptance uses a valid upper bound on the enlarged policy's loss, so
induction over the accepted additions proves safety of every stored policy.

\end{remark}

Numerical acceptance requires bounds on both iteration error and error in
the implemented calculation. For an exact $\gamma$-contraction $T$ with
fixed point $x^\star$, an implemented map $\widehat T$, and an iterate $x$,
\begin{equation}
 \norminf{x-x^\star}\le
 \frac{\norminf{x-\widehat T x}+\norminf{\widehat T x-Tx}}{1-\gamma}.
 \label{eq:implemented-residual}
\end{equation}
The first term measures the difference between the iterate and its
computed backup. We need the second term to bound backup errors, including
transition-row optimization and arithmetic rounding. A small observed residual, even
zero, is insufficient alone. Upper bounds on both terms, their sum, and the division give a valid acceptance bound.
We subtract it from a policy value and add it to a benchmark. Bellman lower and upper inequalities provide another acceptance test.

For \cref{eq:implemented-residual}, let $x^\star=Tx^\star$. The contraction
and triangle inequalities give
\begin{equation*}
 \norminf{x-x^\star}
 \le\norminf{x-Tx}+\norminf{Tx-Tx^\star}
 \le\norminf{x-Tx}+\gamma\norminf{x-x^\star}.
\end{equation*}
Also,
$\norminf{x-Tx}\le\norminf{x-\widehat T x}+\norminf{\widehat T x-Tx}$.
Subtract the contraction term and divide by $1-\gamma>0$. The implemented
map is required only to have a valid upper bound on its error at the
particular iterate $x$.

In the rectangular implementation, exact rational arithmetic evaluates the
final Bellman backup and residual. We divide the residual by $1-\gamma$
exactly and round the value bounds outward. For small
instances, we can recover a fixed point by rational policy evaluation and
accept it after checking its Bellman equations exactly. This calculation
resolves some policies whose loss equals the limit. Probability rows must define a nonempty set of distributions. The verifier
records any endpoint correction needed to enclose a normalized row and
certifies the resulting family.

One dense transition-row sweep sorts a common continuation vector once and scans all
$m$ transition rows, at a cost of $D(n,m)=n\log n+mn$. Let $K_+$, $K_-$, and $K_{\menu}$ count sweeps for
$V^+$, $V^-$, and $W^-_{\menu}$. Let $K_H$ count coupled sweeps. There are at most
$N_{\rm add}=m-n$ optional additions, and a coupled sweep compares
$J(\menu)=\sum_s\abs{\Aset(s)}\abs{\menu(s)}$ action pairs. The bounds in
\cref{tab:scalable-complexity-main} separate these iteration and search costs.

\begin{table}[hbp]
\centering\small
\caption{Arithmetic-operation bounds for dense rectangular certificates.
$K_{\max}$ bounds the number of Bellman sweeps for a policy during construction. Shared
benchmark calculations are performed once.}
\label{tab:scalable-complexity-main}
\begin{tabularx}{\linewidth}{@{}L{0.31\linewidth}Y@{}}
\toprule
Component & Time bound \\
\midrule
Local construction & $O((K_++1)D(n,m)+m)$ \\
Singleton fallback and check & $O((K_-+K_{\menu}+2)D(n,m)+n)$ \\
One separated check & $O((K_{\menu}+1)D(n,m)+n)$ \\
Fixed-order augmentation & $O((N_{\rm add}+1)(K_{\max}+1)D(n,m))$ \\
Best-margin rescanning & $O((N_{\rm add}+1)^2(K_{\max}+1)D(n,m))$ \\
One coupled refinement & $O((K_{\menu}+1)D(n,m)+(K_H+1)[D(n,m)+J(\menu)])$ \\
\bottomrule
\end{tabularx}
\end{table}

When needed, the fallback adds a pessimistic-optimal value calculation and
a singleton check. Fixed-order augmentation tests at most $N_{\rm add}+1$
policies, including its seed.
Best-margin rescanning can test $1+N_{\rm add}(N_{\rm add}+1)/2$ policies. Suppose $N_{\rm check}$ separated checks use counts $K_e$, including the
singleton check when used. Let $I_{\rm seed}$ equal one when the fallback
is used and zero otherwise.
The local pass, fallback, and separated checks together cost
\begin{equation*}
 O\!\left(\left[K_++1+I_{\rm seed}(K_-+1)+\sum_{e=1}^{N_{\rm check}}(K_e+1)\right]D(n,m)+m+N_{\rm check}\right).
\end{equation*}
Each coupled evaluation adds its tabled cost. Dense working memory is
$O(mn+m+n)$ if action-pair branches are scanned without storing them.
Storing all pairwise constants adds $O(J(\menu))$ memory. We count arithmetic operations for value-iteration sweeps.
Policy-iteration variants also solve linear systems. Exact or
interval arithmetic also incurs the cost of its chosen precision. For initial error
at most $E_0>0$ and target error $\tau>0$, ordinary contraction iteration needs
at most
$\max\{0,\lceil\log(E_0/\tau)/(-\log\gamma)\rceil\}$ sweeps when
$0<\gamma<1$. One backup suffices at $\gamma=0$.

\FloatBarrier
\section{Additional numerical results}
\label{app:extra-results}

Our experiments examine agreement with exact solutions, flexibility at loss-limit boundaries, construction time as state count grows, and uncertainty width.
\subsection{Arithmetic checks and timing conventions}
\label{app:experiment-arithmetic-timing}

Rectangular policies are checked independently through exact model-relative loss evaluation or exact-rational Bellman residuals. The residual calculation uses \cref{eq:implemented-residual} to enclose the fixed point. At an exact boundary, rational policy evaluation can establish equality without adding a numerical acceptance tolerance. For recorded point-model probability rows whose binary64 entries do not sum to one exactly, the checker normalizes the row in rational arithmetic and records the change. No such normalization is needed for the dyadic transition rows in the sepsis width experiment.

The reference comparison in Table~\ref{tab:compact-comparison} contains five sizes, $(n,|A|)\in\{(1,2),(1,3),(1,4),(2,2),(2,3)\}$, with two generated MDPs at each size and three loss limits. The runtime experiment contains eight state counts, $1,2,4,8,16,32,64,128$, two or four actions per state, and two replications. Nominal rewards are sampled from $[0,1]$. For a nominal coordinate $c$ and uncertainty fraction $\eta$, the interval endpoints are $(1-\eta)c$ and $(1-\eta)c+\eta$. Transition intervals are intersected with the probability simplex. Both experiments use $\eta=0.25$ for rewards and transitions, and discount factor $0.5$. The reference loss limits are $0.5$, $1$, and $1.5$.

\begin{table}[!hbp]
\centering\small
\setlength{\tabcolsep}{4pt}
\caption{Reference comparison on 30 instance--limit pairs. Certified columns count policies passing the stated certificate, and matches columns count policies attaining the exact maximum action count. Coupled census exhaustively checks the same policy family.}
\label{tab:compact-comparison}
\begin{tabular}{rrrrrrrrrrr}
\toprule
States & Actions & Cases & \multicolumn{2}{c}{Local} & \multicolumn{2}{c}{Best-margin} & \multicolumn{2}{c}{Coupled census} & \multicolumn{2}{c}{Median seconds} \\
 & per state & & Certified & Matches & Certified & Matches & Certified & Matches & Exact & Local \\
\midrule
1 & 2 & 6 & 6 & 6 & 6 & 6 & 6 & 6 & 0.0052 & 0.0015 \\
1 & 3 & 6 & 6 & 6 & 6 & 6 & 6 & 6 & 0.0154 & 0.0022 \\
1 & 4 & 6 & 6 & 6 & 6 & 6 & 6 & 6 & 0.0979 & 0.0028 \\
2 & 2 & 6 & 4 & 3 & 4 & 4 & 6 & 6 & 0.0928 & 0.0029 \\
2 & 3 & 6 & 4 & 4 & 4 & 4 & 6 & 6 & 0.6969 & 0.0047 \\
\bottomrule
\end{tabular}
\end{table}

We evaluate the coupled certificate for all 166 policies across the ten model families at the three loss limits. It uses the same coupled certificate and arithmetic implementation as the boundary census below. The coupled bounds and all 30 maximum-count selections have independent checks. The independent exact membership formulation verifies all 30 selected policies. 

The reference and runtime timing measurements use ARM64 macOS, Python 3.13.11, NumPy 2.4.4, and a Bellman stopping tolerance of $10^{-11}$. The local-construction times measure the floating-point stage. They exclude exact-rational value enclosures, boundary calculations, and independent policy checks. The reported certified counts use the validated calculations, which can resolve exact-boundary cases left undecided by the floating-point stage. Exact-reference times include the finite-lattice search and its algebraic membership decisions. Exact computation takes between 0.0046 and 5.27 seconds across the reference problems. Floating-point local construction takes between 0.0015 and 0.0050 seconds. All 32 runtime instances have certified policies. At 128 states and four actions per state, the median floating-point local-construction time is 0.437 seconds.

The runtime experiment uses loss limit 1 and measures floating-point computation of the upper value envelope and local action-admission tests. Figure~\ref{fig:reference-construction-time} shows both replications at each size and their medians.

\begin{figure}[!htbp]
\centering
\includegraphics[width=.82\linewidth]{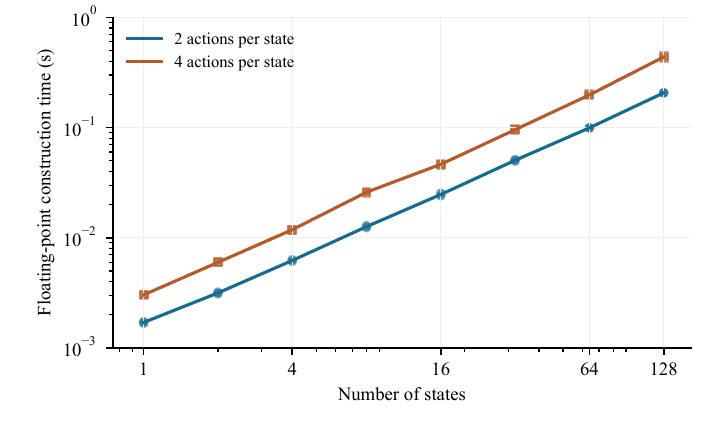}
\caption{Floating-point local-construction times for eight state counts, two or four actions per state, and two replications. Points show individual measurements and lines join medians. These times exclude exact-rational enclosures and independent checks.}
\label{fig:reference-construction-time}
\end{figure}
\subsection{Comparison of exact formulations}
\label{app:exact-formulation-results}

Two independent exact encodings were applied to 730 policies from 68 small
interval MDPs (Table~\ref{tab:backend-audit-appendix}). The instances and policies were specified before the comparison. Each loss limit is set so that the correct answer is known in advance.
A separate branch-and-bound computation first brackets the policy's loss.
A margin of 0.001 to 0.004, four times that computation's tolerance,
separates the limit from the bracket. When the bracket's lower end exceeds
the margin, the limit sits that margin below the bracket, so the correct
answer is unsafe. Otherwise the limit sits the same margin above the
bracket, so the correct answer is safe. This gives 711 unsafe tests and
19 safe tests.
The compact encoding uses one shared Bellman
formula from \cref{thm:compact-membership}. The reference encoding explicitly
lists every pair formed by an optimal deterministic policy and a deterministic
compliant policy. It then solves one shared-model formula for each pair. Both
encodings receive five seconds per policy. A timeout receives the label unknown
and contributes no safe or unsafe conclusion.

\begin{table}[htbp]
\centering
\caption{Comparison of the two exact formulations on 730 policies.}
\label{tab:backend-audit-appendix}
\begin{tabular}{lrr}
\toprule
Outcome & compact formula & enumerated pairs \\
\midrule
Conclusive safe & 19 & 19 \\
Conclusive unsafe & 690 & 690 \\
Timeout or unknown & 21 & 21 \\
Conclusive disagreements & \multicolumn{2}{c}{0 among 709 decisions} \\
Verified unsafe witnesses & 690/690 & 690/690 \\
Median total time, seconds & 0.0033 & 0.0081 \\
Maximum observed time, seconds & 5.2088 & 5.6464 \\
\bottomrule
\end{tabular}
\end{table}

Every conclusive result of both encodings matches the known answer.
All 21 unresolved cases come from the same three-state control instance,
in which the available actions vary by state. Both encodings time out
on the same policies. All 21 are unsafe tests. Each bracket is narrower
than $10^{-10}$, and a model in the family attains a loss above the limit. The compact formula uses 8 to 30 real variables and 22 to
73 top-level constraints in this comparison. The equivalent enumerated family has
up to 81 formulas. Every satisfying assignment describes a model that makes
the policy unsafe. Each witnessing MDP is checked by direct linear policy
evaluation.
Exact rational models witnessing the 21 inconclusive cases satisfy family membership and the required loss inequality.

Only 19 policies receive a conclusive safe result, while 690 receive a verified
unsafe witness. The comparison mainly tests whether the two encodings detect unsafe policies.
The 30-pair exact-reference experiment in the main paper separately
rechecks every returned safe policy.

\subsection{Boundary census}

The boundary experiment contains 68 MDPs, 730 policies, and
213 loss limits (Table~\ref{tab:boundary-census-appendix}). These limits come from the exhaustive reference's upper loss bounds at successive action-count thresholds. Duplicate numerical limits are retained when they correspond to different thresholds.

The direct loss-bound census enumerates the set-valued policies and keeps
those whose computed upper bound on worst-case model-relative loss is at
most the loss limit in every state. Its row reports the largest enabled-action
count among those policies. The coupled and separated census rows
apply their respective validated sufficient tests to the same enumerated
policies. The census evaluates all 730 policies with exact-rational residuals
and outward value bounds. Local construction and best-margin augmentation build a policy without
enumerating the full family.

\begin{table}[htbp]
\centering\small
\caption{Action-count comparisons at 213 boundary loss limits. Matches attain the direct loss-bound reference count. For the shortfall summaries, an unresolved case receives an action-count score of zero.}
\label{tab:boundary-census-appendix}
\begin{tabular}{lrrrr}
\toprule
Method & Matches & Not certified & Mean shortfall & Max. shortfall \\
\midrule
Direct loss-bound census & 213/213 & 0 & 0.000 & 0 \\
Coupled certificate census & 170/213 & 27 & 0.296 & 3 \\
Separated certificate census & 166/213 & 35 & 0.376 & 4 \\
Local construction & 144/213 & 35 & 0.484 & 4 \\
Best-margin augmentation & 166/213 & 35 & 0.376 & 4 \\
\bottomrule
\end{tabular}
\end{table}

Best-margin augmentation improves the local construction at 23 breakpoints. Across all 213 breakpoints, it matches exhaustive evaluation of the separated certificate, including the same 35 limits with no separated certificate. Where that certificate admits a policy, augmentation attains its largest action count. For these instances, certificate conservatism accounts for the gap from the reference count.

The local construction, best-margin augmentation, and exhaustive separated test
each return a policy at 178 of 213 breakpoints. The separated certificate
family is proved empty at the other 35. The coupled census returns a policy
at 186 breakpoints. The direct loss-bound census returns a nonempty policy
at every breakpoint, so the absence of a sufficient certificate at a breakpoint reflects the
limits of the certificate. Zero scores in the table record the absence of a
certificate and contribute the full exact action count to the reported shortfall.

There can be several incomparable maximal policies. Two policies are
incomparable when neither contains the other. Across 2,988 combinations of a model family and loss limit,
there are 3,036 maximal-policy occurrences. Multiple
incomparable policies occur in 382 combinations. At 42 action-count thresholds,
multiple policies attain the smallest direct upper loss bound. A largest action count and a complete
family of maximal policies are therefore different outputs.

\subsection{Numerical checks of the complexity constructions}
\label{app:complexity-checks}

Numerical checks reproduce the exponential construction's predicted maximal-policy counts at all nine tested sizes. Exhaustive solutions agree for all 45 knapsack instances and all 279 Max-Cut thresholds. 

\subsection{Instrumented runtime check}
\label{app:instrumented-scaling}

Instrumented measurements over six state counts reproduce the computational cost ordering in \cref{tab:scalable-complexity-main}. Local construction and one whole-policy check remain faster than fixed-order augmentation by state and action index.  These finite-range measurements use one instance per size and do not establish asymptotic exponents.

\subsection{Sepsis construction and width-path detail}
\label{app:sepsis-width-detail}

The synthetic sepsis model uses the public Gumbel-Max simulator associated
with \citet{oberst}. The source has 720 observed states and
eight actions formed by combinations of antibiotics, ventilation, and
vasopressors. We use its 144-state projection, which removes the five-level
glucose coordinate, and retain ten diabetes-by-current-glucose completions.
Their coordinatewise hull supplies the reward and transition intervals. The
discount factor is $\gamma=0.99$.

Transition rows are projected to a common $2^{-52}$ dyadic grid by largest
remainder. The largest source normalization correction is
$3.331\times10^{-16}$, and the largest dyadic correction is
$1.804\times10^{-16}$. Exact rational verification checks all 11,520 rows and
their box-simplex sets.

The width-path loss limit is selected from the local admission thresholds for
 the full uncertainty family. Let $d_{sa}$ be the smallest uniform loss limit
 that admits action $a$ at state $s$ under the local rule. The threshold for a
 nonempty local set at every state is $\max_s\min_a d_{sa}=89.1117918275$,
 and the threshold for retaining every action is $\max_{s,a}d_{sa}=186.2190220256$.
 We choose a loss limit 8\% of the way between these thresholds,
 \begin{equation*}
  \delta=89.1117918275+0.08(186.2190220256-89.1117918275)
  =96.8803702433.
 \end{equation*}
 This diagnostic loss limit is held fixed as $\lambda$ varies. It uses the
 simulator's reward scale and is separate from the nine-case sepsis comparison
 at loss limits $0.1$, $0.5$, and $1$. The nominal model is the dyadically
 normalized mean of the ten complete simulator models. Starting at
 $\lambda=1$, fixed-order augmentation uses the certified policy from the previous
 wider family as its next seed. The augmented recommendation sets therefore
 expand as the uncertainty family contracts (Table~\ref{tab:sepsis-width-appendix}).

At full width, the upper and lower optimal values differ by up to $\max_s\{V^+(s)-V^-(s)\}\approx26.41$, so, by Theorem~\ref{thm:certificate-soundness}(ii), no separated certificate exists below that limit. On the ten models themselves, the optimal policy of the nominal model loses less than 4.45 at every state. The hull allows combinations of rewards and transitions absent from the ten models. The certificates of \cref{sec:continuous-certificates} instead preserve dependence through shared parameters.

\begin{figure}[t]
\centering
\includegraphics[width=\linewidth]{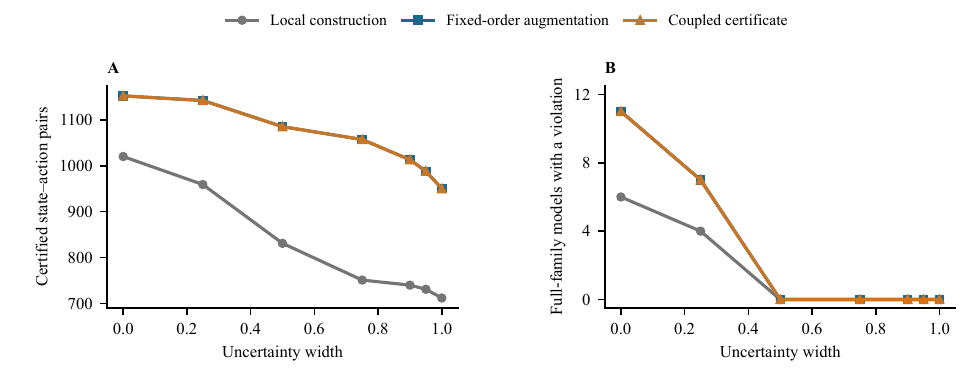}
\caption{Sepsis width-path comparisons. Panel A gives certified state-action pairs as interval width increases. Panel B counts loss-limit violations among twelve full-width models. Curves distinguish local construction, fixed-order augmentation, and the coupled certificate.}
\label{fig:sepsis-width}
\end{figure}

\begin{table}[htbp]
\centering
\caption{Enabled state-action pairs along the nested sepsis width path.}
\label{tab:sepsis-width-appendix}
\begin{tabular}{rrr}
\toprule
$\lambda$ & local & fixed-order \\
\midrule
0    & 1,020 & 1,152 \\
0.25 &   959 & 1,142 \\
0.50 &   831 & 1,085 \\
0.75 &   751 & 1,057 \\
0.90 &   740 & 1,013 \\
0.95 &   731 &   988 \\
1    &   712 &   950 \\
\bottomrule
\end{tabular}
\end{table}

Every sepsis state retains multiple actions under both certified constructions at all
seven widths. At $\lambda=0$, the chosen loss limit permits all 1,152 actions
under the nominal model. At the same loss limit, fixed-order augmentation retains
202 fewer actions when the family expands to full width.
Each certificate protects the family at its specified width. The sample
tests also evaluate the resulting policies on the full-width family, which
can include models outside the contracted family. The augmented policy violates the full-family loss limit on 11 of 12 sampled
full-range models at $\lambda=0$ and seven of 12 at $\lambda=0.25$. The local counts
are six and four. No sampled violation appears at $\lambda\ge0.5$. These sample
counts have no coverage interpretation.
Independent rational evaluation of the Bellman residuals certifies the
separated bound for all 14 local and augmented policies in the table. The smallest margin below the loss limit is $1.32\times10^{-6}$, and no probability intervals require
widening. At each of the seven widths, coupled augmentation adds no actions to its initial policy, so
these checks also certify the corresponding coupled policies. All 21 sepsis
method outputs therefore satisfy the loss requirement for their specified
uncertainty families.
The small width experiment uses two generated MDPs with two states, two
actions per state, and discount factor $0.6$. Their full uncertainty families
use reward and transition fractions 0.16 and 0.10, respectively. The loss limit
for each path lies 18\% of the way from the local-admission onset to the
all-actions threshold. The three methods and seven widths give 42 evaluations across the two models. Exact model-relative checking verifies
every returned policy, and 32 match the exact largest action count. Local, best-margin, and coupled exact-count matches are
10/14, 11/14, and 11/14. The average differences between the reported bounds and exact losses lie in
\begin{equation*}
\begin{array}{c|c}
\text{local}&[0.0912382721,0.0912382724]\\
\text{best-margin}&[0.0793457291,0.0793457293]\\
\text{coupled}&[0.0552921387,0.0552921389].
\end{array}
\end{equation*}

\subsection{Synthetic Interval Models}
\label{app:synthetic-controls}

We evaluate sufficient bounds on 432 synthetic interval MDPs and nine synthetic sepsis cases. Our synthetic grid crosses eight state counts, $1,2,4,8,16,32,64,128$, with two or four actions per state and three replications. Reward and transition interval half-widths are $0,0.05,0.25$. We use a discount factor of $0.9$ and a loss limit of $0.5$. For the nine sepsis cases, we cross uncertainty widths $0.25,0.5,1$ and loss limits $0.1,0.5,1$. These loss limits are smaller than the preceding sepsis width experiment's limit. (The augmented constructions share initial policy candidates and action-addition order.)

\begin{table}[!hbp]
\centering
\caption{Certified policies in the synthetic interval MDP experiments. Counts are out of 432 synthetic cases and nine sepsis cases. Augmentation uses a fixed state-action order. Shared-advantage construction uses the bound in \cref{app:rectangular-advantage}.}
\label{tab:controlled-results}
\begin{tabular}{lrr}
\toprule
Construction & Synthetic ($n=432$) & Sepsis ($n=9$) \\
\midrule
Local separated & 63 & 0 \\
Fixed-order separated & 64 & 0 \\
Coupled & 87 & 0 \\
Shared advantage & 87 & 0 \\
\bottomrule
\end{tabular}
\end{table}

Local separated, augmented separated, coupled, shared-advantage, and nominal fixed-model methods use the same instances and fixed candidate policies. The fixed-policy comparisons measure the bounds themselves. Search comparisons also reflect the actions each construction tries. Independent exact-rational residual checks certify all 301 returned policies on 87 distinct synthetic cases under the convention in \cref{app:experiment-arithmetic-timing}. Construction times cover floating-point value iteration and action selection. They exclude the independent checks. Exact checks on the small instances have a 60-second time limit. Of the applicable checks, 44 complete and ten reach this limit. These exact checks do not apply to the other 387 cases.

The nominal fixed-model method evaluates its policies against each specified family (Table~\ref{tab:supp-controlled-nominal}).

{\scriptsize
\setlength{\tabcolsep}{3pt}
\renewcommand{\arraystretch}{1.08}
\begin{longtable}{@{}lrrrrrrr@{}}
\caption{Fixed-model policies for every controlled uncertainty setting.}\label{tab:supp-controlled-nominal}\\
\toprule
Model & $w_r$ & $w_P$ & $w$ & $\delta$ & Nominal policies & Actions & Seconds \\
\midrule
\endfirsthead
\multicolumn{8}{l}{\emph{Table \thetable\ continued}}\\
\toprule
Model & $w_r$ & $w_P$ & $w$ & $\delta$ & Nominal policies & Actions & Seconds \\
\midrule
\endhead
\midrule\multicolumn{8}{r}{\emph{Continued on next page}}\\
\endfoot
\bottomrule
\endlastfoot
Synthetic & 0.00 & 0.00 & -- & 0.5 & 48 & 13.0 & 0.0061 \\
Synthetic & 0.00 & 0.05 & -- & 0.5 & 48 & 13.0 & 0.0066 \\
Synthetic & 0.00 & 0.25 & -- & 0.5 & 48 & 13.0 & 0.0063 \\
Synthetic & 0.05 & 0.00 & -- & 0.5 & 48 & 13.0 & 0.0059 \\
Synthetic & 0.05 & 0.05 & -- & 0.5 & 48 & 13.0 & 0.0065 \\
Synthetic & 0.05 & 0.25 & -- & 0.5 & 48 & 13.0 & 0.0061 \\
Synthetic & 0.25 & 0.00 & -- & 0.5 & 48 & 13.0 & 0.0058 \\
Synthetic & 0.25 & 0.05 & -- & 0.5 & 48 & 13.0 & 0.0063 \\
Synthetic & 0.25 & 0.25 & -- & 0.5 & 48 & 13.0 & 0.0064 \\
Sepsis & -- & -- & 0.25 & 0.1 & 1 & 176.0 & 0.0067 \\
Sepsis & -- & -- & 0.25 & 0.5 & 1 & 177.0 & 0.0066 \\
Sepsis & -- & -- & 0.25 & 1.0 & 1 & 177.0 & 0.0058 \\
Sepsis & -- & -- & 0.50 & 0.1 & 1 & 176.0 & 0.0062 \\
Sepsis & -- & -- & 0.50 & 0.5 & 1 & 177.0 & 0.0060 \\
Sepsis & -- & -- & 0.50 & 1.0 & 1 & 177.0 & 0.0064 \\
Sepsis & -- & -- & 1.00 & 0.1 & 1 & 176.0 & 0.0061 \\
Sepsis & -- & -- & 1.00 & 0.5 & 1 & 177.0 & 0.0059 \\
Sepsis & -- & -- & 1.00 & 1.0 & 1 & 177.0 & 0.0063 \\
\end{longtable}
}
{\footnotesize\noindent The table counts 432 synthetic and nine sepsis policies certified for the nominal model. Actions and seconds are medians over these policies. For synthetic models, $w_r$ and $w_P$ are reward and transition interval widths. For sepsis, $w$ is the width multiplier. These widths specify the paired experiment. Each policy's guarantee covers only its fixed model.\par}
\par\medskip

No tested construction certifies a positive-uncertainty case with eight or more states. All four constructions certify the 48 cases with zero reward and transition uncertainty. Among the 384 cases with positive uncertainty, local construction certifies 15 and coupled construction certifies 39 (\cref{tab:controlled-results}).

With positive uncertainty, certified policies concentrate in small problems (Figure~\ref{fig:synthetic-certification-map}). None of the cases with reward half-width $0.25$ certify.

The separated floor $V^+-V^-$ is within the limit in 64 cases, and fixed-order augmentation certifies exactly these, as Theorem~\ref{thm:certificate-soundness}(ii) predicts. Local construction certifies 63 of them. The coupled certificate adds 23 cases with the floor above the limit, reaching 87 in total. The largest floor among these additional certified cases is approximately $1.503$, about three times the limit. All 345 coupled unresolved cases have a separated floor above the limit. The smallest is approximately $0.940$. This rules out a separated certificate for each of these cases.

We enclose both value vectors using exact-rational Bellman residuals and optimize each transition row exactly over its box simplex. Subtracting the resulting enclosures gives certified bounds on the largest component of $V^+-V^-$. Every enclosure lies wholly on one side of the $0.5$ limit.

A reward-only diagnostic explains the scale of a separated floor. Fix the
transition probabilities and give every reward an unclipped interval
$[r^0_{sa}-h,r^0_{sa}+h]$ with the same half-width $h$. Shifting all rewards
by $h$ shifts every discounted policy value by $h/(1-\gamma)$, so
$V^+=V^0+h/(1-\gamma)$ and $V^-=V^0-h/(1-\gamma)$, where $V^0$ is the
nominal optimal value. The separated floor is therefore $2h/(1-\gamma)$.
It equals $1$ for $h=0.05$ and $\gamma=0.9$. This envelope calculation
assumes fixed transitions and no reward clipping. It is not a formula for
every synthetic instance or a lower bound on a policy's true loss.

\begin{figure}[t]
\centering
\includegraphics[width=\linewidth]{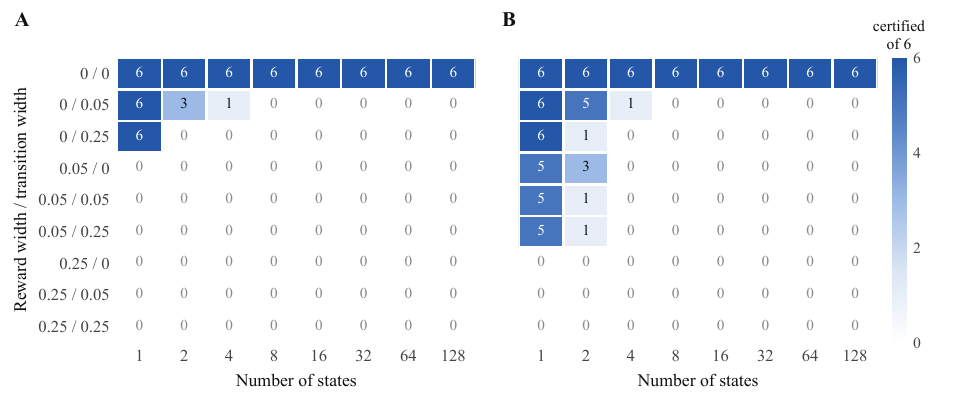}
\caption{Certified synthetic instances for augmented separated construction (Panel A) and coupled construction (Panel B), grouped by state count and reward and transition half-widths. Each cell combines two action counts and three replications.}
\label{fig:synthetic-certification-map}
\end{figure}

The infeasibility check identifies cases where returning no recommendation is necessary. It rules out every safe nonempty set-valued policy in 244 of the 384 positive-uncertainty cases. The constructions certify 39 other cases (Table~\ref{tab:synthetic-infeasibility}). At eight or more states, 201 of 240 cases are infeasible, including all 108 with transition uncertainty at 32 or more states.

In 18 one-state cases, the nominally best action is optimal under every model, so a zero-loss singleton exists below the separated floor. These cases illustrate the observation following Theorem~\ref{thm:certificate-soundness}. For the best action $a$, its lower reward exceeds every other action's upper reward. Its exact loss is therefore $\max\{0,\max_{a'\ne a}r^+_{a'}-r^-_a\}/(1-\gamma)=0$. Fifteen have reward half-width $0.25$, with separated floor approximately $5$. The other three have half-width $0.05$, with floor approximately $1$. Comparisons between distinct actions retain separated continuation bounds, so the coupled certificate also exceeds the limit in these cases.

The remaining 83 cases have neither a certified feasible policy nor an impossibility proof. Of these, 44 have at most four states, and 18 have more than 16.

\begin{table}[htbp]
\centering
\caption{Outcomes for synthetic cases with positive uncertainty, with 48 instances at each state count and loss limit $0.5$. The table distinguishes certified policies from analytically established feasibility.}
\label{tab:synthetic-infeasibility}
\begin{tabular}{rrrrr}
\toprule
States & Construction certified & Analytically feasible & Proved infeasible & Unresolved\\
\midrule
1 & 27 & 18 & 3 & 0\\
2 & 11 & 0 & 11 & 26\\
4 & 1 & 0 & 29 & 18\\
8 & 0 & 0 & 33 & 15\\
16 & 0 & 0 & 42 & 6\\
32 & 0 & 0 & 42 & 6\\
64 & 0 & 0 & 42 & 6\\
128 & 0 & 0 & 42 & 6\\
\midrule
Total & 39 & 18 & 244 & 83\\
\bottomrule
\end{tabular}
\end{table}

For each action index, we form a model favoring that action with upper rewards and optimistic transition rows.
Other actions receive lower rewards and pessimistic rows.
The transition rows maximize or minimize a nominal value vector over the recorded box simplex.
Their model membership and probability sums are checked in rational arithmetic.
For point-model rows, we use the normalization convention in \cref{app:experiment-arithmetic-timing}.
The largest coordinate change is below $1.48\times10^{-16}$.

To establish infeasibility, write $\ell_k(s,a)$ for the one-step loss in model $k$, where $\ell_k(s,a)=V_k^\star(s)-r_k(s,a)-\gamma\sum_{s'} p_k(s'\mid s,a)V_k^\star(s')$.
Exact Bellman residuals give lower bounds $\underline \ell_k(s,a)\le \ell_k(s,a)$.
Suppose a deterministic stationary policy $\sigma$ meets the loss limit $\delta$ in every selected model and at every state.
Its losses satisfy $0\le \rho_k\le\delta$ and
\[
 \rho_k(s)=\ell_k(s,\sigma(s))+\gamma\sum_{s'} p_k(s'\mid s,\sigma(s))\rho_k(s').
\]
Start with all actions and lower bounds $\underline\rho_k(s)=0$.
Every policy $\sigma$ meeting the limit has its actions retained and satisfies $\underline\rho_k\le\rho_k$ throughout. This holds initially because $\rho_k\ge0$.
For each remaining action, compute
\[
 \zeta_k(s,a)=\underline \ell_k(s,a)+\gamma\sum_{s'} p_k(s'\mid s,a)\underline\rho_k(s').
\]
Delete action $a$ at state $s$ if $\zeta_k(s,a)>\delta$ for any model $k$.
Under the invariant, $\zeta_k(s,\sigma(s))\le\rho_k(s)\le\delta$, so $\sigma$ never uses a deleted action.
Then raise $\underline\rho_k(s)$ to the larger of its current value and the minimum $\zeta_k(s,a)$ over remaining actions.
Both the current bound and this minimum are at most $\rho_k(s)$, since the minimum is at most $\zeta_k(s,\sigma(s))\le\rho_k(s)$. The invariant is therefore preserved.
Downward rational rounding preserves these lower bounds.
Repeat these steps while computation permits, stopping if an action set is empty or no bound changes.
An empty action set contradicts the existence of $\sigma$.
Every safe nonempty set-valued policy contains such a compliant policy, so the contradiction also rules out that set-valued policy.
This argument does not rule out an individual randomized or history-dependent policy.
All elimination steps are verified in exact arithmetic, without relying on a solver's infeasibility status.

None of the four robust constructions certifies any of the nine sepsis cases at these loss limits (Table~\ref{tab:controlled-results}). No safe nonempty set-valued policy exists at these loss limits.
For each width, we shrink the simulator completions toward the first completion by that width.
Every resulting model lies in the corresponding interval family.
We rule out a deterministic stationary policy that satisfies the limit in every selected model using the following rational certificate.

For the rational certificate, index the selected models by $k=1,\ldots,K$ and write
$\ell_k(s,a)=V_k^\star(s)-r_k(s,a)-\gamma\sum_{s'} p_k(s'\mid s,a)V_k^\star(s')$.
For any deterministic stationary policy $\sigma$, its nonnegative loss $\rho_k=V_k^\star-V_k^\sigma$ satisfies
\[
 \rho_k(s)=\ell_k(s,\sigma(s))+\gamma\sum_{s'} p_k(s'\mid s,\sigma(s))\rho_k(s').
\]
Let $\underline \ell_k(s,a)\le \ell_k(s,a)$, and define
\[
 (\mathcal J z)(s)=\min_a\left\{\sum_{k=1}^K\underline \ell_k(s,a)
 +\gamma\sum_{s'}\min_k p_k(s'\mid s,a)z(s')\right\}.
\]
Since $\rho_k\ge0$, their sum $\rho_\Sigma=\sum_k \rho_k$ satisfies $\rho_\Sigma\ge\mathcal J \rho_\Sigma$.
The operator $\mathcal J$ is a monotone contraction because its transition rows sum to at most one.
Let $z_{\mathcal J}$ be its fixed point. A verified subsolution $z\le\mathcal Jz$ satisfies $z\le z_{\mathcal J}$. Since $\mathcal J\rho_\Sigma\le\rho_\Sigma$, fixed-point comparison also gives $z_{\mathcal J}\le\rho_\Sigma$. Thus $z\le\rho_\Sigma$ for every such policy.
If $z(s)>K\delta$, at least one selected model exceeds loss limit $\delta$ at state $s$.

We compute these certificates with exact rational arithmetic for the binary input rewards, probabilities, and discount factor.
The interpolation widths are the rational numbers $1/4$, $1/2$, and $1$.
Numerically computed optimal values only propose bounds.
For a proposed value vector $v_k$, we evaluate its Bellman residual exactly and set
$e_k=\|T_k^\star v_k-v_k\|_\infty/(1-\gamma)$.
Then $\|v_k-V_k^\star\|_\infty\le e_k$, so we use
\[
 \underline \ell_k(s,a)=\max\left\{0,\,
 v_k(s)-r_k(s,a)-\gamma\sum_{s'} p_k(s'\mid s,a)v_k(s')-(1+\gamma)e_k\right\}.
\]
The final subsolution inequalities and model inclusion are also checked exactly, with all eight actions retained.
At width $1/4$, completions 0 and 2 give $z(25)>2.8855$ with $K=2$.
At widths $1/2$ and $1$, all ten completions give $z(24)>13.2707$ and $z(41)>13.4130$, respectively.
Each bound exceeds $K$, ruling out loss limits $0.1$, $0.5$, and $1$.
Every nonempty set-valued policy contains a deterministic stationary compliant policy, so no set-valued policy can guarantee those limits for every compliant policy.

\subsection{Shared Uncertainty and Statistical Selection}
\label{sec:shared-uncertainty-controls}
\label{sec:simulation-comparison}

We build controlled models to isolate two ways in which loss can be misjudged. Bounds that ignore shared parameters can overstate loss, while comparisons made one epoch at a time can understate it.

\paragraph{Shared rewards.}
Separate value bounds can discard dependence between actions, as illustrated by one decision with rewards $\xi$ and $\xi-0.01$, where $\xi\in[0.8,1]$ (\cref{app:controlled-statistical-results}). Permitting both actions incurs loss $0.01$, while raw and advantage bounds are approximately $0.21$ because they compare opposite endpoints. Retaining the common $\xi$ gives joint bound $0.01$, which certifies the loss limit $0.03$.

\paragraph{Statistical selection.}
Simulation-based selection keeps actions whose sampled values are not significantly worse than a reference action, using simultaneous comparisons and estimated continuation values (\citealt{marrero2022}; \cref{app:statistical-comparison-design}). Seven two-action models examine one-epoch ties and gaps, ten-epoch accumulation, and the effect of estimated continuation. With 300 inference batches of 512 observations each, small local losses accumulate beyond the loss limit $0.03$ in 2 of 200 ten-epoch replications at significance level 0.05 and 3 at 0.10. With estimated continuation, an optimal action is excluded in 195 of 200 replications at each significance level even though the resulting policies remain within the loss limit (Table~\ref{tab:statistical-controls}). Separate one-decision, two-action common-offset controls add the same uncertain reward shift to both actions, preserving their difference as in the preceding example. These use 100 inference batches of 128 observations and positive reward half-widths of $0.5\delta$ and $\delta$, with $\delta=0.03$. At each width and significance level ($0.05$ or $0.10$), the joint certificate passes all 200 replications while raw and advantage bounds certify none.

\label{app:statistical-selection-results}

\subsubsection{Statistical design}
\label{app:statistical-comparison-design}

The statistical comparison examines limited simulation observations under a
specified model. Simulation-based multiple comparisons provide a published
precedent for constructing sets of similarly performing actions
\citep{marrero2022}. We specify the held-out-control statistical method below, including its continuation-value and zero-variance conventions.
At state $s$ and decision epoch $t$, fix continuation values $\widehat v_{t+1}$ and
generate $n_B+1$ independent batches of $n_K$ observations
$Y_{abk}=R_{abk}+\gamma\widehat v_{t+1}(S'_{abk})$ for each admissible
action $a$. Batches are indexed by $b=0,1,\ldots,n_B$. Let $\bar Y_{ab}$
and $s^2_{ab}$ be the within-batch sample mean and unbiased variance.
The held-out batch $b=0$ selects the control
$c\in\arg\max_a\bar Y_{a0}$, with action order breaking ties. The
remaining batches define
\[
 \widehat Q_a=\frac{1}{n_B}\sum_{b=1}^{n_B}\bar Y_{ab},\qquad
 S_a^2=\frac1{n_B-1}\sum_{b=1}^{n_B}(\bar Y_{ab}-\widehat Q_a)^2.
\]
For a stochastic comparison with rival $a\ne c$, define the studentized
root in inference batch $b$ by
\[
 Z_{ab}=\frac{(\bar Y_{cb}-\bar Y_{ab})-(\widehat Q_c-\widehat Q_a)}
 {\sqrt{(s^2_{cb}+s^2_{ab})/n_K}}.
\]
Let $z_{1-\alpha}$ be the order statistic of rank
$\lceil(1-\alpha)n_B\rceil$ among the $n_B$ values $\max_{a\ne c}Z_{ab}$, where the maximum includes
only stochastic comparisons. If all comparisons are deterministic,
$z_{1-\alpha}=0$. The estimated lower bound for the control--rival mean difference is
\begin{equation}
 L_a=\widehat Q_c-\widehat Q_a
       -z_{1-\alpha}\sqrt{(S_c^2+S_a^2)/n_B}.
 \label{eq:statistical-control-lower}
\end{equation}
The proposed action set contains the control and all rivals with
$L_a\le0$. The statistical test threshold is zero, independently of
the cumulative loss limit $\delta$. Backward induction sets
$\widehat v_t(s)=\max_a\widehat Q_a$, independently of which rivals
are retained. An oracle-continuation comparison substitutes the exact
optimal $v^*_{t+1}$ to isolate the effect of continuation estimation.

All actions and batches are sampled independently. Within each inference
batch, the same control observations enter all rival comparisons, inducing
their intended dependence.
Multinomial batch counts have the distribution of the corresponding
independent draws from the finite reward and successor distributions.
Statistical calculations use ordinary binary64 probabilities. Continuous-family certification uses the outward-rounded arithmetic in \cref{app:experiment-arithmetic-timing}. The empirical-quantile procedure is evaluated by its observed
coverage across independent replications.
Seven controlled models examine ties, gaps of one quarter, one, and
1.5 times the loss limit, ten-epoch accumulation of small local gaps,
and paired estimated- versus exact-continuation two-stage models.
The grid has $n_B\in\{30,100,300\}$, $n_K\in\{32,128,512\}$,
$\alpha\in\{0.05,0.10\}$, $\delta=0.03$, and 200 independent
replications, giving 25,200 replications. A further 2,400 replications use $n_B=100$
and $n_K=128$ to vary a shared reward offset or opposing action means, with
half-widths $0$, $\delta/2$, and $\delta$, while holding the nominal
simulator fixed. The same sampled policy is checked
using raw, advantage, and joint certificates. A separate exact Bernoulli calculation, given below, examines why control
selection cannot be guaranteed from a reward-range bound alone.

For a structurally deterministic action, its conditional mean is used
exactly and $S_a^2=0$. A pair of deterministic actions uses their exact
conditional difference, without a studentized root or confidence margin.
If any stochastic rival comparison has a zero within-batch denominator,
the entire state--epoch comparison retains all admissible actions and
reports no finite lower bounds. Abstention counts accompany coverage,
so vacuous bounds are distinguishable from informative comparisons. An additive-tolerance reference
starts from an exact nominal optimal singleton policy and greedily
accepts an action only when exact fixed-model worst-compliant policy loss
satisfies the loss limit at every state and decision epoch.

Controlled coverage uses independent replications and pointwise two-sided 95\% Clopper--Pearson intervals for each event. Conditional coverage targets the means given estimated continuation, while true-$Q^*$ coverage targets optimal action values. Each whole-replication indicator requires coverage across all state--epoch comparison families, and $\alpha$ indexes the per-state--epoch comparison. The clinical panel has one replication per patient and significance level, paired across widths, giving descriptive comparisons on those patients.

\subsubsection{Additional controlled results}
\label{app:controlled-statistical-results}

\begin{table}[t]
\centering
\caption{Simulation comparisons at $n_B=300$ inference batches and $n_K=512$ observations per batch. Entries give counts at $\alpha=0.05/0.10$, each out of 200 independent replications.}
\label{tab:statistical-controls}
\small
\setlength{\tabcolsep}{4pt}
\begin{tabular}{lcccc}
\toprule
Control & \shortstack{Loss\\violations} & \shortstack{Any optimal\\action excluded} & \shortstack{Conditional\\coverage} & \shortstack{True-$Q^*$\\coverage} \\
\midrule
Tie & 0/0 & 12/20 & 188/180 & 188/180 \\
Quarter-limit gap & 0/0 & 0/0 & 187/166 & 187/166 \\
Loss-limit gap & 0/0 & 0/0 & 186/174 & 186/174 \\
Large gap & 0/0 & 0/0 & 190/178 & 190/178 \\
Ten-epoch accumulation & 2/3 & 0/0 & 110/71 & 110/71 \\
Estimated continuation & 0/0 & 195/195 & 163/157 & 5/5 \\
Oracle continuation & 0/0 & 33/46 & 167/154 & 167/154 \\
\bottomrule
\end{tabular}
\par\smallskip
\begingroup
\footnotesize
Coverage counts require every recorded epoch--state comparison family in a replication to be covered. Conditional targets use the continuation values supplied to the comparison. True-$Q^*$ targets use optimal continuation. The specified significance level $\alpha$ applies separately to each position's comparison family. Loss violations use exact fixed-model worst-compliant loss above $\delta=0.03$.
\endgroup
\end{table}

The five analytic controls examine common uncertainty, conflicting action preferences, zero model uncertainty, equality at the loss limit, and accumulation. With three decisions and discount factor 0.97, the joint certificate encloses the common-baseline loss 0.029109. All five agree with their analytic conclusions.

The controlled statistical grid has 126 settings and the width grid has twelve, each with 200 replications. Oracle continuation makes the conditional and true-$Q^*$ coverage targets coincide. Across the full controlled grid, 248 losses in the loss-limit-gap control exceed 0.03 by approximately $2.66\times10^{-17}$. The comparisons use exact binary64 reward values.

With 300 batches of 512 observations, only the ten-epoch model exceeds the loss limit (Table~\ref{tab:statistical-controls}). All three certificate types certify every common-offset and opposing-means policy at zero width.

The opposing-means control has one decision and rewards $1+\eta\theta$ and $1-\eta\theta$, where $\theta\in[-1,1]$. Each singleton loses $2\eta$ at the endpoint favoring the excluded action, and the two-action policy has the same worst compliant loss. At $\eta=0.03$, every nonempty set-valued policy therefore exceeds loss limit 0.03. At $\eta=0.015$, the mathematical loss equals the limit, while the outward numerical enclosure leaves the check unresolved. 

\subsubsection{Clinical statistical selection}
\label{app:clinical-statistical-results}

The clinical statistical experiment uses the fixed 244-patient panel with 300 inference batches and 512 observations per batch. It uses significance levels 0.05 and 0.10, and factor widths $w\in\{0,0.5,1,2\}$. The nominal simulator, action sets, and random seed are paired across widths. Each significance level has a separate random seed. We check each proposed plan against the full family and seek a certified subset within its proposed sets. The optimal-value benchmark uses the full admissible action sets. Table~\ref{tab:clinical-computation-settings} gives the cover and construction limits.

At significance level 0.05, full-family checks certify 41, 18, 2, and 0 unchanged proposals at widths 0, 0.5, 1, and 2. At significance level 0.10, the corresponding counts are 43, 20, 2, and 0. All other proposals have explicit models at which the loss exceeds the limit (Table~\ref{tab:clinical-statistical-proposals}). Proposal rejection concerns that set-valued treatment plan. Subset existence is evaluated separately in Table~\ref{tab:clinical-statistical-subsets}.

\begin{longtable}{@{}lrrrr@{}}
\caption{Checks of statistically selected treatment plans against the model family.}\label{tab:clinical-statistical-proposals}\\
\toprule
$w$ & $\alpha$ & Certified & Verified violation & Unresolved \\
\midrule
\endfirsthead
\toprule
$w$ & $\alpha$ & Certified & Verified violation & Unresolved \\
\midrule
\endhead
\bottomrule
\endfoot
0 & 0.05 & 41 & 203 & 0 \\
0 & 0.10 & 43 & 201 & 0 \\
0.5 & 0.05 & 18 & 226 & 0 \\
0.5 & 0.10 & 20 & 224 & 0 \\
1 & 0.05 & 2 & 242 & 0 \\
1 & 0.10 & 2 & 242 & 0 \\
2 & 0.05 & 0 & 244 & 0 \\
2 & 0.10 & 0 & 244 & 0 \\
\end{longtable}
{\footnotesize\noindent Every setting uses the same 244 input-selected patients. A verified violation applies to the specified treatment plan. Certified subsets appear in Table~\ref{tab:clinical-statistical-subsets}.\par}

\begin{longtable}{@{}rrrrrrrr@{}}
\caption{Certified subsets of statistically selected actions.}\label{tab:clinical-statistical-subsets}\\
\toprule
$w$ & $\alpha$ & \shortstack{Flexible\\(at least)} & Singleton & Infeasible & Unresolved & \shortstack{Proposal actions\\per pair} & \shortstack{Subset actions\\per pair} \\
\midrule
\endfirsthead
\toprule
$w$ & $\alpha$ & \shortstack{Flexible\\(at least)} & Singleton & Infeasible & Unresolved & \shortstack{Proposal actions\\per pair} & \shortstack{Subset actions\\per pair} \\
\midrule
\endhead
\bottomrule
\endfoot
0 & 0.05 & 244 & 0 & 0 & 0 & 9.68 & 8.26 \\
0 & 0.10 & 244 & 0 & 0 & 0 & 9.56 & 8.21 \\
0.5 & 0.05 & 230 & 3 & 0 & 11 & 9.68 & 4.57 \\
0.5 & 0.10 & 235 & 1 & 0 & 8 & 9.56 & 4.88 \\
1 & 0.05 & 189 & 1 & 0 & 54 & 9.68 & 2.95 \\
1 & 0.10 & 183 & 3 & 0 & 58 & 9.56 & 3.03 \\
2 & 0.05 & 151 & 0 & 0 & 93 & 9.68 & 2.22 \\
2 & 0.10 & 143 & 0 & 0 & 101 & 9.56 & 2.23 \\
\end{longtable}
{\footnotesize\noindent A pair is a living health state and decision year. Every setting uses the same 244 input-selected patients without survey weighting. Flexible counts are lower bounds. Proposal sizes use all patients, and subset sizes condition on certified flexible subsets. Singleton denotes a singleton certificate without a certified flexible subset. Status columns concern existence within the proposed action sets.\par}

At width zero, explicit nominal models reject 203 proposals at significance level 0.05 and 201 at 0.10, while certified flexible subsets exist for all 244 patients at either level. At width one, certified flexible subsets exist for at least 189 and 183 patients, respectively.

At significance level 0.05, at least one nominal-model optimal action is excluded for 14 patients. True-$Q^*$ and conditional coverage across all comparisons each hold for 57 patients. At significance level 0.10, these counts are 43, 20, and 20. These nominal comparisons are paired across widths. Cover certification separately establishes the loss guarantee over the full specified family for every compliant choice from the certified subset.

\subsubsection{What the controls show}

Shared-parameter bounds preserve cancellation between action values. Statistical comparisons answer whether actions can be distinguished from sampled observations. The ten-epoch controls show why that comparison differs from bounding cumulative loss. The estimated-continuation controls show that excluding an optimal action can still leave loss below a positive limit.

\paragraph{Exact control-selection diagnostic.}
Consider $A$ actions with independent reward observations in $[0,\kappa]$,
$N$ observations per action, true optimal action set $A^*$, and empirical
maximizer set $\widehat A_N$. A bound of the form
\[
 \Pr\{\widehat A_N\not\subseteq A^*\}
 \le A\exp\{-N/(2\kappa^2)\}
\]
would suggest $N\ge2\kappa^2\log(A/\beta)$ to make this probability at
most $\beta$, irrespective of the gap between action means. This
inequality is false even in a one-stage problem. Set $A=2$, $\kappa=1$,
$\beta=0.01$, and let the rewards be Bernoulli with means 0.501 and
0.5. The suggested sample size is $N=\lceil2\log(200)\rceil=11$.
For independent $X\sim\operatorname{Binomial}(11,0.501)$ and
$Y\sim\operatorname{Binomial}(11,0.5)$, exact summation gives
\[
 \Pr(Y>X)
 =\sum_{i=0}^{11}\Pr(X=i)\sum_{j=i+1}^{11}\Pr(Y=j)
 =0.4140567696754165\ldots,
\]
whereas the proposed upper bound is $2e^{-11/2}=0.0081735429\ldots$.
Strict inferior selection alone exceeds the bound. No tie convention
can remove the discrepancy. The statistical experiments therefore
measure control-selection errors directly and choose $n_B$ and $n_K$
from the stated design grid.

\subsection{Parametric reference comparison}
\label{app:parametric-reference}

We test the bounds of \cref{sec:continuous-certificates} on one-parameter families built from the ten small reference problems of \cref{sec:computational}. For each nominal model $M_0$, we draw $U_0,U_1$ from the same generator and set
\[
 M(\theta)=\tfrac34 M_0+\tfrac14\bigl[(1-\theta)U_0+\theta U_1\bigr],\qquad 0\le\theta\le1.
\]
Each $M(\theta)$ lies in the interval family of the same reference problem. The discount factor and loss limits are unchanged. We enumerate all 166 nonempty set-valued policies. With one state, loss is convex in $\theta$, so we check both endpoints. With two states, we check $\theta=j/1000$, $j=0,\ldots,1000$, exactly. The \emph{reference count} is the largest action count among policies passing all these point checks, so no safe policy can exceed it.

For each policy passing the point checks, we bisect the cell with the largest advantage bound until that bound certifies the policy or the cover reaches 64 cells. We evaluate all three bounds on the same cover in exact rational arithmetic. With a single cell, the raw, advantage, and joint bounds reach the reference count in 27, 22, and 28 of the 30 instance--limit pairs. With at most eight cells, all three reach it in every pair. So does the construction of \cref{sec:common-procedure}, with $\theta=0$, $1/2$, and $1$ as its chosen models. The certified counts are therefore exact maxima over the whole family. Sixteen of them exclude at least one action.

\FloatBarrier
\section{Infeasibility Certificates from Finite Model Lists}
\label{app:clinical-existence}

Models selected from a continuous uncertainty family provide a
complementary infeasibility test. If no single policy is safe in all of those
models, none can be safe in the entire family. For a finite list, we test existence by
requiring one policy to satisfy the loss limit in all of them. This appendix
first expresses that requirement exactly, then removes actions without
losing a safe solution, and finally weakens the remaining constraints.
If even those weaker constraints have no solution, no safe policy exists on
the original list. A solution to the weaker constraints must still be checked
against the exact requirement.

\subsection{One policy shared by all listed models}

Let $M_1,\ldots,M_{N_{\rm mod}}$ be complete finite-horizon models. Their common action
sets are $\Aset_t(s)$, their decision epochs are $t=0,\ldots,T-1$, and their
terminal values are $h_{M_k}(s)$. Here $0\le\gamma\le1$ is allowed because the
horizon is finite. Backward induction gives each model's optimal values as
\begin{equation*}
 V^\star_{M_k,T}(s)=h_{M_k}(s),\qquad
 V^\star_{M_k,t}(s)=\max_{a\in\Aset_t(s)}
 \{r_{M_k,tsa}+\gamma\sum_jP_{M_k,tsa,j}V^\star_{M_k,t+1}(j)\}.
\end{equation*}
The indices $k,t,s,a$ identify a model, decision epoch, state, and action. A common
deterministic policy $\sigma$ chooses the same action $\sigma_t(s)$ in every
model. Its value uses the same recursion with that action fixed.

Define cumulative policy loss and one-step loss, respectively, by
\begin{equation*}
 \rho^\sigma_{M_k,t}(s)=V^\star_{M_k,t}(s)-V^\sigma_{M_k,t}(s),\qquad
 \ell_{M_k,tsa}=V^\star_{M_k,t}(s)-r_{M_k,tsa}
            -\gamma\sum_jP_{M_k,tsa,j}V^\star_{M_k,t+1}(j).
\end{equation*}
Both quantities are nonnegative. The second takes action $a$ once and uses
optimal continuation. If $\sigma_t(s)=a$, subtracting the two value
recursions gives
\begin{equation}
 \rho^\sigma_{M_k,T}=0,\qquad
 \rho^\sigma_{M_k,t}(s)=\ell_{M_k,tsa}
       +\gamma\sum_jP_{M_k,tsa,j}\rho^\sigma_{M_k,t+1}(j).
 \label{eq:finite-loss-recursion}
\end{equation}
Both policies receive the same terminal value in each model, giving zero terminal loss even when $h_{M_k}$ differs across models.

For a uniform loss limit $\delta\ge0$, the policy is safe at every position in
every model if and only if there are upper bounds $Y$ satisfying
\begin{align}
 0&\le Y_{M_k,t}(s)\le \delta,\qquad Y_{M_k,T}(s)=0,\notag\\
 Y_{M_k,t}(s)&\ge \ell_{M_k,ts,\sigma_t(s)}
 +\gamma\sum_jP_{M_k,ts,\sigma_t(s),j}Y_{M_k,t+1}(j).
 \label{eq:clinical-exact-supersolution}
\end{align}
A safe policy can take $Y=\rho^\sigma$. Conversely, starting from the zero
terminal loss, backward induction gives $\rho^\sigma\le Y\le \delta$.
One shared action choice at each position, together with these linear
inequalities for every model, is therefore an exact existence formulation.
By downward closure, it also decides whether any nonempty safe set-valued
policy exists. The screening arguments below require the loss limit to hold
at \emph{every} epoch and state.

\subsection{Removing actions without losing a safe solution}

If $\ell_{M_k,tsa}>\delta$ for any listed model, a policy choosing $a$ at $(t,s)$
already loses too much before any later loss is added.
\Cref{eq:finite-loss-recursion} therefore permits removal of this action.
If all actions at one position satisfy this test, no safe policy exists.
An implementation may retain one of those actions to keep its action arrays
nonempty. Doing so weakens the screen while preserving the validity of a later
impossibility result.

A second rule compares two actions across every continuation value that a
safe policy could attain. Safety and optimality imply
\begin{equation*}
 V^\star_{M_k,t+1}-\delta\one\le V^\sigma_{M_k,t+1}\le V^\star_{M_k,t+1}.
\end{equation*}
For actions $a',a$, write
$\Delta P_{M_k,tsj}(a',a)=P_{M_k,tsa',j}-P_{M_k,tsa,j}$. The smallest improvement from
replacing $a$ by $a'$ over this entire continuation box is
\begin{equation}
 \mu_{M_k,ts}(a',a)=r_{M_k,tsa'}-r_{M_k,tsa}
 +\gamma\sum_j\Delta P_{M_k,tsj}(a',a)
 \begin{cases}
 V^\star_{M_k,t+1}(j)-\delta,&\Delta P_{M_k,tsj}(a',a)>0,\\
 V^\star_{M_k,t+1}(j),&\Delta P_{M_k,tsj}(a',a)\le0.
 \end{cases}
 \label{eq:clinical-box-dominance}
\end{equation}
A positive coefficient is minimized at the lower endpoint. A negative
coefficient is minimized at the upper endpoint. Thus, if
$\mu_{M_k,ts}(a',a)>0$ for every model $k$, action $a'$ improves on $a$ for
all possible safe continuation vectors. We call this strict dominance on
the safe continuation boxes. Remove $a$ only if a replacement remains,
either $a'$ itself or an action reached through verified dominance replacements.

\begin{proposition}[Existence-preserving screen]
\label{prop:clinical-existence-screen}
One-step-loss removals and strict continuation-box dominance removals,
with retained replacements as specified above, preserve the existence of a
policy satisfying the loss limit in every listed model, state, and decision epoch.
\end{proposition}

\begin{proof}{Proof.}
No safe policy can use an action removed by the one-step rule, by
\cref{eq:finite-loss-recursion}. For a dominance removal, consider a safe
policy using $a$. Its continuation values lie in the specified boxes.
Replacing $a$ by $a'$ at that position therefore raises its value in every
model. Later values are unchanged. Earlier values cannot decrease because
their continuation coefficients are nonnegative. The modified policy is
still safe and makes the same replacement in every model.

Apply this argument to successive removals. At a fixed position, strict
dominance has no cycle. Evaluating a putative cycle at any continuation
vector in any model's box would require a strictly increasing action value
to return to itself. Dominance is also transitive. If $a'$ improves on $a$ and $a''$ improves on
$a'$ throughout the same box, adding their action-value differences shows
that $a''$ improves on $a$ throughout that box. A finite replacement chain therefore ends at a
retained action with the required improvement. An action ruled out by the
one-step rule cannot be the sole replacement of an action used by a safe
policy, since the preceding replacement argument would make it safe.

Thus a safe original policy can be transformed into a safe policy on the
retained action domain. Conversely, every retained-domain policy is allowed
on the original domain. 
\Halmos
\end{proof}

Only verified strict comparisons justify a removal. Retaining an action
whose margin is unresolved is always permissible. Extra retained actions
preserve the validity of an impossibility result, although they may make the
test harder to solve.

\subsection{A rational outer relaxation}

The exact existence formula can be large. We simplify its coefficients in
a direction that admits every exact solution. For a nonnegative number $z$
and integer $q\ge1$, define
$\lfloor z\rfloor_q=2^{-q}\lfloor2^qz\rfloor$, and set
\begin{equation*}
 \underline\ell_{M_k,tsa}=\lfloor \ell_{M_k,tsa}\rfloor_q,\qquad
 \underline c_{M_k,tsa,j}=\lfloor\gamma P_{M_k,tsa,j}\rfloor_q.
\end{equation*}
Each coefficient is rounded downward to a multiple of $2^{-q}$. Choose one
retained action at each position, shared across models, and introduce
$0\le L_{M_k,t}(s)\le \delta$. For every chosen action $a$, require
\begin{equation}
 L_{M_k,t}(s)\ge\underline\ell_{M_k,tsa}
      +\sum_j\underline c_{M_k,tsa,j}L_{M_k,t+1}(j),\qquad L_{M_k,T}=0.
 \label{eq:clinical-outer-recurrence}
\end{equation}
The coefficients and loss variables are nonnegative. Rounding coefficients
downward therefore decreases the right-hand side, so every exact feasible
assignment remains feasible. This inclusion is why the formula is called
an outer relaxation.

\begin{theorem}[Sound impossibility certificate]
\label{thm:clinical-outer-unsat}
Choose any nonempty subset of the listed models and apply the
existence-preserving screen on that subset. If its outer relaxation
\eqref{eq:clinical-outer-recurrence} is unsatisfiable, no policy is safe on the full list. Consequently, no nonempty
safe set-valued policy exists on that list.
\end{theorem}

\begin{proof}{Proof.}
Suppose a policy were safe for the full list. It would be safe on the
chosen subset. Screening would preserve some safe policy on the retained
action domain. Assign $L$ to that policy's exact cumulative losses.
Since $\underline\ell\le\ell$, $0\le\underline c\le\gamma P$, and
$\rho^\sigma\ge0$, its exact recursion implies every rounded inequality.
Safety supplies $0\le L\le \delta$, and the common terminal value supplies
$L_{M_k,T}=0$. This assignment would satisfy the outer relaxation
\eqref{eq:clinical-outer-recurrence}, contradicting unsatisfiability. The equivalence between singleton and set-valued-policy
existence proves the final statement.
\Halmos
\end{proof}

If a user-supplied action restriction has not been proved to preserve
existence, unsatisfiability establishes impossibility only on that restricted
domain. Likewise, impossibility on the finite list transfers to a larger
uncertainty family only when the listed models actually belong to it.
Extending safety on the finite list to the larger family requires an
additional certificate.

For stored IEEE-754 binary64 inputs, exact arithmetic can interpret every
coefficient as the rational number represented by its bits. Backward
induction, screening, and the final Boolean and rational linear formula can
then avoid floating-point feasibility tolerances. The proofs above use
nonnegative continuation coefficients and a finite horizon. They therefore
also apply algebraically to the recursions defined by stored coefficients,
even if rounding makes a transition row's exact rational sum differ from
one. In that case the result concerns those numerical recursions. Applying
it to an underlying analytic MDP requires a separate argument relating the
stored coefficients to that MDP. A solver timeout leaves existence
unresolved. A satisfying outer assignment must pass the unrounded formula
or an exact policy evaluation before it establishes feasibility.

\subsection{Enclosed point models and propagated loss bounds}
\label{app:enclosed-infeasibility}

For the continuous clinical family, each listed parameter belongs to the
original domain and is enclosed as a degenerate parameter cell. Outward
reward, transition, and terminal-value bounds give bounds
$\underline V^\star_{k,t}$ and $\overline Q^\star_{k,tsa}$ on the
optimal value and optimal action value of that same model. Both use the
full original action sets. Define direct lower bounds on action loss and
continuation coefficients by
\begin{equation}
 \underline\ell_{k,tsa}
 =\max\{0,\underline V^\star_{k,t}(s)-\overline Q^\star_{k,tsa}\},
 \qquad
 \underline c_{k,tsa,j}=\gamma\underline P_{k,tsa,j}.
 \label{eq:enclosed-loss-coefficients}
\end{equation}
The lower transition bounds are intersected with nonnegativity. The
optimal benchmark is kept fixed when actions are removed from a candidate
domain. In particular, the coefficients in
\cref{eq:enclosed-loss-coefficients} are bounds on the original loss
recursion, not the action losses of a model assembled from lower endpoints.
The clinical calculations interpret $\gamma=0.97$ and $\delta=0.03$ exactly.

\begin{corollary}[Infeasibility from enclosed family members]
\label{cor:enclosed-infeasibility}
Suppose each point-model enclosure contains its family member and the
derived value bounds use the original admissible actions. Replace the
coefficients of \cref{eq:clinical-outer-recurrence} by
\cref{eq:enclosed-loss-coefficients}, with optional downward binary rounding.
If the resulting necessary conditions have a checked contradiction,
no policy satisfies the loss limit throughout the continuous family.
The same conclusion holds after any action removals proved to preserve
existence for these enclosed losses.
\end{corollary}
\begin{proof}{Proof.}
A policy safe throughout the family is safe in every listed point model.
Its exact losses satisfy \cref{eq:finite-loss-recursion} and lie between
zero and $\delta$. The enclosure assumptions give
$0\le\underline\ell_{k,tsa}\le\ell_{M_k,tsa}$ and
$0\le\underline c_{k,tsa,j}\le\gamma P_{M_k,tsa,j}$.
Thus its exact losses satisfy the relaxed inequalities, including any
additional downward rounding. Existence-preserving removals leave a safe
policy satisfying the same necessary conditions. A contradiction to those
conditions rules out the assumed policy. A safe nonempty set-valued policy
would contain a safe deterministic policy, so it is ruled out as well.
\Halmos
\end{proof}

The clinical screen propagates these direct loss bounds backward in time.
Let $\underline\rho_{k,T}=0$, and let
$\mathcal P_{k,tsa}$ be the point enclosure's box of transition rows
intersected with the probability simplex. At a position $(t,s)$, compute
\begin{equation}
 \zeta_{k,tsa}=\underline\ell_{k,tsa}
       +\gamma\min_{p\in\mathcal P_{k,tsa}}
                       \sum_jp_j\underline\rho_{k,t+1}(j).
 \label{eq:enclosed-propagation}
\end{equation}
The endpoint sum $\sum_j\underline P_{k,tsa,j}
\underline\rho_{k,t+1}(j)$ is also a valid, possibly smaller continuation
bound. Remove action $a$ when $\zeta_{k,tsa}>\delta$ for any listed
model. For the surviving set $A'_t(s)$, retain the bounds
$\underline\rho_{k,t}(s)=\min_{a\in A'_t(s)}\zeta_{k,tsa}$.
An empty surviving action set at a living position is an infeasibility certificate. The death
states have zero loss and require no separate loss restriction.

\begin{proposition}[Sound backward elimination]
\label{prop:enclosed-propagation}
Backward elimination using \cref{eq:enclosed-propagation} removes no
action used by a policy satisfying the loss limit at every position in
every listed model. At each completed position, its retained bounds lie
below that policy's cumulative losses.
\end{proposition}
\begin{proof}{Proof.}
The terminal bounds are exact. Suppose the claim holds at the next
epoch and let $\sigma$ be any safe policy. Its true transition row belongs
to $\mathcal P_{k,tsa}$. Nonnegativity and
\cref{eq:finite-loss-recursion} give
$\rho^\sigma_{M_k,t}(s)\ge\zeta_{k,ts,\sigma_t(s)}$.
A strict bound above $\delta$ therefore excludes that action from every
safe policy. The safe policy's action remains among the survivors, so
minimizing over survivors preserves a lower bound on its loss. Backward
induction proves both claims. If a position empties, no safe policy could
have supplied its required action.\Halmos
\end{proof}

The minimum is taken separately for each model only to obtain lower
bounds. The argument does not assert that the minimizing actions form a
common policy. Every removal records its model, position, action, and
strict rational inequality. The checker reconstructs the bounds from the
point enclosures and checks row minimization in exact arithmetic.

\subsection{Bounds that couple two listed models}
\label{app:pair-loss-certificate}

A weighted combination of two models' losses retains a further common-action
constraint. Choose two listed models, write them as $1$ and $2$, and use
rational weights $w_j=j/N$ for $j=0,\ldots,N$. Let
$b_{t,s,j}\ge0$ be proposed lower bounds on
$w_j\rho^\sigma_{1,t}(s)+(1-w_j)\rho^\sigma_{2,t}(s)$, with
$b_{T,s,j}=0$. Write $\mathcal B_{t,s}(w)$ for the linear interpolation
of these bounds between adjacent weights. For a fixed weight $w$ and
action $a$, define, for each successor $s'$, the nonnegative masses
\[
 x_{s'}=w\underline P_{1,tsa,s'},\qquad
 y_{s'}=(1-w)\underline P_{2,tsa,s'},\qquad
 m_{s'}=x_{s'}+y_{s'}.
\]
Terms with $m_{s'}=0$ contribute zero. At each grid weight, require
\begin{equation}
 b_{t,s,j}\le
 w_j\underline\ell_{1,tsa}+(1-w_j)\underline\ell_{2,tsa}
 +\gamma\sum_{s':m_{s'}>0}m_{s'}\,
       \mathcal B_{t+1,s'}(x_{s'}/m_{s'})
 \label{eq:pair-loss-subsolution}
\end{equation}
for every original action. All quantities in this inequality are rational.

\begin{proposition}[Common-policy pair bound]
\label{prop:pair-loss-subsolution}
Bounds satisfying \cref{eq:pair-loss-subsolution} for every original
action lie below the indicated weighted loss of every deterministic
policy. Any bound exceeding $\delta$ at a living position is therefore
an infeasibility certificate.
\end{proposition}
\begin{proof}{Proof.}
Fix a policy. Its weighted loss is linear in the weight, so lower bounds
at adjacent weights interpolate to a lower bound at every intervening
weight. Assume the claim holds at epoch $t+1$. Each summand on the
right of \cref{eq:pair-loss-subsolution} is at most
$x_{s'}\rho^\sigma_{1,t+1}(s')+
y_{s'}\rho^\sigma_{2,t+1}(s')$.
The lower probability endpoints and nonnegative losses bound this by the
corresponding term with the true probabilities. Applying the inequality
to the policy's shared action and using the two exact loss recursions
proves the claim at epoch $t$. The terminal case starts the induction.
If both model losses were at most $\delta$, every convex combination
would be at most $\delta$, proving the final assertion.\Halmos
\end{proof}

The same inequalities support elimination. At each position, remove an
action if the right-hand side exceeds $\delta$ at some grid weight.
For surviving actions, check \cref{eq:pair-loss-subsolution} at every
weight. The preceding induction, now restricted to policies that are
safe at all later positions, shows that these removals preserve every
safe policy. An empty action set proves infeasibility. The checker
verifies every removed action's strict witness and every surviving
action's subsolution inequalities on the original action domain.

\subsection{Checked contradictions and existence bounds}
\label{app:checked-existence-bounds}

When backward elimination alone is inconclusive, a finite action-branch
tree can certify a contradiction to the necessary conditions. These
conditions may include an exclusion of a complete policy whose loss has
already been proved to exceed $\delta$ in a listed model. At each node,
rational propagation justifies every removal in its current domain.
An internal node branches on an undecided position and has one child
for every remaining action, with no action omitted. A leaf either has
an empty action domain or fixes a complete policy whose point-model loss
has a checked strict lower bound above $\delta$. The latter leaf rules
out only that fixed policy. Induction from the leaves to the root shows
that a complete checked tree rules out every safe policy.
The checker reconstructs the formula from the point enclosures, checks
each domain reduction, and checks complete branch coverage. A solver's
infeasibility status alone is not an infeasibility certificate.

For unresolved patients, the bounds concern $d^\star$ in
\cref{sec:master-construction}, rather than the loss of a particular
candidate. One unconditional lower bound is
\[
 \max_{(t,s)\text{ living}}\min_{a\in\Aset_t(s)}
                  \max_k\underline\ell_{k,tsa}.
\]
An original-domain pair subsolution also gives an unconditional lower
bound. Propagated bounds obtained after threshold-dependent removals
apply directly to policies safe at $\delta$. To report such a number
as an unconditional lower bound on $d^\star$, cap it at $\delta$.
Every other policy already exceeds $\delta$, which proves the capped
bound. An empty action set or a checked branch contradiction establishes
$d^\star>\delta$ without supplying a numerical margin beyond the limit.
A complete-cover bound for any admissible singleton policy gives an
upper bound on $d^\star$. The smallest such checked bound is retained.
Pointwise rejection of one candidate supplies no lower bound on
$d^\star$. Each reported endpoint is backed by its own checked evidence.

\FloatBarrier
\section{Hypertension models and supplementary experimental specifications}
\label{app:continuous-experiment-details}

The hypertension experiment examines how uncertainty in future blood pressure
and cholesterol affects the treatment choices that can be retained. This
appendix describes the model, uncertainty sets, treatment \mbox{rules, and experimental comparisons}.

Table~\ref{tab:clinical-settings} gives the source inputs, and Table~\ref{tab:clinical-parameter-guide} specifies their mathematical roles.
\begin{table}[!htbp]
\centering
\caption{Inputs to the hypertension decision models and their sources.}
\label{tab:clinical-settings}
\begin{tabularx}{\linewidth}{@{}L{0.19\linewidth}YL{0.25\linewidth}@{}}
\toprule
Input & Specification & Source \\
\midrule
Source cohort & 2009--2016 NHANES. The eligible sample represents 66.5 million adults, including 32.0 million choice-capable adults & \citeauthor{nhanes} (\citeyear{nhanes}, \citeyear{nhanesweighting}) \\
Horizon and discount & 10 annual decision years. Terminal continuation follows the last year. $\gamma=0.97$ & Model specification of \citet{garciahypertension} \\
Health states & 10 per patient and year, comprising 6 living and 4 death states & Model specification of \citet{garciahypertension} \\
Actions & Up to 196 antihypertensive drug combinations before admissibility restrictions & Model specification of \citet{garciahypertension} \\
Risk equation & Revised pooled cohort equation & \citet{yadlowsky2018} \\
Treatment effects & Class-specific BP-response estimates and age-dependent event relative risks & \citet{law2003,law2009} \\
Medication burden & Combination-specific disutility using the inherited base burden and class symptom frequencies & \citet{law2003,sussman2013,schell2016}, with the adaptation in \citet{garciahypertension} \\
Quality factors & Age- and health-state quality-of-life weights & \citet{kohlilynch2019} \\
Life expectancy & Age- and sex-specific remaining life expectancy and mortality & \citet{ariasxu2019} \\
Terminal continuation & Remaining life expectancy times a health-state mortality adjustment and terminal quality factor & \citet{ariasxu2019,pandya2015,kohlilynch2019} \\
\bottomrule
\end{tabularx}
\par\smallskip{\footnotesize\noindent The sample contains 4,590 patients, including 2,443 choice-capable patients and 2,147 with only no treatment at every living health state and year. Eligibility follows the inherited age, race, and reported-event exclusions. The population totals are our sums of pooled NHANES examination weights for these groups. The model retains the inherited numerical inputs and adaptations. The cited clinical studies support the parameter choices within that model.\par}
\end{table}

\begin{table}[tbp]
\centering
\small
\caption{Inputs to the primary hypertension experiment. Patient-specific values come from the model and forecast data.}
\label{tab:clinical-parameter-guide}
\begin{tabularx}{\linewidth}{L{0.21\linewidth}L{0.41\linewidth}Y}
\toprule
Input & Value or specification & Source within this companion\\
\midrule
Decision horizon & Ten annual decisions. Terminal continuation follows the final year. & \Cref{app:clinical-risk-bridge}\\
Discount factor & $\gamma=0.97$. & \Cref{app:clinical-risk-bridge}\\
States and actions & Ten health states, six living. 196 drug combinations before admissibility restrictions. & \Cref{app:clinical-risk-bridge,app:clinical-common-feasibility}\\
Nominal forecasts & Supplied patient-year centers $c_{ijt}$ for systolic BP, total cholesterol, and high-density lipoprotein cholesterol. & \Cref{app:clinical-factor-bands}\\
Forecast bands & Positive endpoints $L_{ijt},U_{ijt}$ from reconstructed covariance and residual scales. Baseline factors remain fixed. & \Cref{eq:clinical-factor-scale,eq:clinical-factor-map}\\
Shared parameters & $z\in[-1,1]^3$. Primary width $w=1$ and sensitivity widths $0,0.5,2$. Nominal model at $z=0$. & \Cref{app:clinical-factor-bands}\\
Untreated event risk & Supplied revised pooled cohort equation with annual multiplier $0.082$. 70\% coronary heart disease and 30\% stroke. & \Cref{app:clinical-risk-bridge}\\
Treatment response & Source BP reductions and age-dependent event relative risks, evaluated at nominal forecasts and fixed across the primary family. & \Cref{app:clinical-common-feasibility}\\
Annual medication burden & Source combination-specific $\beta_a$, fixed across the primary family. Immediate reward $\max\{0,u_{its}-\beta_a\}$. & \Cref{app:clinical-common-feasibility}\\
Mortality and quality of life & Source age-, sex-, and health-state inputs determine transitions and quality factors. & \Cref{app:clinical-risk-bridge,app:clinical-common-feasibility}\\
Terminal value & Source remaining life expectancy, mortality adjustment, and terminal quality factor at the final decision-row age. & \Cref{app:clinical-risk-bridge,app:dose-horizon-design}\\
Treated-action thresholds & Nominal post-treatment systolic BP at least 120 mmHg and diastolic BP at least 55 mmHg. Additional model restrictions apply. & \Cref{app:clinical-common-feasibility}\\
No-treatment rule & Nominal systolic BP at most 150 mmHg and diastolic BP at most 90 mmHg. The model permits no treatment when all actions fail these conditions. & \Cref{app:clinical-common-feasibility}\\
Loss limits & Primary limit $\delta=0.03$ QALYs. Larger limits 0.05 and 0.07 for sensitivity, at every living health-state/year pair. & \Cref{app:clinical-enclosures}\\
\bottomrule
\end{tabularx}
\par\vspace{4pt}
\vspace{2pt}
\parbox{\linewidth}{\footnotesize The source model is described by \citet{garciahypertension}. The risk equation follows \citet{yadlowsky2018}. The forecast bands specify the sensitivity analysis.}
\end{table}

\subsection{Source cohort, reference model, and risk calculation}
\label{app:clinical-risk-bridge}

The source hypertension model supplies 4,590 patients from NHANES \citep{nhanes}, ten annual
profiles per patient, ten health states, and 196 ordered treatment
combinations before admissibility restrictions \citep{garciahypertension}.
The model inputs include all 4,590 source patients and their
45,900 forecast rows. Clinical result summaries restrict the analysis to
the 2,443 choice-capable patients, identified from source action sets before
examining search outcomes. The decision horizon is ten years and the discount
factor is 0.97, interpreted exactly. Six states represent living
health conditions. With ten annual decisions, they give 60 living health-state/year pairs per patient. We use the source model's event-history, mortality, quality-of-life,
and terminal life-expectancy conventions. The supplied
forecast table determines the centers of the uncertainty family.

Untreated cardiovascular risk uses the supplied \texttt{rev\_arisk}
implementation of the revised pooled cohort equation
\citep{yadlowsky2018}. The experiment fixes this implementation's printed coefficients,
annual multiplier $0.082$, and 70\%/30\% allocation to coronary heart
disease (CHD) and stroke. The
source race code is converted to the calculator's Black indicator as
$1-\texttt{race}$, and the treated indicator is zero. Treatment benefits
are subsequently applied once in the transition construction. This
interpretation treats the supplied risk factors as untreated values, as in
the source model.
With fixed demographic indicators and age in decision epoch $t$, the annual equation
can be written
\begin{equation}
 q_{it}(S,C,H)
 =0.082\,\operatorname{logistic}
   \left(a_{it}+b_iS^2+c_{it}S+e_i\frac{C}{H}\right),
 \qquad \operatorname{logistic}(u)=\frac{1}{1+\exp(-u)},
 \label{eq:clinical-factor-risk}
\end{equation}
where $S$ denotes systolic BP in mmHg, and $C,H$ denote
total and HDL cholesterol in mg/dL. The grouped coefficients in
\eqref{eq:clinical-factor-risk} are obtained from the supplied
implementation, including its demographic interactions. The model evaluates each epoch's probability directly,
without adjusting the ten annual probabilities to reproduce \mbox{a baseline ten-year prediction}.

We flag profiles outside age 40--79 inclusive,
$90<S<200$, $130<C<320$, or
$20<H<100$. The flags identify values outside these ranges without excluding patients or
changing their recorded values. There are
313 flagged patients. Of these, 292 are flagged at baseline and 21 only
in a later forecast epoch. These patients use the risk equation outside
its stated covariate ranges.

\subsection{Survey weights and the analysis population}
\label{app:clinical-survey-weights}

The source extraction combines the four 2009--2016 NHANES cycles, retains
non-Hispanic Black and White adults aged 40--60 at baseline, and excludes
affirmative reports of myocardial infarction or stroke. The source code
calls the myocardial-infarction variable \texttt{CHD}, but its NHANES item
is \texttt{MCQ160E}, rather than an indicator for all coronary disease.
The stroke item is \texttt{MCQ160F}. The cohort is defined by these
\mbox{recorded exclusions \citep{nhanesmedicalconditions}}.

Each source weight is the two-year examination weight divided by four,
following the NHANES rule for pooling these cycles \citep{nhanesweighting}.
Write $\varpi_i$ for this weight, counted once for each patient $i$, rather
than once for each of its ten forecast rows. For an analysis set $I$ and
a subset $J\subseteq I$, the represented population is
$\sum_{i\in I}\varpi_i$, and the weighted share is
$\sum_{i\in J}\varpi_i/\sum_{i\in I}\varpi_i$.
An outcome $y_i$ has weighted mean
$\sum_{i\in I}\varpi_i y_i/\sum_{i\in I}\varpi_i$.
Weighted quantiles linearly interpolate the sorted outcome values at
midpoint cumulative weights. Records with zero weight contribute neither
to these means nor to the quantile interpolation.

Survey weights represent 66,504,886.84 adults, including 32,028,141.06
choice-capable adults and 34,476,745.78 whose action sets force singleton plans.
There are 106 zero-weight patients in the full source, including 86 in the
choice-capable group. They remain in the computational census and have
zero contribution to population summaries. The source file omits the
survey strata and primary sampling units needed for design-based standard
errors. The reported survey summaries are weighted point estimates and
descriptive quantiles.

A patient is choice-capable if the nominal source action sets permit at least two actions at some decision epoch and living health state. This classification is independent of certification. Certification shares use this
group as their denominator. Compliant treatment plan summaries further condition on the
same certified-flexible patients for both rules, with class-avoidance
comparisons restricted to patients admitting that class restriction.
These are nominal-model predictions for the eligible source population.
The input-based sensitivity panel and its summary measures are defined in \cref{app:clinical-panel-design}.

\subsection{Regression reconstruction and forecast bands}
\label{app:clinical-factor-bands}

The supplied forecast centers are retained, while a reconstruction of the
NHANES preprocessing and regression workflow supplies the uncertainty scales
\citep{nhanes}. The reconstruction combines the source data with public
NHANES inputs. Medication classification includes antihypertensive classes
outside the five offered by the decision model and retains the source
prescription-count and combination coding conventions. Missing values are filled using random-forest imputation. The patient records are
then expanded using rounded survey weights and used to generate a synthetic
dataset. Regressions are fitted to the synthetic records labeled as untreated.
The imputation and synthesis seeds are 123 and 456, respectively. Synthetic records represent the source inputs rather \mbox{than additional independent observations}.

Five factor models are fitted for systolic BP, diastolic BP,
total cholesterol, HDL, and LDL. The candidate regressors are age, squared
age, sex, race, smoking, and diabetes, with bidirectional stepwise
selection followed by least squares. Fitted coefficients, design coding,
coefficient covariances, residual scales, and fitted objects specify the
reconstruction. An independent
least-squares fit using \texttt{biglm} checks each selected formula's
predictions.
For individual $i$, factor $j$, and decision epoch $t$, write $c_{ijt}>0$ for the
supplied center and $\Delta x_{ijt}$ for the change from baseline in the selected
regression design vector. Demographic indicators are fixed, so only the
included age terms change. Let $\widehat V_j$ and
$\widehat\sigma_j^2$ denote the fitted coefficient covariance and residual
variance. The uncertainty scale is
\begin{equation}
 s_{ijt}^2=(\Delta x_{ijt})^{\top}\widehat V_j\Delta x_{ijt}
             +2\widehat\sigma_j^2\quad(t>0),
 \qquad s_{ij0}^2=0.
 \label{eq:clinical-factor-scale}
\end{equation}
The first term allows uncertainty in the fitted change from baseline. The
second treats the change's two residuals as uncorrelated with equal variance.
The fitted covariance covers the selected regression coefficients. The
uncertainty scales treat the imputation, synthesis, and model-selection steps
as fixed.

For $t>0$, take the 2.5th and 97.5th percentiles of
$N(c_{ijt},s_{ijt}^2)$ conditional on positivity, expanding either endpoint
when necessary to include $c_{ijt}$. Denote the resulting positive
endpoints by $L_{ijt}\le c_{ijt}\le U_{ijt}$. At baseline,
$L_{ij0}=U_{ij0}=c_{ij0}$. These quantile-based endpoints define forecast bands. Simultaneous predictive coverage across years and factors has not been established.
The recorded numerical endpoints become fixed inputs to the certified
model family.

One latent coordinate $z_j\in[-1,1]$ per factor links all future epochs through
\begin{equation}
 x_{ijt}(z_j;w)=c_{ijt}
 \begin{cases}
 \exp\{wz_j\log(U_{ijt}/c_{ijt})\},&z_j\ge0,\\
 \exp\{-wz_j\log(L_{ijt}/c_{ijt})\},&z_j<0.
 \end{cases}
 \label{eq:clinical-factor-map}
\end{equation}
The primary family uses the three coordinates for systolic BP, total cholesterol,
and HDL, with $z\in[-1,1]^3$ and $w=1$. Each coordinate is shared across
epochs, states, and actions, so the family does not allow a separate
deviation at every health state and year. Baseline factors remain fixed.
Widths $w\in\{0,1/2,1,2\}$ respectively give the point model, a narrower
family, the reference family, and a wider family. Positivity is preserved
by the logarithmic map. At $w=1$, $z_j=-1,0,1$ give the lower endpoint, center, and upper endpoint, respectively. The domain $[-1,1]$ therefore normalizes the forecast band. Increasing $w$ multiplies log displacement from the center. At $w=2$ that displacement doubles. The families are nested as width increases.

\subsection{Treatment response, transitions, and admissibility}
\label{app:clinical-common-feasibility}

The primary family varies untreated risk while keeping treatment responses
fixed. The inherited response functions use class-specific estimates from Table~2 of \citet{law2003} and the age-specific risk slopes of \citet{law2009}. Predicted systolic BP and diastolic BP reductions $D^S_{ita}$ and
$D^D_{ita}$, medication burdens $\beta_a$, and admissible action sets are therefore
evaluated at the reference forecasts. For event $e$, the treatment
response is
\[
 E_{itae}=\exp\{\log(RR_{ite})D^S_{ita}/20\},
\]
using the source age-dependent relative risk $RR_{ite}$. For a state with
history multiplier $m$, untreated event risk $p$ is transformed to
$mp/\{1+(m-1)p\}$, then multiplied by $E_{itae}$. The immediate reward
is $\max\{0,u_{its}-\beta_a\}$, where $u_{its}$ is the inherited quality factor \citep{kohlilynch2019}. The inherited medication burden uses class symptom frequencies from \citet{law2003} and the source model's base disutility, retaining its adaptation of \citet{sussman2013,schell2016}.

The source model assigns transition probabilities first to fatal stroke,
then to fatal CHD, other death, surviving stroke, surviving CHD, and finally
to the remaining history state. Each probability is limited to the probability
mass still available. The interval calculations include these limits
whenever the probabilities before clipping might sum to more than one.
The source death states lead to the terminal death state with zero
subsequent reward.

We use the source rules to determine which actions are admissible. A treated
action requires reference post-treatment systolic BP at least 120, reference
post-treatment diastolic BP at least 55, nonnegative reference systolic BP, and a
nonnegative diastolic BP reduction. The no-treatment action is admissible when reference
systolic BP is at most 150 and diastolic BP is at most 90. If these rules leave no action,
the source fallback permits no treatment. Dead-state action sets contain
only no treatment. This source-model fallback occurs for 50 patients in
268 modeled patient-years and makes every primary action set nonempty.
Admissibility is evaluated at the reference forecasts, and the
complete-family certificate bounds QALY loss on those fixed sets.

The least systolic-BP-reducing single-drug option in the model is an ACE inhibitor, with reduction $8.5+0.1(S-154)$ at nominal systolic BP $S$. Its post-treatment floor requires $0.9S+6.9\ge120$, so treatment requires nominal systolic BP of approximately 126 mmHg or above. Of the 2,147 patients without choice, 2,088 never reach this threshold and the other 59 fail the diastolic-BP rules whenever the systolic rule permits treatment. Among the 2,443 choice-capable patients, 1,865 have alternatives at baseline and 578 first gain them in a later year. This classification uses the input action sets and precedes optimization.

A separate four-factor sensitivity analysis varies diastolic BP, recomputes treatment
responses, and intersects the action sets across the family. For the
drug-class action set, affine response slopes follow the source drug order
and printed coefficients. Recorded reference reductions and
post-treatment pressures anchor the extension, preserving the source
comparisons at the reference profile. The no-treatment fallback is
applied within each model before the common action set is formed. We count treatments excluded by common admissibility and empty common action sets at all four
widths for all 4,590 patients. An empty common set means that no
single action at a required health state and year is admissible throughout
this declared family.
The forecast bands can include systolic BP below diastolic BP because the
coordinates vary separately. We flag such combinations.
For the common 244-patient panel defined in
\cref{app:clinical-panel-design}, the responsive analysis
searches for certified treatment plans at $w=1$. Point diagnostics compare the
common-action optimum with model-specific-action optima at the center
and the 16 corners. These point comparisons show how the optimal benchmark changes when the
action sets change. Certification separately checks the entire family.

\subsection{Numerical inclusion and certificate checks}
\label{app:clinical-enclosures}

Clinical treatment plans use the joint bound on complete risk-score covers (\cref{app:clinical-risk-cover}). A cell encloses rewards, transitions, terminal values, and action contrasts throughout its parameter region. The monotone map in \eqref{eq:clinical-factor-map} bounds each factor on a cell. The quadratic systolic BP term requires checking endpoints and any interior stationary point, and the positive cholesterol ratio is bounded on its factor box. Logistic monotonicity then gives an annual-risk enclosure.

Interval arithmetic encloses logarithms and exponentials with outward rounding \citep{johanssonarb}. Printed coefficients are exact decimal numbers, while binary64 inputs and forecast-band endpoints are the exact rational numbers represented by their bits. Conversion to binary64 moves each bound outward when necessary. Bellman calculations optimize transition vectors over their interval bounds subject to probabilities summing to one, using exact rational arithmetic. Table~\ref{tab:clinical-computation-settings} lists arithmetic precision and computation limits.

The joint action contrast preserves the shared untreated-risk term. For unclipped transitions, the treatment-effect difference multiplies the same history-adjusted event risk. Same-action contrasts are exactly zero, and clipped rows use conservative interval bounds. In the responsive analysis, exponential affine response differences include any interior stationary point as well as endpoints.

The construction uses the initial policies and fixed-order augmentation in Algorithm~\ref{alg:family-search}. A certified incumbent can initialize the procedure. Each accepted addition satisfies the complete-cover test, and a separate check validates the final cover and action-addition trace.

When every action set is singleton, the unique treatment plan is each model's optimum on those sets and has zero loss. Such input-forced singletons are distinguished from certified singleton treatment plans on inputs that permit choice. Infeasibility requires the exact proofs in \cref{app:clinical-existence}. For responsive common action sets, the checks reconstruct the admissibility diagnosis as well as the cover. Statistical proposal checks keep the proposed actions fixed, while subset construction restricts additions to those actions and keeps the unrestricted benchmark.

\subsubsection{Risk-score construction and checks}
\label{app:clinical-risk-cover}

The clinical cover retains dependence across epochs through a reference risk score. Every action loss is measured against the optimum on the full fixed admissible action sets. With fixed treatment response, uncertain inputs enter each future epoch's transition model through untreated event risk. For patient $i$, write the logistic score as
\[
 \psi_t(z)=a_{it}+b_iS_t(z)^2+c_{it}S_t(z)+e_iC_t(z)/H_t(z),
\]
where $S_t,C_t,H_t$ denote systolic BP, total cholesterol, and HDL cholesterol. The coefficients $a_{it},c_{it}$ can vary across epochs, while $b_i,e_i$ are fixed for each patient, as in \eqref{eq:clinical-factor-risk}. The baseline epoch is fixed.

For each cover, a fixed reference year $r$ is chosen and $\psi_r(z)$ is calculated from the same forecast factors and risk coefficients. The primary covers use reference years 4, 6, or 8 (counting the initial decision year as year 1). This choice changes only the enclosing geometry.

A sign orthant is one of the eight regions of $[-1,1]^3$ in which each coordinate is restricted to $[-1,0]$ or $[0,1]$. Within one orthant, let $B_j$ be a finite cover by parameter subboxes, let $R_j$ enclose the reference score $\psi_r(B_j)$, and let $D_{tj}$ enclose $(\psi_t-\psi_r)(B_j)$. For a score interval $I$, the epoch score is enclosed by
\[
\operatorname{hull}_{j:\,R_j\cap I\ne\varnothing}\bigl((I\cap R_j)+D_{tj}\bigr).
\]
Every parameter point whose reference score belongs to $I$ lies in one of the included subboxes. The enclosure therefore contains its score, as proved in \cref{lem:risk-score-cover}. Logistic monotonicity bounds the annual risk. Complete coverage of the parameter orthants and their score ranges then gives a complete cover of the family. Structural variants use their specified transition, reward, and terminal equations. Generic-dose action sets use their relative-risk products.

The construction starts from the sign orthants and refines score intervals or parameter boxes. A reference epoch and an outward-rounded partition determine score and cross-epoch difference bounds. The allowed precisions, partition sizes, and refinement limits appear in Table~\ref{tab:clinical-computation-settings}. Each accepted action addition and final treatment plan satisfy the joint complete-cover bound. For statistical subsets, additions are restricted to the proposed sets, while the optimal-value benchmark uses the full admissible action sets.

Certificate checks establish parameter-domain coverage, score subdivision coverage, interval inclusion, and the exact loss threshold, with the model family reconstructed for each certificate. An independently written score-enclosure calculation uses a finer partition and the same outward-rounded Bellman computation as the construction. We check point witnesses and the action additions separately. Infeasibility proofs are checked through their exact arithmetic witnesses as described in \cref{app:clinical-existence}. The responsive family uses parameter-box cells that also enclose its uncertain treatment responses and admissibility.

\subsection{Unresolved existence bounds}
\label{app:unresolved-existence-bounds}

For each of the 47 unresolved patients, representing 703,065 adults, lower and upper bounds enclose the smallest achievable worst-case loss among common deterministic treatment plans. A checked finite-list argument supplies the lower bound, and a checked complete-cover evaluation of an admissible treatment plan supplies the upper bound. Every interval contains 0.03 QALYs. Lower bounds range from 0.0247 to 0.0300 QALYs, with upper bounds from 0.0300 to 0.0345 QALYs. These bounds are rounded to the nearest value at four decimal places. A bounded exact-search procedure ends without a decision for 37 patients, representing 531,663 adults. Computation ends during a search or check for ten patients, representing 171,402 adults. Feasibility at 0.03 QALYs remains undetermined. The verified upper bounds establish a 0.035-QALY guarantee. Table~\ref{tab:clinical-computation-settings} reports the bounds and applicable limits.

\begin{figure}[t]
\centering
\includegraphics[width=\linewidth]{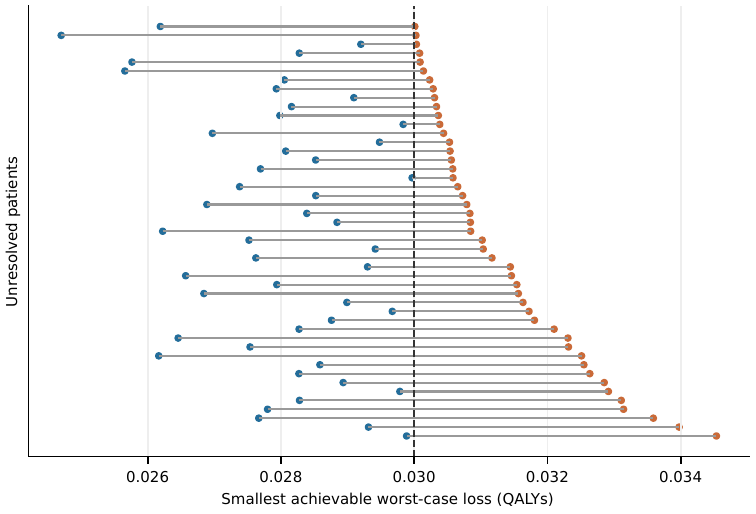}
\caption{Bounds on the smallest achievable worst-case loss for the 47 unresolved patients. Each segment joins the finite-list lower bound to the complete-cover upper bound of a treatment plan, ordered by upper endpoint. The dashed line marks 0.03 QALYs.}
\label{fig:unresolved-existence-bounds}
\end{figure}

\subsection{Sensitivity panel}
\label{app:clinical-inventory}
\label{app:clinical-panel-design}

The fixed panel contains 244 of the 2,443 choice-capable patients. Selection uses nominal annual risk, age, sex, race, source action-set size, relative forecast-band width, the covariate-range flag defined in \cref{app:clinical-risk-bridge}, and how often the no-treatment fallback applies. These eight inputs are converted to percentile ranks. Selection starts with nine distinct patients representing input extremes or flagged inputs, then repeatedly selects the patient with the largest minimum Euclidean distance to the selected set. The random seed 2026091701 fixes ties. This input-based rule uses no certification outcomes.

All width, responsive, dose, horizon, post-event, and clinical statistical settings use these same patients. The panel summarizes the selected patients with unweighted counts and percentages. Its input-based selection supports exploratory comparisons across settings. National population summaries use the full choice-capable cohort. Treatment-plan size is summarized among patients with a certified flexible treatment plan. Counts of distinct patients with any valid certificate at a setting give lower bounds on flexibility, including certificates valid for a wider family or a smaller loss limit with the same action sets and benchmark.

\subsection{Dose, horizon, and structural sensitivity panel}
\label{app:dose-horizon-design}

Dose sensitivity varies the available numbers of standard- and half-dose medications. The two generic action sets use the inputs in Table~\ref{tab:generic-dose-inputs}.
Blood-pressure reduction, cardiovascular risk reduction, and treatment
burden follow the modeling approach used in hypertension studies
\citep{sussman2013,schell2016}, with BP-lowering evidence providing clinical
motivation \citep{sundstrom2014,sundstrom2015}. These generic values are assumptions used only in this sensitivity experiment.
\begin{table}[tbp]
\centering
\small
\caption{Specified inputs for the generic-dose sensitivity experiment.}
\label{tab:generic-dose-inputs}
\begin{tabular}{lrr}
\toprule
Quantity & Standard dose & Half dose\\
\midrule
systolic BP reduction (mmHg) & 5.5 & 3.7\\
diastolic BP reduction (mmHg) & 3.3 & 2.2\\
CHD relative risk & 0.87 & 0.93\\
Stroke relative risk & 0.79 & 0.86\\
Annual QALY burden & 0.002 & 0.001\\
\bottomrule
\end{tabular}
\end{table}
For $n$ standard-dose and $h$ half-dose agents, BP reductions and QALY
burdens add, while event-specific relative risks multiply. In particular,
\[
 D^S=5.5n+3.7h,\qquad D^D=3.3n+2.2h,\qquad
 b=0.002n+0.001h,
\]
with CHD multiplier $0.87^n0.93^h$ and stroke multiplier $0.79^n0.86^h$.
These effects remain fixed as the untreated risk factors vary. The
multiplicative response, its application to the broader CHD outcome, and
the two-to-one standard/half-dose burden ratio are explicit modeling
assumptions. The baseline burden in \citet{sussman2013} and
\citet{schell2016} is 0.001 per medication per year.

The 21-action set contains the nonnegative integer pairs $(n,h)$ with
$n+h\le5$. The six-action set is its $h=0$ subset, using
identical equations for shared actions. Medication count $n+h$ and nominal
dose equivalent $n+h/2$ are distinct quantities. These generic action sets
have no drug-class identities. The drug-class 196-action set is a
class-specific response representation.

For each panel patient, the dose design combines two action sets, horizons of five and ten epochs, and three loss limits. The five-epoch model uses the first five profiles and the source terminal factor at its last decision-row age, discounted by $\gamma^5$. The ten-epoch drug-class model separately varies zero terminal reward, absorbing post-event living states, zero living post-event reward, and a 60/40 coronary--stroke allocation. Each setting uses the joint risk-score construction with the limits in Table~\ref{tab:clinical-computation-settings}.

\subsection{Decision outcomes and clinical interpretation}
\label{app:decision-outcome-design}

The performance guarantee permits different choices to have different
outcomes within the loss limit. We evaluate these differences in the
nominal model, starting each person in the healthy state and using the
certified treatment plans.
Separate restricted dynamic programs compute the highest and lowest expected QALYs allowed by each certified treatment plan, using their own continuation values. One selection rule chooses the fewest current medications. Another chooses the lower median
allowed action, ordered by medication count, burden, and action identifier. Two others separately minimize expected cumulative undiscounted medication disutility and expected regimen switches. Class-avoidance rules maximize
nominal QALYs while avoiding each drug class. The two cumulative cost
objectives are optimized by dynamic programming over the decision horizon.
The median is taken over the ordered allowed actions, with each action
retained separately even when several have the same medication count. Class avoidance
is unavailable when a required menu becomes empty. There is no
fallback to an unrestricted plan.
Switch minimization augments the state by the previously realized action.
Initial treatment is unobserved, so no initial switch cost is assigned,
and no switch cost accrues after death. Medication costs accrue while
alive. Minimizing them alone can favor shorter survival, so accompanying
QALYs and event outcomes are necessary for interpretation. Independent
nominal forward and backward evaluations check agreement. We compute these nominal-model outcomes in floating point, without certified bounds.
Four illustrative treatment plans are selected by the rule in \cref{app:representative-patients}. Their paths show choices conditional on remaining healthy.

First-event incidence uses a separate event-free alive occupancy. First
CHD or stroke events leave this risk set, competing other-cause death
is recorded separately, and only event-free survivors remain at risk.
Post-event mass cannot re-enter. First-event incidence, competing death
before the first event, and event-free survival at the horizon sum to one.
Restricted event-free occupied years sum event-free alive probabilities
at the starts of the decision intervals. End-of-interval totals are also
recorded. Repeated visits to event states are not counted as first events.
Discounted finite-horizon and terminal lifetime QALYs are reported
separately.

Our implementation of the guidelines applies the source model's medication-count
adaptation of the 2017 guideline \citep{wheltonguideline2017} on the full admissible
action sets. Its risk input is the selected calculator's ten-year risk with the
source history multipliers. Counts map to the
lowest-burden matching combination, with source action order breaking
ties. When that count is unavailable, the comparison uses the nearest admissible count. Guideline calculations assume zero initial medications.
Among adults with certified flexible treatment plans, evaluated from the healthy initial state under the nominal model, the best selection gains a mean of 0.072 QALYs over the guidelines, with median 0.042. The worst selection gains a mean of 0.066 QALYs, with median 0.036.
Across-person summaries use the pooled survey weights and matched
individuals for compliant treatment plan comparisons, as specified in
\cref{app:clinical-survey-weights}.

\FloatBarrier
\section{Supplementary Clinical Results}
\label{app:clinical-results}

This appendix reports population outcomes, model sensitivities, and the computation settings supporting their certificates. The primary analysis includes all 2,443 choice-capable patients and uses survey weights to represent 32.0 million adults. Sensitivity results use the panel defined in \cref{app:clinical-panel-design}.

\subsection{Primary outcomes and treatment flexibility}

At the primary loss limit, 2,300 patients have a certified treatment plan, comprising 2,283 flexible treatment plans and seventeen singletons. Infeasibility is proved for 96 patients, and 47 remain unresolved. Certified treatment plans account for 94.4\% of represented adults, with 93.6\% having a flexible treatment plan (Table~\ref{tab:clinical-outcomes}). All four classes remain in the population denominator. Among flexible treatment plans, the weighted median is 3.45 actions per living health-state/year pair and 35 flexible health-state/year pairs out of 60. For every flexible treatment plan and four singleton plans, we tried adding each missing action once. For thirteen singleton plans, the computation limit was reached first. Singleton and unresolved patients could still admit a flexible treatment plan, whereas infeasibility certificates exclude it. The demonstrated flexible share is therefore a lower bound on the population share admitting such a plan. Counting singleton and unresolved patients as potentially flexible gives the interval 93.6\%--96.6\%. This interval bounds the share admitting flexible plans. It does not reflect sampling error. Testing all additions once can still leave larger or different flexible plans undiscovered.

\begin{longtable}{@{}lrrr@{}}
\caption{Treatment-plan outcomes in the choice-capable population.}\label{tab:clinical-outcomes}\\
\toprule
Outcome & Patients & Adults (millions) & Share (\%) \\
\midrule
\endfirsthead
\toprule
Outcome & Patients & Adults (millions) & Share (\%) \\
\midrule
\endhead
\bottomrule
\endfoot
Certified flexible & 2,283 & 29.985 & 93.62 \\
Certified singleton & 17 & 0.238 & 0.74 \\
Infeasible & 96 & 1.102 & 3.44 \\
Unresolved & 47 & 0.703 & 2.20 \\
\end{longtable}
{\footnotesize\noindent Survey weights represent 32.03 million adults across 2,443 patients. The loss limit is 0.03 QALYs and uncertainty width is one.\par}

The nominal-model plan starts from a nominal-model optimal treatment plan and uses fixed-order augmentation on that point model. Its evaluated loss is the maximum over living health-state/year pairs and the $3^3=27$ models formed by assigning each of the three forecast factors its low, central, or high value at width one. Outward-rounded loss bounds at a maximizing evaluated model prove rejection for 2,440 patients. For the other three patients, containment in a certified primary treatment plan establishes the family-wide guarantee by \cref{lem:menu-monotonicity}. These classes account for 99.8\% and 0.2\% of adults, respectively. Among rejected treatment plans, the maximum evaluated loss over all living health-state/year pairs has a weighted median of 0.44 QALYs, with quartiles 0.28 and 0.72 QALYs. From the healthy state at the first decision epoch, the maximum over the same 27 models has a weighted median of 0.30 QALYs, with quartiles 0.17 and 0.55 QALYs. For 99.0\% of adults whose nominal-model plan is rejected, corresponding to 2,424 of the 2,440 patients, a checked model gives loss above 0.03 QALYs from that initial health state and year.

\subsection{Patient characteristics}
\label{app:clinical-heterogeneity}

Baseline systolic BP, nominal annual cardiovascular risk, and number of admissible treatments define the strata in Figure~\ref{fig:clinical-flexibility-profile}. The number of admissible treatments is the number of admissible actions summed across all 60 living health-state/year pairs and divided by 60. The groups describe overlapping patient characteristics without multivariable adjustment. Within a treatment plan, the average action count and the count of flexible health-state/year pairs give each pair equal weight. Survey weights then summarize these patient-level measures. Among adults with certified flexible treatment plans, 73.3\% have at least two admissible actions in the healthy state of the first year. The certified plans retain at least two actions there for 98.9\% of that group. The other 26.7\% have a singleton source action set at that initial state and year and gain choices later. Favorable and unfavorable forecasts use $z=(-1,-1,1)$ and $z=(1,1,-1)$, respectively. Survey-weighted healthy-state optimal medication counts over years two through ten average 0.95 under favorable forecasts and 3.47 under unfavorable forecasts for infeasible adults. Higher-risk adults averaging over twenty admissible treatments have corresponding means 3.39 and 3.95, from point evaluations, not certificates.

\subsection{Uncertainty width and the loss limit}
\label{app:clinical-width-results}
\label{sec:clinical-sensitivity}

We vary forecast width to examine how much uncertainty treatment choices can accommodate. We also increase the loss limit to measure how allowing more performance loss affects flexibility. Table~\ref{tab:clinical-width-limits} reports the 244-patient panel for each setting. A certificate remains valid at a larger loss limit or on a contained uncertainty family with the same action sets and benchmark. We use these implications to establish the displayed lower bounds.

All 244 panel patients have certified flexible plans at widths zero and one half under each loss limit. At width one, at least 234 patients have such plans at the 0.03-QALY limit. The count is at least 243 at each larger limit. At width two, the corresponding counts are at least 155, 188, and 192. The primary panel setting also includes seven infeasible patients and three unresolved patients.

For the full choice-capable population, every plan certified at 0.03 QALYs remains valid at 0.05 and 0.07 QALYs. Thus certified flexible plans exist for at least 93.6\% of adults at each larger limit. Complete-cover bounds establish some treatment plan within 0.05 QALYs for at least 97.6\% of adults.

\begin{longtable}{@{}lrrrrr@{}}
\caption{Certified treatment plans by uncertainty width and loss limit.}\label{tab:clinical-width-limits}\\
\toprule
$w$ & $\delta$ & \shortstack{Flexible\\(at least)} & Singleton & Infeasible & Unresolved \\
\midrule
\endfirsthead
\toprule
$w$ & $\delta$ & \shortstack{Flexible\\(at least)} & Singleton & Infeasible & Unresolved \\
\midrule
\endhead
\bottomrule
\endfoot
0 & 0.03 & 244 & 0 & 0 & 0 \\
0 & 0.05 & 244 & 0 & 0 & 0 \\
0 & 0.07 & 244 & 0 & 0 & 0 \\
0.5 & 0.03 & 244 & 0 & 0 & 0 \\
0.5 & 0.05 & 244 & 0 & 0 & 0 \\
0.5 & 0.07 & 244 & 0 & 0 & 0 \\
1 & 0.03 & 234 & 0 & 7 & 3 \\
1 & 0.05 & 243 & 1 & 0 & 0 \\
1 & 0.07 & 243 & 1 & 0 & 0 \\
2 & 0.03 & 155 & 0 & 7 & 82 \\
2 & 0.05 & 188 & 0 & 0 & 56 \\
2 & 0.07 & 192 & 0 & 0 & 52 \\
\end{longtable}
{\footnotesize\noindent Every setting uses the same 244 input-selected patients without survey weighting. Flexible counts are lower bounds. A certificate remains valid for a contained model family or a larger loss limit. Singleton denotes a singleton certificate without a certified flexible treatment plan. Infeasibility and unresolved status concern existence under the specified setting.\par}

\subsection{Responsive treatment effects and common admissibility}
\label{app:clinical-physical-results}

We vary treatment response to examine whether choices remain available when the same treatment has different effects across models. This analysis also varies diastolic BP and requires each allowed action to be admissible under every model at the relevant health state and year. Parameter-box covers enclose the additional uncertain treatment effects (\cref{app:clinical-common-feasibility}).

At width one, eleven of the 244 patients have nonempty common action sets everywhere. The other 233 have an empty common set at some health state and year. One of the eleven has a certified flexible plan at each loss limit, while ten remain unresolved (Table~\ref{tab:clinical-responsive}). The number admitting a flexible plan under common admissibility therefore lies between one and eleven. Requiring a treatment to be admissible across all models can itself restrict choice before the loss guarantee is considered.

\begin{longtable}{@{}lrrrrr@{}}
\caption{Treatment plans with uncertain treatment response.}\label{tab:clinical-responsive}\\
\toprule
$\delta$ & \shortstack{Empty common\\action set} & \shortstack{Flexible\\(at least)} & Singleton & Unresolved & \shortstack{Actions per\\pair} \\
\midrule
\endfirsthead
\toprule
$\delta$ & \shortstack{Empty common\\action set} & \shortstack{Flexible\\(at least)} & Singleton & Unresolved & \shortstack{Actions per\\pair} \\
\midrule
\endhead
\bottomrule
\endfoot
0.03 & 233 & 1 & 0 & 10 & 15.15 \\
0.05 & 233 & 1 & 0 & 10 & 35.37 \\
0.07 & 233 & 1 & 0 & 10 & 43.65 \\
\end{longtable}
{\footnotesize\noindent A pair is a living health state and decision year. Every setting uses the same 244 input-selected patients without survey weighting. The action sets contain treatments admissible throughout the specified four-factor family. Empty common action sets are distinct from infeasibility under the primary fixed action sets. Actions per pair gives the median among certified flexible treatment plans.\par}

\subsection{Dose, horizon, and post-event assumptions}
\label{app:clinical-structure-results}

We vary the available dose combinations and the horizon to examine how treatment options and planning length affect flexibility. Each setting uses the same 244 patients and either six or 21 generic actions (\cref{app:dose-horizon-design}). At 0.03 QALYs, the six-action set gives certified flexible plans to at least 182 patients over five years. The corresponding count over ten years is at least 173. With 21 actions, the counts are at least 230 and 188 for the respective horizons. Input-based upper bounds for the six-action settings are 187 and 222. Those for the 21-action settings are 232 and 240, because other patients have only one admissible treatment everywhere. Table~\ref{tab:clinical-dose-horizon} gives the outcomes at each loss limit.

\begin{longtable}{@{}p{1.65in}rrrrrrr@{}}
\caption{Certified treatment plans by action set and horizon.}\label{tab:clinical-dose-horizon}\\
\toprule
Setting & $\delta$ & \shortstack{Flexible\\(at least)} & Singleton & Infeasible & Unresolved & \shortstack{Actions per\\pair} & \shortstack{Flexible\\pairs} \\
\midrule
\endfirsthead
\toprule
Setting & $\delta$ & \shortstack{Flexible\\(at least)} & Singleton & Infeasible & Unresolved & \shortstack{Actions per\\pair} & \shortstack{Flexible\\pairs} \\
\midrule
\endhead
\bottomrule
\endfoot
21 actions, $T=5$ & 0.03 & 230 & 14 & 0 & 0 & 2.88 & 16 \\
21 actions, $T=5$ & 0.05 & 232 & 12 & 0 & 0 & 3.32 & 20 \\
21 actions, $T=5$ & 0.07 & 232 & 12 & 0 & 0 & 3.58 & 22 \\
21 actions, $T=10$ & 0.03 & 188 & 4 & 0 & 52 & 1.98 & 29 \\
21 actions, $T=10$ & 0.05 & 236 & 4 & 0 & 4 & 2.66 & 36 \\
21 actions, $T=10$ & 0.07 & 240 & 4 & 0 & 0 & 2.96 & 40 \\
6 actions, $T=5$ & 0.03 & 182 & 62 & 0 & 0 & 1.67 & 13 \\
6 actions, $T=5$ & 0.05 & 186 & 58 & 0 & 0 & 1.82 & 14 \\
6 actions, $T=5$ & 0.07 & 187 & 57 & 0 & 0 & 1.93 & 16 \\
6 actions, $T=10$ & 0.03 & 173 & 23 & 0 & 48 & 1.27 & 12 \\
6 actions, $T=10$ & 0.05 & 222 & 22 & 0 & 0 & 1.48 & 21 \\
6 actions, $T=10$ & 0.07 & 222 & 22 & 0 & 0 & 1.58 & 25 \\
\end{longtable}
{\footnotesize\noindent A pair is a living health state and decision year. Every setting uses the same 244 input-selected patients without survey weighting. Flexible counts are lower bounds from valid certificates. Singleton denotes patients with a singleton certificate and no certified flexible treatment plan. Size columns give medians among the certified flexible treatment plans. No infeasibility proof is established for these settings.\par}

We next vary post-event dynamics and rewards to examine how assumptions about illness and later health affect flexibility. At 0.03 QALYs, absorbing post-event states yield certified flexible plans for at least 241 patients. Removing terminal reward yields at least 229, and removing living post-event rewards yields at least 212. A 60/40 coronary--stroke allocation yields at least 198. Table~\ref{tab:clinical-model-variants} gives each variant's outcomes.

\begin{longtable}{@{}p{1.65in}rrrrrrr@{}}
\caption{Certified treatment plans under alternative model assumptions.}\label{tab:clinical-model-variants}\\
\toprule
Setting & $\delta$ & \shortstack{Flexible\\(at least)} & Singleton & Infeasible & Unresolved & \shortstack{Actions per\\pair} & \shortstack{Flexible\\pairs} \\
\midrule
\endfirsthead
\toprule
Setting & $\delta$ & \shortstack{Flexible\\(at least)} & Singleton & Infeasible & Unresolved & \shortstack{Actions per\\pair} & \shortstack{Flexible\\pairs} \\
\midrule
\endhead
\bottomrule
\endfoot
60/40 event split & 0.03 & 198 & 3 & 0 & 43 & 3.16 & 33 \\
60/40 event split & 0.05 & 214 & 2 & 0 & 28 & 4.02 & 40 \\
60/40 event split & 0.07 & 219 & 5 & 0 & 20 & 4.60 & 46 \\
Absorbing post-event states & 0.03 & 241 & 0 & 0 & 3 & 6.98 & 55 \\
Absorbing post-event states & 0.05 & 244 & 0 & 0 & 0 & 9.54 & 56 \\
Absorbing post-event states & 0.07 & 244 & 0 & 0 & 0 & 9.67 & 56 \\
Zero post-event rewards & 0.03 & 212 & 8 & 0 & 24 & 3.73 & 37 \\
Zero post-event rewards & 0.05 & 221 & 12 & 0 & 11 & 4.28 & 44 \\
Zero post-event rewards & 0.07 & 227 & 10 & 0 & 7 & 5.27 & 47 \\
No terminal continuation & 0.03 & 229 & 0 & 0 & 15 & 6.68 & 46 \\
No terminal continuation & 0.05 & 242 & 0 & 0 & 2 & 9.14 & 57 \\
No terminal continuation & 0.07 & 244 & 0 & 0 & 0 & 9.74 & 59 \\
\end{longtable}
{\footnotesize\noindent A pair is a living health state and decision year. Every setting uses the same 244 input-selected patients without survey weighting. Flexible counts are lower bounds from valid certificates. Singleton denotes patients with a singleton certificate and no certified flexible treatment plan. Size columns give medians among the certified flexible treatment plans. No infeasibility proof is established for these settings.\par}

\subsection{Computation settings and unresolved existence}
\label{app:sensitivity-stopping}

A cover-region limit bounds the number of active regions in a complete parameter cover. An outer-box limit bounds the parameter boxes used for score enclosures. Candidate and repair limits bound numerical proposals, which require a certificate before acceptance. An enlargement limit bounds the number of action additions tested in the fixed order. Exact-search limits bound the effort spent proving that no common treatment plan meets the loss limit. Table~\ref{tab:clinical-computation-settings} states these limits and arithmetic settings by analysis. Ranges describe the settings of the certificates and checks supporting the results.

\begin{longtable}{@{}L{1.25in}L{2.95in}L{1.35in}@{}}
\caption{Computation limits and bounds for unresolved patients.}\label{tab:clinical-computation-settings}\\
\toprule
Analysis or patient & Limit or stopping condition & Value \\
\midrule
\endfirsthead
\toprule
Analysis or patient & Limit or stopping condition & Value \\
\midrule
\endhead
\bottomrule
\endfoot
Primary & Cover cells & 512--2,048 \\
Primary & Interval precision (bits) & 128--256 \\
Primary & Risk-score grid exponent & 40--46 \\
Primary & Inner enclosure divisions & 16--32 \\
Primary & Outer enclosure boxes & 64--512 \\
Primary & Candidate proposal allowance (seconds) & 600 \\
Primary & Candidate cover allowance (seconds) & 3,600 \\
Primary & Finite-list bound preparation allowance (seconds) & 3,600 \\
Primary & Exact existence allowance (seconds) & 5,400 \\
Primary & Exact existence branch nodes & 200,000 \\
Primary & Model-list size & 64 \\
Primary & Candidates per proposal round & 4 \\
Primary & Threshold evaluation allowance (seconds) & 1,800 \\
Primary & Negative proof check allowance (seconds) & 7,200 \\
Primary & Treatment-plan check allowance (seconds) & 7,200--43,200 \\
Primary & Enlargement allowance (seconds) & 14,400--43,200 \\
Primary & Enlargement work interval (seconds) & 7,200--28,800 \\
Primary & Initial-plan cover check allowance (seconds) & 900--1,800 \\
Primary & Minimum score-interval width & $2^{-46}$--$2^{-8}$ \\
Existence analysis & Candidate allowance (seconds) & 1,800--7,200 \\
Existence analysis & Proposal allowance (seconds) & 300--1,800 \\
Existence analysis & Cover refinement allowance (seconds) & 900--7,200 \\
Existence analysis & Finite-list bound preparation allowance (seconds) & 1,800 \\
Existence analysis & Exact existence allowance (seconds) & 7,200--36,000 \\
Existence analysis & Exact branch nodes & 2,000,000--8,000,000 \\
Existence analysis & Exact search passes & 1--2 \\
Existence analysis & Projected exact-search allowance (seconds) & 900--14,400 \\
Existence analysis & Paired-model exact allowance (seconds) & 1,200--7,200 \\
Existence analysis & Existence-search allowance per patient task (seconds) & 21,600--43,200 \\
Existence analysis & Proof-check allowance (seconds) & 43,200 \\
Existence analysis & Cover cells & 512--2,048 \\
Existence analysis & Outer enclosure boxes & 64 \\
Existence analysis & Interval precision (bits) & 256 \\
Sensitivity settings & Seed candidates & 5 \\
Sensitivity settings & Candidate-proposal allowance (seconds) & 15 \\
Sensitivity settings & Candidate-check allowance (seconds) & 300 \\
Sensitivity settings & Enlargement allowance (seconds) & 900 \\
Sensitivity settings & Hard per-patient allowance (seconds) & 14,400 \\
Sensitivity settings & Coordinate evaluations & 2,000 \\
Sensitivity settings & Cover cells & 512 \\
Sensitivity settings & Interval precision (bits) & 128 \\
Sensitivity settings & Risk-score grid exponent & 32--40 \\
Sensitivity settings & Inner enclosure divisions & 16 \\
Sensitivity settings & Outer enclosure boxes & 64 \\
Sensitivity settings & Minimum score-interval width & $2^{-11}$ \\
Sensitivity settings & Memory per worker (GiB) & 12 \\
Responsive model & Parameter-box cover cells & 512 \\
Responsive model & Method allowance across loss limits (seconds) & 1,800 \\
Responsive model & Initial allowance (seconds) & 600 \\
Statistical settings & Fixed-plan check allowance (seconds) & 300 \\
Statistical settings & Inference batches & 300 \\
Statistical settings & Observations per batch & 512 \\
Parameter-box checks & Parameter-box cover cells & 1--512 \\
Parameter-box checks & Method allowance across loss limits (seconds) & 1,800 \\
Parameter-box checks & Initial allowance (seconds) & 600 \\
Patient 17 & Exact-search procedure ends without decision & $[0.029885, 0.034536]$ \\
Patient 19 & Exact-search procedure ends without decision & $[0.028584, 0.032547]$ \\
Patient 46 & Exact-search procedure ends without decision & $[0.027983, 0.030362]$ \\
Patient 55 & Exact-search procedure ends without decision & $[0.029784, 0.032915]$ \\
Patient 158 & Exact-search procedure ends without decision & $[0.028277, 0.033108]$ \\
Patient 439 & Computation ends during search or check & $[0.028070, 0.030538]$ \\
Patient 710 & Exact-search procedure ends without decision & $[0.028387, 0.030836]$ \\
Patient 753 & Exact-search procedure ends without decision & $[0.028758, 0.031805]$ \\
Patient 893 & Exact-search procedure ends without decision & $[0.029198, 0.030033]$ \\
Patient 930 & Exact-search procedure ends without decision & $[0.026456, 0.032307]$ \\
Patient 1158 & Exact-search procedure ends without decision & $[0.029670, 0.031723]$ \\
Patient 1247 & Exact-search procedure ends without decision & $[0.028933, 0.032852]$ \\
Patient 1342 & Exact-search procedure ends without decision & $[0.026164, 0.032512]$ \\
Patient 1381 & Computation ends during search or check & $[0.028520, 0.030557]$ \\
Patient 1665 & Exact-search procedure ends without decision & $[0.029299, 0.031443]$ \\
Patient 1731 & Exact-search procedure ends without decision & $[0.027937, 0.031540]$ \\
Patient 1989 & Exact-search procedure ends without decision & $[0.027623, 0.031167]$ \\
Patient 2055 & Exact-search procedure ends without decision & $[0.028050, 0.030232]$ \\
Patient 2243 & Exact-search procedure ends without decision & $[0.028271, 0.032100]$ \\
Patient 2364 & Exact-search procedure ends without decision & $[0.026970, 0.030440]$ \\
Patient 2365 & Exact-search procedure ends without decision & $[0.028522, 0.030726]$ \\
Patient 2369 & Exact-search procedure ends without decision & $[0.029478, 0.030530]$ \\
Patient 2463 & Computation ends during search or check & $[0.027519, 0.031020]$ \\
Patient 2501 & Exact-search procedure ends without decision & $[0.028269, 0.032636]$ \\
Patient 2513 & Exact-search procedure ends without decision & $[0.029313, 0.033976]$ \\
Patient 2533 & Exact-search procedure ends without decision & $[0.027798, 0.033142]$ \\
Patient 2557 & Exact-search procedure ends without decision & $[0.027379, 0.030653]$ \\
Patient 2582 & Exact-search procedure ends without decision & $[0.027535, 0.032316]$ \\
Patient 2620 & Computation ends during search or check & $[0.025763, 0.030088]$ \\
Patient 2734 & Exact-search procedure ends without decision & $[0.029415, 0.031036]$ \\
Patient 2985 & Exact-search procedure ends without decision & $[0.029834, 0.030384]$ \\
Patient 3059 & Exact-search procedure ends without decision & $[0.027929, 0.030284]$ \\
Patient 3169 & Computation ends during search or check & $[0.029968, 0.030582]$ \\
Patient 3199 & Exact-search procedure ends without decision & $[0.029093, 0.030305]$ \\
Patient 3341 & Exact-search procedure ends without decision & $[0.028989, 0.031633]$ \\
Patient 3481 & Exact-search procedure ends without decision & $[0.027692, 0.030578]$ \\
Patient 3756 & Computation ends during search or check & $[0.026569, 0.031459]$ \\
Patient 3759 & Computation ends during search or check & $[0.027666, 0.033589]$ \\
Patient 3908 & Exact-search procedure ends without decision & $[0.028274, 0.030081]$ \\
Patient 3944 & Exact-search procedure ends without decision & $[0.024700, 0.030026]$ \\
Patient 3947 & Computation ends during search or check & $[0.026189, 0.030009]$ \\
Patient 3956 & Exact-search procedure ends without decision & $[0.025656, 0.030138]$ \\
Patient 4082 & Exact-search procedure ends without decision & $[0.026843, 0.031563]$ \\
Patient 4177 & Computation ends during search or check & $[0.028157, 0.030335]$ \\
Patient 4348 & Exact-search procedure ends without decision & $[0.026887, 0.030789]$ \\
Patient 4517 & Computation ends during search or check & $[0.028840, 0.030843]$ \\
Patient 4575 & Exact-search procedure ends without decision & $[0.026224, 0.030847]$ \\
\end{longtable}
{\footnotesize\noindent Allowances specify maximum computation. Ranges describe limits that vary across patients. Plans certified during existence analysis use its listed limits. Task allowances are shared by their component operations. Nested allowances are not additive. The table lists operation limits, not cumulative work across all operations for a patient. Each proposal round generates candidate plans for cover checks. An exact-search pass applies the existence search to a finite model list. Grid spacing is two raised to the negative of the grid exponent, in score units. Interval-arithmetic precision is specified separately. The interval for each unresolved patient bounds its smallest achievable worst-case loss in QALYs. Printed lower endpoints round down and upper endpoints round up. The exact-search procedure ends without a decision for 37 patients, and computation ends during a search or check for 10 patients. Sensitivity settings apply to the width/loss-limit, action-set/horizon, model-variant, and statistical-subset analyses. The primary panel setting uses primary limits. The responsive-model and statistical rows give their additional limits. Parameter-box checks use the listed cover and per-method limits. A work interval bounds one continuation of the enlargement procedure.\par}

The unresolved-patient display in \cref{app:unresolved-existence-bounds} reports lower and upper bounds on the smallest achievable worst-case loss, together with the stopping reasons. Testing each missing action once yields a certified plan whose action count can be below the largest feasible count.

\subsection{Illustrative Patient Treatment Plans}
\label{app:representative-patients}

Four illustrative patients show the treatments within individual certified treatment plans (Figure~\ref{fig:representative-patients}). Within each baseline systolic BP group, certified flexible patients are ordered by nominal annual cardiovascular risk, with patient identifier breaking ties, and the lower median rank is selected. The selected identifiers are 4406, 1933, 3451, and 4358 in increasing systolic BP group order.

Each panel displays the healthy-state menu at every decision epoch. Three paths show the choices of the best-QALY, fewest-current-medications, and fewest-regimen-changes rules conditional on remaining healthy. The last rule uses its preceding action at each epoch, with no initial switching cost. Fixed offsets distinguish drug combinations with the same medication count.

\begin{figure}[!htbp]
\centering
\includegraphics[width=\textwidth]{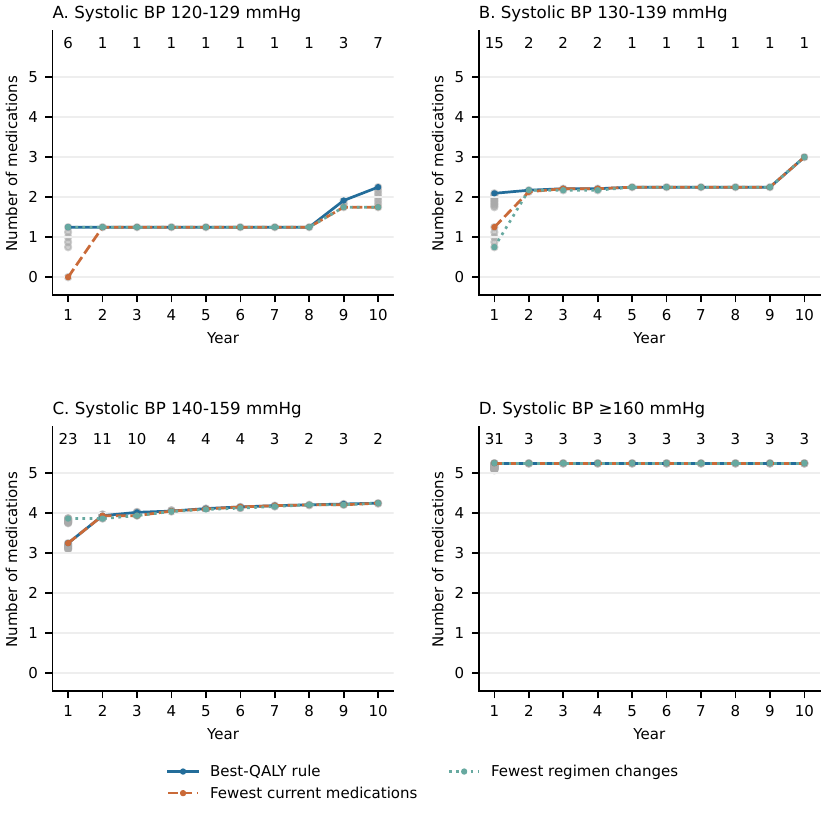}
\caption{Menus and selection-rule paths conditional on remaining healthy. Panels A--D represent baseline systolic BP groups 120--129, 130--139, 140--159, and at least 160 mmHg. Dots denote drug combinations and numbers give their count at each decision epoch. Paths overlap when the rules choose the same action.}
\label{fig:representative-patients}
\end{figure}

For patient 4406, the best-QALY, fewest-current-medications, and fewest-regimen-changes rules begin with one, zero, and one medication, respectively. Their initial counts are two, one, and one for patient 1933. They are three, three, and four for patient 3451. All three rules begin with five medications for patient 4358.

\FloatBarrier
\end{APPENDICES}
\end{document}